%% file: main.tex
\documentclass{article}

\input{settings/packages}
\input{settings/commands}

\begin{document}

\input{settings/cover}

\input{sections/abstract}

\section{Introduction}
\input{sections/introduction}

\section{Training trees}
\input{sections/defining_trees}

\section{Kauri: KMeans as unsupervised reward ideal}
\input{sections/kauri}

\section{Experiments}
\input{sections/experiments}

\section{Final words}
\input{sections/conclusion}

\bibliography{bibliography}
\bibliographystyle{plainnat}

\newpage
\appendix
\onecolumn

\section{Similarities with hierarchical clustering}
\label{app:hierarchical_clustering}
\input{appendix/related_hierarchical_methods}

\section{Proof of equality between centroid distances and sample-wise distances}
\label{app:centroid_equality}
\input{appendix/proof_equality}

\section{Relationship of kernel KMeans and MMD-GEMINI}
\label{app:kauri_objective}
\input{appendix/kauri_objective}

\section{Demonstration of expressions for the KAURI gains}
\label{app:kauri_gains}
\input{appendix/expression_gains}

\section{A fast implementation for KAURI}
\label{app:implementation_kauri}
\input{appendix/dynamic_algorithm}

\section{Datasets characteristics}
\label{app:dataset_summary}
\input{appendix/dataset_characteristics}

\section{Related motivating examples}
\label{app:imm_example}
\input{appendix/example_imm}


\end{document}

%% file: settings/packages.tex
\usepackage{graphicx}       
\usepackage{booktabs}       
\usepackage{subfig}         

\usepackage{amsmath, amssymb}       
\usepackage{stmaryrd}
\usepackage{bbm}                    

\usepackage{wrapfig}                

\usepackage{tikz}%
\usetikzlibrary{positioning, calc, decorations.pathreplacing, patterns, backgrounds}%
\usepackage{environ} 

\usepackage{natbib}
\setcitestyle{round,authoryear}

\usepackage{url}

\usepackage{algorithm}
\usepackage{algorithmic}

\usepackage[margin=1in]{geometry}

\usepackage{pgfplots}
\DeclareUnicodeCharacter{2212}{−}
\usepgfplotslibrary{groupplots,dateplot}
\usetikzlibrary{patterns,shapes.arrows}
\pgfplotsset{compat=newest}


%% file: settings/commands.tex
\renewcommand{\vec}{\pmb}
\newcommand{\x}{\vec{x}}

\newcommand{\real}{\mathbb{R}}
\newcommand{\rkhs}{\mathcal{H}}

\newcommand{\I}{\mathcal{I}}

\newcommand{\dataspace}{\mathcal{X}}
\newcommand{\cluster}{\mathcal{C}}
\newcommand{\dataset}{\mathcal{D}}
\newcommand{\objective}{\mathcal{L}}
\renewcommand{\split}{\mathcal{S}}
\newcommand{\leaf}{\mathcal{T}}

\newcommand{\stargain}{\Delta\objective^\star}
\newcommand{\doublestargain}{\Delta\objective^{\star\star}}
\newcommand{\switchgain}{\Delta\objective^\rightleftarrows}
\newcommand{\reallocationgain}{\Delta\objective^{\hookrightarrow}}

\newcommand{\Kmax}{K_{\max}}
\newcommand{\dmax}{d_{\max}}
\newcommand{\Tmax}{T_{\max}}

\newcommand{\p}{p_\theta}

\NewDocumentCommand{\distance}{o o m}{
  \ensuremath{\IfNoValueTF{#1}
    {#3}
    {\IfNoValueTF{#2}
     {#3}
     {#3\left( #1 \| #2 \right)}
    }}%
}
\NewDocumentCommand{\ipm}{o o}{
    \distance[#1][#2]{D_\text{IPM}}
}
\NewDocumentCommand{\fdiv}{o o}{
    \distance[#1][#2]{D_f}
}
\NewDocumentCommand{\hellinger}{o o}{
    \distance[#1][#2]{D_{H^2}}
}
\NewDocumentCommand{\kl}{o o}{
    \distance[#1][#2]{D_\text{KL}}
}
\NewDocumentCommand{\mmd}{o o}{
    \distance[#1][#2]{D_\text{MMD}}
}
\NewDocumentCommand{\squaredmmd}{o o}{
    \distance[#1][#2]{D_{\text{MMD}^2}}
}
\NewDocumentCommand{\wasserstein}{o o}{
    \distance[#1][#2]{D_\mathcal{W}}
}
\NewDocumentCommand{\tv}{o o}{
    \distance[#1][#2]{D_\text{TV}}
}

\NewDocumentCommand{\klsym}{o o}{
    \distance[#1][#2]{D_\text{KL-sym}}
}

\newcommand{\second}{{\prime\prime}}
\newcommand{\inv}[1]{\frac{1}{#1}}

\newcommand{\card}[1]{\ensuremath{\left|#1\right|}}

\NewDocumentCommand{\norm}{m}{
 \ensuremath{\|#1\|}
}

\NewDocumentCommand{\ind}{o}{\ensuremath{
    \IfNoValueTF{#1}
        {\mathbbm{1}}
        {\mathbbm{1}\!\left[#1\right]}
    }
}

\newcommand{\expectation}[2]{\mathbb{E}_{#1}\left[#2\right]}

\NewDocumentCommand{\gemini}{o o m}{\ensuremath{
    \IfNoValueTF{#1}
    {\IfNoValueTF{#2}
       {\I}
       {\I^{#2}}
    }
    {\IfNoValueTF{#2}
       {\I_{#1}}
       {\I_{#1}^{#2}}
    }
    \left(#3\right)
}}

\renewcommand{\cite}{\citep}

\newcommand*\Let[2]{\STATE #1 $\gets$ #2}

\newcommand{\std}[1]{\textsubscript{#1}}

\makeatletter
\newsavebox{\measure@tikzpicture}
\NewEnviron{scaletikzpicturetowidth}[1]{%
  \def\tikz@width{#1}%
  \def\tikzscale{1}\begin{lrbox}{\measure@tikzpicture}%
  \BODY
  \end{lrbox}%
  \pgfmathparse{#1/\wd\measure@tikzpicture}%
  \edef\tikzscale{\pgfmathresult}%
  \BODY
}
\makeatother

%% file: settings/cover.tex










\begin{center}
    {\Large
     {\sc Kernel KMeans clustering splits for end-to-end unsupervised decision trees}
    }

    \bigskip

    Louis Ohl \textsuperscript{1,2,3} \qquad Pierre-Alexandre Mattei \textsuperscript{1,4} \qquad Mickaël Leclercq \textsuperscript{2}\\Arnaud Droit \textsuperscript{2} \qquad Frédéric Precioso \textsuperscript{1,3}

    \bigskip

    {\it
    \textsuperscript{1} Université Côte d'Azur, Inria, Maasai team, I3S / LJAD, CNRS\\
    \textsuperscript{2} Université Laval, CHU de Québec Research Centre, ADLab\\
    \textsuperscript{3} I3S, CNRS\\
    \textsuperscript{4} LJAD CNRS\\
    }
\end{center}

%% file: sections/abstract.tex
\begin{abstract}
    Trees are convenient models for obtaining explainable predictions on relatively small datasets. Although there are many proposals for the end-to-end construction of such trees in supervised learning, learning a tree end-to-end for clustering without labels remains an open challenge. As most works focus on interpreting with trees the result of another clustering algorithm, we present here a novel end-to-end trained unsupervised binary tree for clustering: Kauri. This method performs a greedy maximisation of the kernel KMeans objective without requiring the definition of centroids. We compare this model on multiple datasets with recent unsupervised trees and show that Kauri performs identically when using a linear kernel. For other kernels, Kauri often outperforms the concatenation of kernel KMeans and a CART decision tree.
\end{abstract}

%% file: sections/introduction.tex
Decision tree classifiers are one of the most intuitive models in machine learning owing to their intrinsic interpretability~\cite[Section 3.2]{molnar_interpretable_2020}. Trees consist of a set of hierarchically sorted nodes starting from one single root node. Each node comprises two or more conditions called rules, each of which leading to a different child node. Once a node does not have any child, a decision is returned. A childless node is named a leaf.


While the end model is eventually interpretable, building it implies some questions to be addressed, notably regarding the number of nodes, the feature (or set of features) on which to apply a decision rule, the construction of a decision rule i.e. the number of thresholds and hence the number of children per node. Learning the structure is easier in the case of supervised learning, whereas the absence of labels makes the construction of unsupervised trees more challenging. In recent related works, the problem was oftentimes addressed with twofold methods~\cite{tavallali_k-means_2021, laber_shallow_2023}: first learning clusters using another algorithm e.g. KMeans, then applying a supervised decision tree to uncover explanations of the clusters. However, such \emph{unsupervised trees} are not fully unsupervised in fact since their training still requires the presence of external labels for guidance which are provided by KMeans.

To alleviate this dependence on an exterior clustering algorithm:

\begin{itemize}
    \item We show how the kernel KMeans objective can be rephrased to avoid the computations of centroids, leveraging simple gains to compute for split proposals in decision trees.
    \item We then introduce an end-to-end unsupervised tree for clustering: KMeans as unsupervised reward ideal (Kauri), which is not restricted to a fixed number of leaves. To the best of our knowledge, this is the first end-to-end kernelised clustering tree.
    \item We show that Kauri displays equal performance in clustering to kernel KMeans+Tree using end-to-end training while obtaining shallower structures, notably for kernels other than the linear kernel. 
    \item We finally show that Kauri addresses some limitations of the kernel KMeans algorithm.
\end{itemize}

%% file: sections/defining_trees.tex

\subsection{How do we train supervised trees?}

In supervised learning, we have access to targets $y$ which guides our tree construction for separating our samples. In this field, we can refer to the well-known classification and regression tree (CART)~\cite{breiman_classification_1984}. At each node, we evaluate the quality of a split, i.e. a proposed rule on a given feature and data-dependent threshold, through gain metrics. We then add to the tree structure the split that achieved the highest possible gain. Common implementations of supervised trees use the Gini criterion developed by the statistician Corrado~\citet{gini_variability_1912}, which indicates how \emph{pure} a tree node is given the proportion of different labels in its samples \citep{casquilho_ginisimpson_2018}. Later works then proposed different gain metrics like the difference of mutual information in the ID3~\cite{quinlan_induction_1986} and C4.5~\cite{quinlan_c4_2014} algorithms.

The greedy nature of tree optimisation can lead to the construction of very deep trees, which harms the interpretable nature of the model~\cite{lustrek_what_2016}. This motivates for example the construction of multiple trees that are equivalent in terms of decision, yet different in terms of structure presenting thus an overview of the Rashomon set for interpretations~\cite{xin_exploring_2022}. Other approaches tried to overcome the deterministic non-differentiable nature of the rule-based tree by introducing differentiable leaves~\cite{fang_unsupervised_1991} which allows to train trees through gradient descent. For example, \citet{yang_deep_2018} build a set of differentiable bins per feature which combinations through a matrix multiplication returns the prediction.


\subsection{How do we train unsupervised trees?}

In clustering, we do not have access to labels making all previous notions of gains unusable, so we need other tools for guiding the splitting procedure of the decision trees. A common approach is then to keep the algorithm supervised as described in the previous section, yet providing labels that were derived from a clustering algorithm e.g. KMeans~\cite{laber_shallow_2023, held_unsupervised_1997}. In this sense, centroids derived from KMeans can also be involved in split procedures~\cite{tavallali_k-means_2021}, even to the point that the data from which the centroids are derived do not need to be collected~\cite{gamlath_nearly-tight_2021}. However, such methods do not properly construct the tree \emph{from scratch} in an unsupervised way despite potential changes in the gain formulations. We are interested in a method that can provide a directly integrated objective to optimise tree training. Other gains derived from entropy formulations have also been proposed~\cite{bock_information_1994, basak_interpretable_2005}. We even note the use of mutual information to achieve deeper and deeper refinements of binary clusters~\cite{karakos_unsupervised_2005}.

Oftentimes, these approaches assume that a leaf describes fully a cluster, e.g. \citet{blockeel_top-down_1998}. Combining leaves into a single cluster requires then post hoc methods~\cite{fraiman_interpretable_2013}. In such a case, an elegant approach for constructing an unsupervised tree was proposed by~\citet{liu_clustering_2000} by adding uniform noise to the data and assigning a decision tree to separate the noise from the true data. Such trees put in different leaves dense areas of the data, which can then be labelled manually.

To ensure that several leaves can be assigned to a single cluster, related work also focused on the complete initialisation of a tree and refinement according to a global objective function. For example, \citet{bertsimas_interpretable_2021} directly maximise the silhouette score or the Dunn index, which are internal clustering metrics and require the initialisation of the tree through greedy construction or KMeans labels. The objective is optimised using a mixed integer optimisation formulation of the tree structure. Lately, \citet{gabidolla_optimal_2022} proposed to optimise an oblique tree, a structure with logistic regressions at each node, through the alternative optimisation of a distance-based objective, e.g. KMeans, providing pseudo-labels to a tree alternating optimisation problem~\citep{carreira-perpinan_alternating_2018}.

When no rules need to be specified in the tree architecture, the problem becomes related to divisive hierarchical clustering for which end-to-end solutions exist, as discussed in App.~\ref{app:hierarchical_clustering}.

%% file: sections/kauri.tex
The Kauri tree is a non-differentiable binary decision tree that looks in many ways alike the CART algorithm. It constructs from scratch a binary tree giving hard clustering assignments to the data by using an objective equivalent to the optimisation of a kernel KMeans. In the Kauri structure, a cluster can be described by several leaves.

\subsection{Objective function}
\label{ssec:kernel_kmeans_relationship}

We consider that we have a dataset of $n$ samples: $\dataset = \{\x_i\}_{i=1}^n$. 
We write the partition into $\Kmax$ clusters as:
\begin{equation}
    \cluster_k = \{\x_i \in \dataspace_k\}, \forall k \leq \Kmax,
\end{equation}
with $\{\dataspace_k\}_{k=1}^{\Kmax}$ a partition of the data space $\dataspace\subseteq \real^d$.

The kernel KMeans algorithm minimises the cluster sum of squares in a Hilbert space $\rkhs$ with projection $\varphi$ and kernel $\kappa$ with respect to $\Kmax$ centroids $\vec{\mu}_k \ldots \vec{\mu}_{\Kmax}$:

\begin{equation}\label{eq:kernel_kmeans}
    \objective_\text{KMeans} = \sum_{k=1}^{\Kmax} \sum_{\x\in\cluster_k}\norm{\varphi(\x) - \vec{\mu}_k}^2_\rkhs.
\end{equation}

Instead of computing the sample-wise distance to the centroid using the kernel trick~\citep{dhillon_kernel_2004}, Kauri optimises this objective without explicitly computing centroids per cluster. To that end, we use the following simple equality (detailed in App.~\ref{app:centroid_equality}):

\begin{equation}
\sum_{\x \in \cluster_k} \norm{\varphi(\x) - \vec{\mu}_k}^2_\rkhs  = \inv{2\card{\cluster_k}}\sum_{\x, \vec{y} \in \cluster_k} \norm{\varphi(\x) - \varphi(\vec{y})}^2_\rkhs.
\end{equation}
Once inserted into the kernel KMeans objective, we get an alternative formulation without centroids:
\begin{equation}
    \objective_\text{KMeans} = \sum_{k=1}^{\Kmax} \inv{2\card{\cluster_k}}\sum_{\x,\vec{y}\in\cluster_k} \norm{\varphi(\x)-\varphi(\vec{y})}^2_\rkhs.
\end{equation}
Using the kernel trick, we can rephrase this objective as:
\begin{equation}
    \objective_\text{KMeans} = \sum_{k=1}^{\Kmax} \inv{2\card{\cluster_k}} \sum_{\x,\vec{y}\in\cluster_k} (\kappa(\x,\x) + \kappa(\vec{y},\vec{y}) - 2\kappa(\x,\vec{y}) ),
\end{equation}
where the two first kernel terms can be summarised as the size of clusters weighting the diagonal elements of the kernel. Finally, the third term is the grand sum of the kernel of the cluster, and thus:
\begin{equation}\label{eq:alternative_kernel_kmeans}
    \objective_\text{KMeans} = \sum_{\x\in\dataset} \kappa(\x,\x)-\sum_{k=1}^{\Kmax} \inv{\card{\cluster_k}} \sum_{\x,\vec{y}\in\cluster_k} \kappa(\x,\vec{y}).
\end{equation}
For the sake of simplicity, we introduce the function $\sigma$ that sums the kernel values $\kappa(\x_i,\x_j)=\langle \varphi(\x_i), \varphi(\x_j)\rangle$ of samples indexed by two sets:
\begin{equation}
    \sigma(E \times F) = \sum_{\substack{\x_i\in E\\\x_j\in F}} \kappa(\x_i, \x_j).
\end{equation}
We will refer to the $\sigma$ function as the \emph{kernel stock}. This function is bilinear with respect to the input spaces. We provide in Figure~\ref{fig:kernel_stocks} a visual explanation of its different usages.

We finally note that the first term of Eq.~(\ref{eq:alternative_kernel_kmeans}) is a constant because it does not depend on the clustering. With the only the second term remaining, we can remove the minus sign and hence maximise the objective:
\begin{equation}\label{eq:objective}
    \objective_\text{Kauri} = \sum_{k=1}^{\Kmax} \frac{\sigma(\cluster_k^2)}{\card{\cluster_k}},
\end{equation}
Therefore, maximising our objective function $\objective$ is equivalent up to a constant to minimising a KMeans objective for any kernel. However, in contrast to Eq.~(\ref{eq:kernel_kmeans}), it is not a function of centroids, but a function of a partition. This objective can also be connected to MMD-GEMINI~\citep{ohl_generalised_2022} as shown in App.~\ref{app:kauri_objective}.

\input{figs_tex/fig_kernel_stocks}

\subsection{Tree branching}

For supervised trees like CART or ID3, the types of splits are binary and guided by the labels which tell us to which class each child node should go. For unsupervised trees, we must consider all possibilities: to which cluster goes the left child, to which cluster goes the right child on what set of features to do the split, on what threshold in this feature to split and on which nodes. We note $\leaf_p$ the set of samples reaching the $p$-th node. For a split, let $\split_L$ be the subset of samples from the node samples $\leaf_p$ that will go to the left child node and $\split_R$ the complementary subset of samples that will go to the right child node. Each child node will be assigned to a different cluster, whether new, already existing, or equal to the parent node's cluster assignment. Let $k_p$ be the current cluster membership of the parent node $p$, $k_L$ the future cluster membership for the left child node and $k_R$ the future cluster membership of the right child node, then $\split_L \cup \split_R = \leaf_p \subseteq \cluster_{k_p}$ and after splitting: $\split_L \subseteq \cluster_{k_L}$ and $\split_R\subseteq \cluster_{k_R}$.

We enforce the following constraints: a child node must stay in the parent node's cluster if both children leaving would empty the parent's cluster; the creation of a new cluster can only be done under the condition that the number of clusters does not exceed a specified limit $K_\text{max}$. We also impose a maximum number of leaves $\Tmax$ which can be equal to at most the number of samples $n$. It is nonetheless possible that the algorithm stops the splitting procedure if all gains become negative before reaching the maximum number of leaves allowed. Unlike~\citep{gabidolla_optimal_2022} who recently proposed unsupervised oblique trees, we choose to keep splits on a single feature because this lowers the complexity of greedy exploration.

Thus, learning consists in greedily exploring from all nodes the best split and either taking this split to build a new cluster or merging with another cluster. We now present the objective function and related gains depending on the children's cluster memberships.

\subsection{Gain metrics}
\label{ssec:gain_metrics}

We can derive from the objective of Eq.~(\ref{eq:objective}) four gains that evaluate how much score we get by assigning one child node to a new cluster, assigning both child nodes to two new clusters, merging one child node to another cluster, or merging both child nodes to different clusters. We denote by $\cluster^\prime_\bullet$ the clusters after the split operation and by $\cluster_\bullet$ the clusters before the split. Hence, the global gain metric is:
\begin{equation}
    \Delta\objective(\split_L: k_p\rightarrow k_L, \split_R: k_p \rightarrow k_R) = \frac{\sigma({\cluster^\prime_{k_L}}^2)}{\card{\cluster^\prime_{k_L}}}+\frac{\sigma({\cluster^\prime_{k_R}}^2)}{\card{\cluster^\prime_{k_R}}}+\frac{\sigma({\cluster^\prime_{k_p}}^2)}{\card{\cluster^\prime_{k_p}}}-\frac{\sigma(\cluster_{k_L}^2)}{\card{\cluster_{k_L}}}-\frac{\sigma(\cluster_{k_R}^2)}{\card{\cluster_{k_R}}}-\frac{\sigma(\cluster_{k_p}^2)}{\card{\cluster_{k_p}}},
\end{equation}
which corresponds to subtracting the contribution of the {kernel stocks} of the former clusters and adding the {kernel stocks} of the new clusters after splitting.

From this global gain metric, we derive four different gains: the \emph{star gain} $\stargain$ for assigning either the left or right child of a leaf to a new cluster, the \emph{double star gain} $\doublestargain$ for assigning the left and right children of a leaf to two new clusters, the \emph{switch gain} $\switchgain$ for assigning either the left or right child of a leaf to another existing cluster and the \emph{reallocation gain} $\reallocationgain$ for assigning respectively the left and right children to different existing clusters. In practice, for the $p$-th leaf with split proposals $\split_L$ and $\split_R$, we do not need to compute the future cluster kernel stocks $\sigma(\cluster^\prime_\bullet)$ and instead use the stocks between the splits and the current clusters: $\sigma(\split_{L/R} \times \cluster_\bullet)$. We provide a detailed explanation of the derivations of the gains in App.~\ref{app:kauri_gains} leading to the complete algorithm description in App.~\ref{app:implementation_kauri}. The code is provided in supplementary materials.

%% file: figs_tex/fig_kernel_stocks.tex
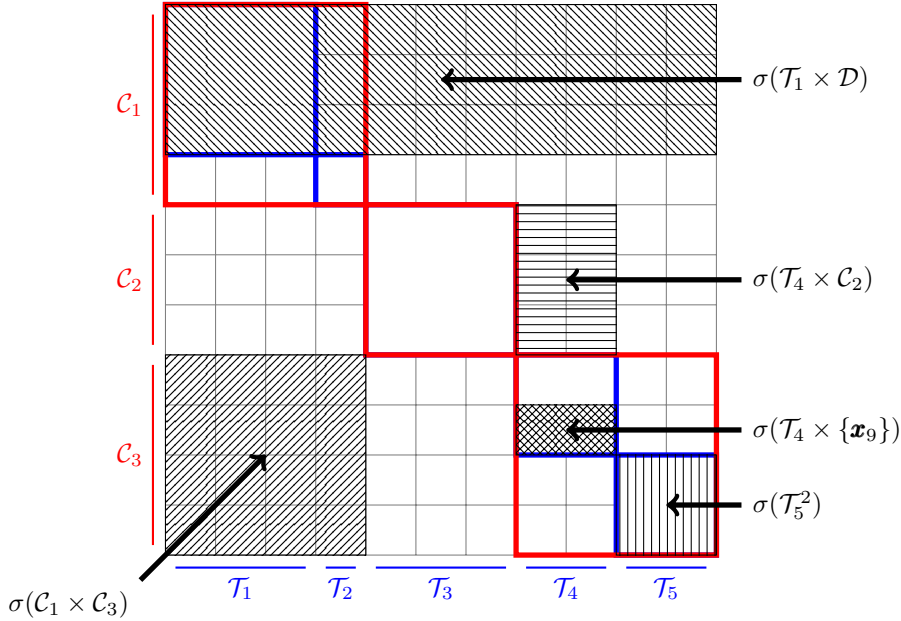
\begin{figure}
    \centering
    \begin{scaletikzpicturetowidth}{0.65\linewidth}
    \begin{tikzpicture}[scale=\tikzscale]
    
        \draw[step=1.0, gray, thin] (0, 0) grid (11,11);
        
        \draw[blue, line width = 0.2em, fill=white] (0,8) rectangle ++(3,3);
        \draw[blue, line width = 0.2em, fill=white] (3,7) rectangle ++(1,1);
        \draw[blue, line width = 0.2em, fill=white] (4,4) rectangle ++(3,3);
        \draw[blue, line width = 0.2em, fill=white] (7,2) rectangle ++(2,2);
        \draw[blue, line width = 0.2em, fill=white] (9,0) rectangle ++(2,2);
        
        \draw[red, line width = 0.2em] (0,7) rectangle ++(4,4);
        \draw[red, line width = 0.2em] (4,4) rectangle ++(3,3);
        \draw[red, line width = 0.2em] (7,0) rectangle ++(4,4);
        
        \draw[pattern = north west lines, distance=0.1pt] (0,8) rectangle ++(11,3); 
        \draw[pattern = horizontal lines] (7,4) rectangle ++(2,3); 
        \draw[pattern = north east lines] (0,0) rectangle ++(4,4); 
        \draw[pattern = vertical lines] (9,0) rectangle ++(2,2); 
        \draw[pattern = crosshatch] (7,2) rectangle ++(2,1); 
        
        \draw[red, thick] (-0.25, 7.2) -- +(0,3.6) node[midway, left] {$\cluster_1$};
        \draw[red, thick] (-0.25, 4.2) -- ++(0,2.6) node[midway, left] {$\cluster_2$};
        \draw[red, thick] (-0.25, 0.2) -- ++(0,3.6) node[midway, left] {$\cluster_3$};
        
        \draw[blue, thick] (0.2,-0.25) -- ++ (2.6, 0) node[midway, below] {$\leaf_1$};
        \draw[blue, thick] (3.2,-0.25) -- ++ (0.6, 0) node[midway, below] {$\leaf_2$};
        \draw[blue, thick] (4.2,-0.25) -- ++ (2.6, 0) node[midway, below] {$\leaf_3$};
        \draw[blue, thick] (7.2,-0.25) -- ++ (1.6, 0) node[midway, below] {$\leaf_4$};
        \draw[blue, thick] (9.2,-0.25) -- ++ (1.6, 0) node[midway, below] {$\leaf_5$};
        
        \draw[<-, black, line width = 0.2em] (5.5,9.5) -- +(6, 0) node[right] {$\sigma(\leaf_1\times \dataset)$};
        \draw[<-, black, line width = 0.2em] (8,5.5) -- +(3.5, 0) node[right] {$\sigma(\leaf_4\times \cluster_2)$};
        \draw[<-, black, line width = 0.2em] (2,2) -- +(-2.5, -2.5) node[below left] {$\sigma(\cluster_1\times \cluster_3)$};
        \draw[<-, black, line width = 0.2em] (10,1) -- +(1.5, 0) node[right] {$\sigma(\leaf_5^2)$};
        \draw[<-, black, line width = 0.2em] (8,2.5) -- +(3.5, 0) node[right] {$\sigma(\leaf_4\times \{\x_9\})$};
        
    \end{tikzpicture}\end{scaletikzpicturetowidth}
    
    \caption{A toy example with a dataset consisting of 11 samples partitioned in 3 clusters using 5 leaves in a tree. The matrix represents the kernel between all pairs of samples and dashed areas correspond to the sum of kernel elements according to the \emph{kernel stock} function $\sigma$.}
    \label{fig:kernel_stocks}
\end{figure}

%% file: sections/experiments.tex
\input{figs_tex/tab_benchmark_small_ari}
\input{figs_tex/tab_benchmark_small_wad}

\subsection{General setup}

We covered datasets used for benchmarking by~\citet{laber_shallow_2023} and~\citet{frost_exkmc_2020} and those from the Fundamental Clustering Data Suite by~\citet{bertsimas_interpretable_2021}. We chose to present only a subset of the tested datasets in the main part of the paper and put all the remaining datasets in App.~\ref{app:dataset_summary} to avoid lengthy tables. The summary of the datasets used in the experiments can also be seen in App.~\ref{app:dataset_summary}. We will assess the general clustering performances and explanation power of the models before showing qualitative examples.

We used minmax scaling for all datasets in order to be comparable with the ICOT~\citep{bertsimas_interpretable_2021} method, which assumes features in the [0,1] range. All categorical variables were one-hot-encoded, except for the US congressional votes dataset, where we encoded specifically the answer yes as 1, the no as -1 and the unknown votes as 0. In other datasets, we tossed away all samples that presented missing values for the sake of simplicity.

We used 4 different metrics to assess performance: the adjusted rand index (ARI, \citealp{hubert_comparing_1985}) and the kernel KMeans score normalised by the reference score of the sole kernel KMeans algorithm for cluster quality, and the weighted average depth (WAD, \citealp{mcsherry_explanation_2002, lustrek_what_2016}) or weighted average explanation size (WAES, \citealp{laber_shallow_2023}) for the parsimony of the tree structure.

We compare the performances of Kauri against recent methods for unsupervised tree constructions, namely ExShallow~\citep{laber_shallow_2023}, RDM~\citep{makarychev_explainable_2022}, IMM~\citep{moshkovitz_explainable_2020}, ExKMC~\citep{frost_exkmc_2020} and ICOT~\citep{bertsimas_interpretable_2021} for which we found available packages or implementations. These methods are twofold and start by fitting KMeans centroids to the data. Then, they learn a tree to explain the clusters obtained. The differences in all methods lie in attempts to limit the depth of the tree for the sake of simple explanations as deep trees tend to lose expressivity in explanation. Specifically in the case of ICOT, the KMeans only serves as a warm-start initialisation for a greedy tree, which is then refined through a mixed integer optimisation problem. We also choose to provide a combination of kernel KMeans and a standard CART decision tree classifier as a baseline for all algorithms, noted KMeans+DT. To the best of our knowledge, only ExKMC~\citep{frost_exkmc_2020} was surprisingly compared to such baseline, and \citet{bertsimas_interpretable_2021} compared their model to KMeans.

\subsection{Performances with as many leaves as clusters}
\label{ssec:benchmark}

As some related works focus on trees with one leaf per cluster, we limit the Kauri tree and the KMeans+DT to as many leaves as clusters. However, we could not limit the number of leaves for ICOT, so we restrained the depth to 5 levels. Consequently, the scores of ICOT may outperform other methods due to a more free architecture.

Since some algorithms are deterministic in nature, we introduce stochasticity in results by selecting 80\% of the training data over 30 runs. We report the ARI for all algorithms in Table~\ref{tab:benchmark_small_ari} and the WAD in Table~\ref{tab:benchmark_small_wad}. The results for the kernel KMeans score were left in App.~\ref{app:dataset_summary} because they are very similar to the ARI scores.

We observe in Table~\ref{tab:benchmark_small_ari} that Kauri often performs on par with related works. In general, the performances of all methods are quite similar regarding the ARI. In particular, the ARI of Kauri is often close to the baseline KMeans + DT, except for the wine dataset. We believe that the differences of scores are in general negligible as they are covered by the standard deviations throughout the multiple subsets of data. However, for very different scores like the ARI on the Target dataset, we believe that this difference can be explained by the order of the choice of splits in the trees owing to the presence of the KMeans objective among methods or just the usage of labels. We did not manage to run ICOT on datasets with a large number of features or clusters. This joins the observations of~\citet{bertsimas_interpretable_2021} who reported a runtime of more than an hour for the Engytime dataset. Finally, we observe good performances in Table~\ref{tab:benchmark_small_wad}  regarding the depth of the nodes at which the cluster decisions are returned. We purposefully removed in Table~\ref{tab:benchmark_small_wad} the Atom, Cancer, and Chainlink datasets because they are binary, which implies that the WAD is always equal to 2 when the model is constrained to 2 leaves. Moreover, we did not manage to obtain the tree structure underlying in ICOT to compute the WAD.

\subsection{Performances with more leaves than clusters}

\input{figs_tex/tab_benchmark_large_ari}

\input{figs_tex/tab_benchmark_large_waes}
\input{figs_tex/fig_synthetic}


We run here the exact same benchmark as proposed in section~\ref{ssec:benchmark}, except that we seek to compare Kauri with the ExKMC method. To this end, all trees are now limited to 4 times more leaves than clusters following the result of~\citet{frost_exkmc_2020}. Contrary to section~\ref{ssec:benchmark}, the excessive number of leaves will necessarily imply that multiple leaves might explain a single cluster. We chose then to measure the WAES because we wanted to emphasise the complexity of the explanation of a cluster rather than the depth of trees. We report the ARI in Table~\ref{tab:benchmark_large_ari} and the WAES in Table~\ref{tab:benchmark_large_waes}.

We observe in Table~\ref{tab:benchmark_large_ari} that Kauri often performs on par with ExKMC and the baseline regarding the clustering performances. However, we find that our WAES scores are steadily lower with the baseline compared to ExKMC on most datasets in Table~\ref{tab:benchmark_large_waes}. In this context, we believe that the difference between the baseline and Kauri is mainly due to the contrast between the respective optimisation of a Gini impurity on clustering labels and the greedy optimisation of KMeans.



To highlight some differences in behaviour between Kauri and KMeans + DT, we show in Figure~\ref{fig:synthetic} how the angles of the decision boundary and the number of samples in the dataset can change the performances in seemingly identical distributions. Indeed, KMeans easily builds linear boundaries that are not axis-aligned,  hence as the boundaries become less and less aligned with the axes, the decision trees struggle to maintain a low number of leaves to mimic these ``diagonal" boundaries. This effect gets worse if the number of samples to separate is high on this decision boundary. However, as soon as the decision boundaries are axis-aligned, the decision tree becomes again a fierce competitor. Both trees have unlimited leaves and stop only when no gain is longer possible. 

\subsection{Varying the kernel}

\input{figs_tex/tab_kernels_ari}
\input{figs_tex/tab_kernels_kmeans}

\input{figs_tex/fig_wine_pca}

Our comparisons have been done so far with related works using the linear kernel in kernel KMeans. However, using a different kernel leads to the absence of a definition of centroids in the Euclidean space where the data lie. Consequently, previously compared methods are not compatible with such a setup because they generally require a centroid, and we are only left with the KMeans+DT baseline as a competitor. We explore 4 different kernels with default parameters from \texttt{scikit-learn}~\citep{pedregosa_scikit-learn_2011}: $\chi^2$, additive $\chi^2$, Laplacian and RBF kernels with Table~\ref{tab:benchmark_kernels_ari} for the ARI and Table~\ref{tab:benchmark_kernels_kmeans} for the normalised KMeans score.

We generally observe equal or stronger ARI scores for the KAURI algorithm in Table~\ref{tab:benchmark_kernels_ari}. The same observation goes for the KMeans score in Table~\ref{tab:benchmark_kernels_kmeans}, especially for the Laplacian kernel in both tables. Note that we had to reduce the number of datasets to present in both tables. Other results are described in App.~\ref{app:dataset_summary}. This is due to the phenomenon of empty clusters that arises in the kernel KMeans algorithm. Consequently, the basis on which the decision tree is learnt does not provide enough clusters, thus lowering the ARI and increasing the kernel KMeans score. Similarly, the reference kernel KMeans score used for normalisation suffered from the same problem. Empirically, we observed this behaviour for 2 implementations of the kernel KMeans algorithm.\footnote{\url{https://gist.github.com/mblondel/6230787}, and the \texttt{tslearn} v0.6.3 implementation~\citep{tavenard_tslearn_2020}.}

\input{figs_tex/fig_kernel_kmeans_clusters}
\input{figs_tex/fig_us_votes_tree}

As a simple example, we ran 100 kernel KMeans models with an additive $\chi^2$ kernel or a polynomial kernel for the Mice protein dataset and the Target dataset. From Figure~\ref{fig:kernel_kmeans_clusters}, we observe that for a relatively small number of clusters, the kernel KMeans may converge to several empty clusters, with the worst effect from the polynomial kernel in this example. To complete Figure~\ref{fig:kernel_kmeans_clusters}, the number of clusters found with the best kernel KMeans score was 8 for the Mice protein dataset and 6 for the Target dataset. In contrast, datasets with only 2 clusters to find often converged to the good number of clusters, making scores comparable.

This shows that with the end-to-end construction of the Kauri tree, we do not suffer from dependence to the basis KMeans algorithm, and manage to get the correct user-desired number of clusters.

In the case of the presented datasets in Table~\ref{tab:benchmark_kernels_ari} where kernel KMeans managed to converge to the desired number of clusters, we think that the improved performances by Kauri come from its greedy nature. The greedy optimisation leads to a local minimum where no branch can be further found, even with fewer leaves than the maximum specified. Consequently, the obtained tree remains in a simpler state than a decision tree that would overfit the labels from kernel KMeans. We provide a simple visualisation through the PCA and clustering of the wine dataset in Figure~\ref{fig:pca_wine}.

\subsection{A qualitative example of the obtained decision tree}

In this example, we focus on the congressional votes dataset which details 16 key votes from the 435 members of the US Congress in 1985. The targets of the dataset are the Republican or Democrat affiliations of the voters. Note that with a linear kernel, unknown votes encoded as 0 do not favour a ``yes'' (1) or ``no'' (-1) vote. The Kauri tree that was fitted on this dataset is described in Figure~\ref{fig:kauri_us_votes}. The obtained clusters translate very well the Republican and Democrat opposition through arming and international assistance, with one cluster containing 73\% of Republicans and the second one adding up to 96\% of Democrats. The ARI is 0.47 for this tree which corresponds to an unsupervised accuracy of 84\%.

%% file: figs_tex/tab_benchmark_small_ari.tex
\begin{table*}
    \centering
    \caption{ARI scores \std{std} (greater is better) after 30 runs on random subsamples of 80\% of the input datasets. Entries marked X were not run because of excessive runtime due to large numbers of features. All models are limited to finding as many leaves as clusters.}
    \label{tab:benchmark_small_ari}
    \begin{tabular}{ccccccc}
        \toprule
        Dataset & Kauri & KMeans+DT & ICOT & IMM & ExShallow & RDM \\
        \midrule
        Atom & 0.19\std{0.03} & 0.17\std{0.03} & 0.17\std{0.05} & \textbf{0.20\std{0.04}} & 0.19\std{0.03} & 0.17\std{0.01} \\
        Cancer & 0.74\std{0.01} & 0.73\std{0.01} & \textbf{0.78\std{0.05}} & 0.73\std{0.02} & 0.73\std{0.01} & 0.65\std{0.02} \\
        Chainlink & \textbf{0.10\std{0.01}} & 0.09\std{0.02} & 0.08\std{0.04} & \textbf{0.10\std{0.01}} & 0.09\std{0.01} & 0.08\std{0.01} \\
        Digits & 0.40\std{0.04} & 0.42\std{0.03} & X & 0.36\std{0.04} & \textbf{0.43\std{0.02}} & 0.23\std{0.04} \\
        Hepta & \textbf{1.00\std{0.01}} & \textbf{1.00\std{0.00}} & 0.79\std{0.16} & \textbf{1.00\std{0.00}} & \textbf{1.00\std{0.00}} & 0.93\std{0.02} \\
        Iris & \textbf{0.79\std{0.09}} & 0.73\std{0.04} & 0.57\std{0.02} & 0.74\std{0.06} & 0.78\std{0.05} & 0.51\std{0.03} \\
        Lsun & 0.89\std{0.02} & \textbf{0.91\std{0.04}} & 0.78\std{0.20} & 0.86\std{0.04} & 0.90\std{0.05} & 0.58\std{0.10} \\
        Mice & \textbf{0.23\std{0.02}} & 0.20\std{0.03} & X & 0.17\std{0.03} & 0.21\std{0.03} & 0.12\std{0.03} \\
        Target & \textbf{0.64\std{0.01}} & 0.56\std{0.05} & X & 0.63\std{0.02} & \textbf{0.64\std{0.02}} & 0.28\std{0.02} \\
        Wine & 0.67\std{0.08} & 0.73\std{0.04} & 0.48\std{0.16} & \textbf{0.75\std{0.03}} & \textbf{0.75\std{0.04}} & 0.29\std{0.09} \\
        \bottomrule
\end{tabular}
\end{table*}

%% file: figs_tex/tab_benchmark_small_wad.tex
\begin{table*}
    \centering
    \caption{WAD scores \std{std} (lower is better) after 30 runs on random subsamples of 80\% of the input datasets. All models are limited to finding as many leaves as clusters.}
    \label{tab:benchmark_small_wad}
    \begin{tabular}{cccccc}
        \toprule
        Dataset & Kauri & KMeans+DT & IMM & ExShallow & RDM \\
        \midrule
        Digits & \textbf{4.47\std{0.08}} & 4.60\std{0.21} & 6.55\std{0.36} & 4.88\std{0.21} & 4.48\std{0.17} \\
        Hepta & 4.80\std{0.06} & \textbf{4.75\std{0.05}} & 4.88\std{0.07} & 4.77\std{0.06} & 4.91\std{0.06} \\
        Iris & \textbf{2.67\std{0.03}} & \textbf{2.67\std{0.02}} & \textbf{2.67\std{0.01}} & \textbf{2.67\std{0.02}} & 2.68\std{0.02} \\
        Lsun & \textbf{2.48\std{0.01}} & \textbf{2.48\std{0.02}} & 2.71\std{0.11} & 2.58\std{0.13} & 2.59\std{0.09} \\
        Mice & \textbf{4.03\std{0.04}} & 4.14\std{0.10} & 5.61\std{0.59} & 4.16\std{0.11} & 4.38\std{0.31} \\
        Target & 4.20\std{0.02} & \textbf{3.74\std{0.21}} & 4.30\std{0.07} & 4.30\std{0.02} & 4.03\std{0.02} \\
        Wine & \textbf{2.62\std{0.05}} & 2.65\std{0.04} & 2.70\std{0.02} & 2.70\std{0.02} & 2.63\std{0.10} \\
        \bottomrule
    \end{tabular}
\end{table*}

%% file: figs_tex/tab_benchmark_large_ari.tex
\begin{table}[!th]
\centering
    \caption{Performances after 30 runs on random subsamples of 80\% of the input datasets. All models are limited to finding 4 times more leaves than clusters.}
    \subfloat[ARI scores \std{std} (greater is better)]{
    \begin{tabular}{cccc}
        \toprule
        Method & Kauri & KMeans+DT & ExKMC \\
        \midrule
        Atom & \textbf{0.18\std{0.03}} & 0.17\std{0.02} & 0.17\std{0.02} \\
        Cancer & 0.86\std{0.01} & 0.84\std{0.02} & \textbf{0.87\std{0.01}} \\
        Chainlink & 0.10\std{0.01} & 0.09\std{0.01} & \textbf{0.11\std{0.01}} \\
        Digits & 0.55\std{0.03} & \textbf{0.58\std{0.02}} & \textbf{0.58\std{0.02}} \\
        Hepta & \textbf{1.00\std{0.00}} & \textbf{1.00\std{0.00}} & \textbf{1.00\std{0.00}} \\
        Iris & \textbf{0.72\std{0.03}} & \textbf{0.72\std{0.03}} & \textbf{0.72\std{0.04}} \\
        Lsun & \textbf{0.88\std{0.03}} & \textbf{0.88\std{0.03}} & 0.87\std{0.03} \\
        Mice & \textbf{0.22\std{0.01}} & 0.20\std{0.02} & 0.20\std{0.02} \\
        Target & 0.63\std{0.02} & 0.63\std{0.01} & \textbf{0.64\std{0.02}} \\
        Wine & 0.85\std{0.04} & \textbf{0.87\std{0.04}} & 0.85\std{0.04} \\
        \bottomrule
    \end{tabular}
    \label{tab:benchmark_large_ari}}
    \subfloat[WAES scores \std{std} (lower is better)]{
    \begin{tabular}{cccc}
        \toprule
        Dataset & Kauri & KMeans+DT & ExKMC \\
        \midrule
        Atom & \textbf{2.37\std{0.41}} & 2.89\std{0.44} & 3.18\std{0.72} \\
        Cancer & \textbf{3.45\std{0.16}} & 3.88\std{0.13} & 4.02\std{0.12} \\
        Chainlink & \textbf{2.59\std{0.32}} & 2.78\std{0.31} & 3.26\std{0.39} \\
        Digits & \textbf{6.59\std{0.16}} & \textbf{6.59\std{0.15}} & 8.59\std{0.35} \\
        Hepta & 4.39\std{0.05} & \textbf{4.37\std{0.05}} & 4.46\std{0.05} \\
        Iris & \textbf{3.32\std{0.35}} & 3.48\std{0.20} & 3.65\std{0.35} \\
        Lsun & \textbf{2.82\std{0.27}} & 2.86\std{0.27} & 3.39\std{0.46} \\
        Mice & 6.54\std{0.29} & \textbf{6.39\std{0.31}} & 8.00\std{0.45} \\
        Target & \textbf{4.17\std{0.04}} & 4.22\std{0.05} & 4.18\std{0.07} \\
        Wine & \textbf{3.68\std{0.39}} & 3.86\std{0.30} & 4.36\std{0.52} \\
        \bottomrule
    \end{tabular}\label{tab:benchmark_large_waes}
    }
    \end{table}

%% file: figs_tex/fig_synthetic.tex
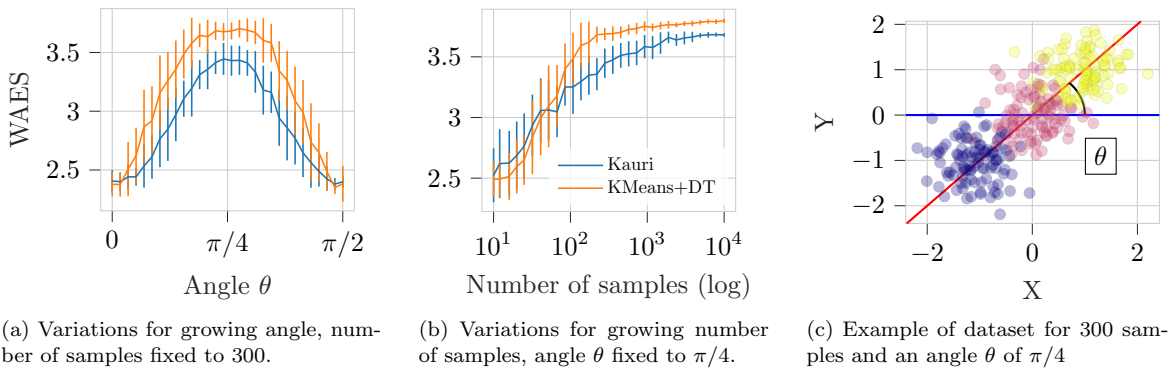
\begin{figure}[!th]
    \centering
    \subfloat[][Variations for growing angle, number of samples fixed to 300.]{\input{figs/synthetic_angle}}\hfil
    \subfloat[][Variations for growing number of samples, angle $\theta$ fixed to $\pi/4$.]{\input{figs/synthetic_samples}}\hfil
    \subfloat[][Example of dataset for 300 samples and an angle $\theta$ of $\pi/4$]{\input{figs/synthetic_example}\label{sfig:synthetic_dataset_rotation}}\hfil
    \caption{Variations of WAES scores for aligned isotropic 2d Gaussian distributions separated by Kauri or KMeans+Tree as the angle of the alignment (red line in \ref{sfig:synthetic_dataset_rotation}) with the x-axis (blue line in \ref{sfig:synthetic_dataset_rotation}) $\theta$ grows or the number of samples increases over 30 runs. The distance between the means is $\sqrt{2}$ and the scale matrices are $0.2\vec{I}_2$.}
    \label{fig:synthetic}
\end{figure}

%% file: figs/synthetic_angle.tex
\begin{tikzpicture}

\definecolor{darkorange25512714}{RGB}{255,127,14}
\definecolor{darkslategray38}{RGB}{38,38,38}
\definecolor{lightgray204}{RGB}{204,204,204}
\definecolor{steelblue31119180}{RGB}{31,119,180}

\begin{axis}[
axis line style={lightgray204},
legend cell align={left},
legend style={
  fill opacity=0.8,
  draw opacity=1,
  text opacity=1,
  at={(0.5,0.09)},
  anchor=south,
  draw=none,
  nodes={scale=0.7, transform shape}
},
tick align=outside,
tick pos=left,
x grid style={lightgray204},
xlabel=\textcolor{darkslategray38}{Angle $\theta$},
xmajorgrids,
xmin=-0.0785398163397448, xmax=1.64933614313464,
xtick style={color=darkslategray38},
xtick={0,0.785398163397448,1.5707963267949},
xticklabels={0,\(\displaystyle \pi/4\),\(\displaystyle \pi/2\)},
y grid style={lightgray204},
ylabel=\textcolor{darkslategray38}{WAES},
ymajorgrids,
ymin=2.15218010730024, ymax=3.87347197561062,
ytick style={color=darkslategray38},
width=0.3\linewidth
]
\path [draw=steelblue31119180, semithick]
(axis cs:0,2.31501693924404)
--(axis cs:0,2.49720528297819);

\path [draw=steelblue31119180, semithick]
(axis cs:0.0541653905791343,2.32368343164957)
--(axis cs:0.0541653905791343,2.47676101279487);

\path [draw=steelblue31119180, semithick]
(axis cs:0.108330781158269,2.33694810834063)
--(axis cs:0.108330781158269,2.54460744721493);

\path [draw=steelblue31119180, semithick]
(axis cs:0.162496171737403,2.36682220929204)
--(axis cs:0.162496171737403,2.51695556848574);

\path [draw=steelblue31119180, semithick]
(axis cs:0.216661562316537,2.34929271370843)
--(axis cs:0.216661562316537,2.71870728629157);

\path [draw=steelblue31119180, semithick]
(axis cs:0.270826952895672,2.42531687404342)
--(axis cs:0.270826952895672,2.79801645928992);

\path [draw=steelblue31119180, semithick]
(axis cs:0.324992343474806,2.52369715970123)
--(axis cs:0.324992343474806,3.00719172918766);

\path [draw=steelblue31119180, semithick]
(axis cs:0.37915773405394,2.5768658344926)
--(axis cs:0.37915773405394,3.11802305439629);

\path [draw=steelblue31119180, semithick]
(axis cs:0.433323124633075,2.74521976911353)
--(axis cs:0.433323124633075,3.19833578644203);

\path [draw=steelblue31119180, semithick]
(axis cs:0.487488515212209,2.9528142138093)
--(axis cs:0.487488515212209,3.26007467507958);

\path [draw=steelblue31119180, semithick]
(axis cs:0.541653905791344,2.95490650980637)
--(axis cs:0.541653905791344,3.43576015686029);

\path [draw=steelblue31119180, semithick]
(axis cs:0.595819296370478,3.1870998233622)
--(axis cs:0.595819296370478,3.43512239886003);

\path [draw=steelblue31119180, semithick]
(axis cs:0.649984686949612,3.18699205902104)
--(axis cs:0.649984686949612,3.52656349653452);

\path [draw=steelblue31119180, semithick]
(axis cs:0.704150077528747,3.3167828439934)
--(axis cs:0.704150077528747,3.51055048933994);

\path [draw=steelblue31119180, semithick]
(axis cs:0.758315468107881,3.30943887016046)
--(axis cs:0.758315468107881,3.58056112983954);

\path [draw=steelblue31119180, semithick]
(axis cs:0.812480858687015,3.30177380923077)
--(axis cs:0.812480858687015,3.55933730188034);

\path [draw=steelblue31119180, semithick]
(axis cs:0.86664624926615,3.30993421612048)
--(axis cs:0.86664624926615,3.55695467276841);

\path [draw=steelblue31119180, semithick]
(axis cs:0.920811639845284,3.30158162743387)
--(axis cs:0.920811639845284,3.52086281701057);

\path [draw=steelblue31119180, semithick]
(axis cs:0.974977030424419,3.15796232840226)
--(axis cs:0.974977030424419,3.51203767159774);

\path [draw=steelblue31119180, semithick]
(axis cs:1.02914242100355,2.97215128338227)
--(axis cs:1.02914242100355,3.38740427217329);

\path [draw=steelblue31119180, semithick]
(axis cs:1.08330781158269,2.98545982634907)
--(axis cs:1.08330781158269,3.33787350698427);

\path [draw=steelblue31119180, semithick]
(axis cs:1.13747320216182,2.7053182029657)
--(axis cs:1.13747320216182,3.17734846370097);

\path [draw=steelblue31119180, semithick]
(axis cs:1.19163859274096,2.58166576416326)
--(axis cs:1.19163859274096,3.13544534694785);

\path [draw=steelblue31119180, semithick]
(axis cs:1.24580398332009,2.5471208764882)
--(axis cs:1.24580398332009,2.99421245684513);

\path [draw=steelblue31119180, semithick]
(axis cs:1.29996937389922,2.43915528201235)
--(axis cs:1.29996937389922,2.85306694020987);

\path [draw=steelblue31119180, semithick]
(axis cs:1.35413476447836,2.34843774106307)
--(axis cs:1.35413476447836,2.77578448115915);

\path [draw=steelblue31119180, semithick]
(axis cs:1.40830015505749,2.35889646009594)
--(axis cs:1.40830015505749,2.6184368732374);

\path [draw=steelblue31119180, semithick]
(axis cs:1.46246554563663,2.32287944851239)
--(axis cs:1.46246554563663,2.52689832926538);

\path [draw=steelblue31119180, semithick]
(axis cs:1.51663093621576,2.32885000105305)
--(axis cs:1.51663093621576,2.43092777672473);

\path [draw=steelblue31119180, semithick]
(axis cs:1.5707963267949,2.34869638759485)
--(axis cs:1.5707963267949,2.45063694573848);

\path [draw=darkorange25512714, semithick]
(axis cs:0,2.27571726934349)
--(axis cs:0,2.47983828621206);

\path [draw=darkorange25512714, semithick]
(axis cs:0.0541653905791343,2.27290759799728)
--(axis cs:0.0541653905791343,2.48020351311383);

\path [draw=darkorange25512714, semithick]
(axis cs:0.108330781158269,2.27636104228154)
--(axis cs:0.108330781158269,2.73830562438513);

\path [draw=darkorange25512714, semithick]
(axis cs:0.162496171737403,2.38592738671579)
--(axis cs:0.162496171737403,2.85985039106199);

\path [draw=darkorange25512714, semithick]
(axis cs:0.216661562316537,2.55345153427281)
--(axis cs:0.216661562316537,3.18188179906052);

\path [draw=darkorange25512714, semithick]
(axis cs:0.270826952895672,2.62529106905532)
--(axis cs:0.270826952895672,3.2129311531669);

\path [draw=darkorange25512714, semithick]
(axis cs:0.324992343474806,2.84073232343915)
--(axis cs:0.324992343474806,3.47437878767196);

\path [draw=darkorange25512714, semithick]
(axis cs:0.37915773405394,3.01752868795515)
--(axis cs:0.37915773405394,3.50247131204485);

\path [draw=darkorange25512714, semithick]
(axis cs:0.433323124633075,3.11319615149664)
--(axis cs:0.433323124633075,3.60169273739225);

\path [draw=darkorange25512714, semithick]
(axis cs:0.487488515212209,3.27770278671841)
--(axis cs:0.487488515212209,3.69496387994826);

\path [draw=darkorange25512714, semithick]
(axis cs:0.541653905791344,3.36299078608025)
--(axis cs:0.541653905791344,3.79523143614197);

\path [draw=darkorange25512714, semithick]
(axis cs:0.595819296370478,3.54554534248855)
--(axis cs:0.595819296370478,3.74667687973367);

\path [draw=darkorange25512714, semithick]
(axis cs:0.649984686949612,3.48969849928313)
--(axis cs:0.649984686949612,3.77807927849465);

\path [draw=darkorange25512714, semithick]
(axis cs:0.704150077528747,3.60709844428199)
--(axis cs:0.704150077528747,3.76201266682912);

\path [draw=darkorange25512714, semithick]
(axis cs:0.758315468107881,3.60163049795083)
--(axis cs:0.758315468107881,3.75192505760472);

\path [draw=darkorange25512714, semithick]
(axis cs:0.812480858687015,3.61076635253374)
--(axis cs:0.812480858687015,3.75456698079959);

\path [draw=darkorange25512714, semithick]
(axis cs:0.86664624926615,3.62844839068008)
--(axis cs:0.86664624926615,3.77688494265326);

\path [draw=darkorange25512714, semithick]
(axis cs:0.920811639845284,3.59096855775284)
--(axis cs:0.920811639845284,3.7894758866916);

\path [draw=darkorange25512714, semithick]
(axis cs:0.974977030424419,3.56988920639842)
--(axis cs:0.974977030424419,3.76966634915714);

\path [draw=darkorange25512714, semithick]
(axis cs:1.02914242100355,3.41377372129099)
--(axis cs:1.02914242100355,3.79200405648679);

\path [draw=darkorange25512714, semithick]
(axis cs:1.08330781158269,3.42099342352516)
--(axis cs:1.08330781158269,3.74433990980817);

\path [draw=darkorange25512714, semithick]
(axis cs:1.13747320216182,3.09185143172338)
--(axis cs:1.13747320216182,3.59725967938773);

\path [draw=darkorange25512714, semithick]
(axis cs:1.19163859274096,3.01239881804339)
--(axis cs:1.19163859274096,3.51382340417883);

\path [draw=darkorange25512714, semithick]
(axis cs:1.24580398332009,2.79643425008055)
--(axis cs:1.24580398332009,3.38112130547501);

\path [draw=darkorange25512714, semithick]
(axis cs:1.29996937389922,2.71738790537423)
--(axis cs:1.29996937389922,3.26483431684799);

\path [draw=darkorange25512714, semithick]
(axis cs:1.35413476447836,2.4047924395316)
--(axis cs:1.35413476447836,3.06809644935729);

\path [draw=darkorange25512714, semithick]
(axis cs:1.40830015505749,2.39892577625974)
--(axis cs:1.40830015505749,2.89796311262915);

\path [draw=darkorange25512714, semithick]
(axis cs:1.46246554563663,2.32165150325453)
--(axis cs:1.46246554563663,2.6536818300788);

\path [draw=darkorange25512714, semithick]
(axis cs:1.51663093621576,2.29428017387449)
--(axis cs:1.51663093621576,2.40749760390328);

\path [draw=darkorange25512714, semithick]
(axis cs:1.5707963267949,2.23042064676889)
--(axis cs:1.5707963267949,2.53491268656444);

\addplot [semithick, steelblue31119180]
table {%
0 2.40611100196838
0.0541653633117676 2.40022230148315
0.108330726623535 2.44077777862549
0.162496209144592 2.4418888092041
0.21666157245636 2.53399991989136
0.270826935768127 2.61166667938232
0.324992299079895 2.76544451713562
0.379157781600952 2.84744453430176
0.43332314491272 2.97177767753601
0.487488508224487 3.10644435882568
0.541653871536255 3.19533324241638
0.595819234848022 3.31111121177673
0.64998459815979 3.35677766799927
0.704150080680847 3.41366672515869
0.758315443992615 3.4449999332428
0.812480926513672 3.43055558204651
0.866646289825439 3.43344449996948
0.920811653137207 3.41122221946716
0.974977016448975 3.33500003814697
1.02914237976074 3.17977786064148
1.0833078622818 3.16166663169861
1.13747322559357 2.94133329391479
1.19163858890533 2.85855555534363
1.2458039522171 2.77066659927368
1.29996931552887 2.64611101150513
1.35413479804993 2.56211113929749
1.40830016136169 2.48866677284241
1.46246552467346 2.42488884925842
1.51663088798523 2.37988877296448
1.57079637050629 2.39966654777527
};
\addplot [semithick, darkorange25512714]
table {%
0 2.37777781486511
0.0541653633117676 2.37655544281006
0.108330726623535 2.50733327865601
0.162496209144592 2.62288880348206
0.21666157245636 2.86766672134399
0.270826935768127 2.91911101341248
0.324992299079895 3.15755558013916
0.379157781600952 3.25999999046326
0.43332314491272 3.35744452476501
0.487488508224487 3.48633337020874
0.541653871536255 3.57911109924316
0.595819234848022 3.64611101150513
0.64998459815979 3.63388895988464
0.704150080680847 3.6845555305481
0.758315443992615 3.67677783966064
0.812480926513672 3.68266677856445
0.866646289825439 3.70266675949097
0.920811653137207 3.69022226333618
0.974977016448975 3.66977787017822
1.02914237976074 3.60288882255554
1.0833078622818 3.58266663551331
1.13747322559357 3.34455561637878
1.19163858890533 3.26311111450195
1.2458039522171 3.08877778053284
1.29996931552887 2.99111104011536
1.35413479804993 2.7364444732666
1.40830016136169 2.64844441413879
1.46246552467346 2.48766660690308
1.51663088798523 2.35088896751404
1.57079637050629 2.38266658782959
};
\end{axis}

\end{tikzpicture}

%% file: figs/synthetic_samples.tex
\begin{tikzpicture}

\definecolor{darkorange25512714}{RGB}{255,127,14}
\definecolor{darkslategray38}{RGB}{38,38,38}
\definecolor{lightgray204}{RGB}{204,204,204}
\definecolor{steelblue31119180}{RGB}{31,119,180}

\begin{axis}[
axis line style={lightgray204},
legend cell align={left},
legend style={
  fill opacity=0.8,
  draw opacity=1,
  text opacity=1,
  at={(0.97,0.03)},
  anchor=south east,
  draw=none,
  nodes={scale=0.7, transform shape}
},
log basis x={10},
tick align=outside,
tick pos=left,
x grid style={lightgray204},
xlabel=\textcolor{darkslategray38}{Number of samples (log)},
xmajorgrids,
xmin=7.07945784384138, xmax=14125.3754462276,
xmode=log,
xtick style={color=darkslategray38},
xtick={0.1,1,10,100,1000,10000,100000,1000000},
xticklabels={
  \(\displaystyle {10^{-1}}\),
  \(\displaystyle {10^{0}}\),
  \(\displaystyle {10^{1}}\),
  \(\displaystyle {10^{2}}\),
  \(\displaystyle {10^{3}}\),
  \(\displaystyle {10^{4}}\),
  \(\displaystyle {10^{5}}\),
  \(\displaystyle {10^{6}}\)
},
y grid style={lightgray204},
ymajorgrids,
ymin=2.22521907478369, ymax=3.89216836088228,
ytick style={color=darkslategray38},
width=0.3\linewidth
]
\path [draw=steelblue31119180, semithick]
(axis cs:10,2.30098949687908)
--(axis cs:10,2.74567716978759);

\path [draw=steelblue31119180, semithick]
(axis cs:12,2.3352138390603)
--(axis cs:12,2.90367504982859);

\path [draw=steelblue31119180, semithick]
(axis cs:16,2.36034342356881)
--(axis cs:16,2.88965657643119);

\path [draw=steelblue31119180, semithick]
(axis cs:20,2.40598604889016)
--(axis cs:20,2.96734728444317);

\path [draw=steelblue31119180, semithick]
(axis cs:25,2.49263563113471)
--(axis cs:25,3.03536436886529);

\path [draw=steelblue31119180, semithick]
(axis cs:32,2.68168199044061)
--(axis cs:32,3.18915134289272);

\path [draw=steelblue31119180, semithick]
(axis cs:41,2.79328274111616)
--(axis cs:41,3.32216441335539);

\path [draw=steelblue31119180, semithick]
(axis cs:52,2.81789763965552)
--(axis cs:52,3.30646133470345);

\path [draw=steelblue31119180, semithick]
(axis cs:67,2.83771048661457)
--(axis cs:67,3.25582185169389);

\path [draw=steelblue31119180, semithick]
(axis cs:85,3.09823177462369)
--(axis cs:85,3.40372900969003);

\path [draw=steelblue31119180, semithick]
(axis cs:108,3.02849658912868)
--(axis cs:108,3.47458983062441);

\path [draw=steelblue31119180, semithick]
(axis cs:137,3.14178263661727)
--(axis cs:137,3.4514047113146);

\path [draw=steelblue31119180, semithick]
(axis cs:174,3.18756454216381)
--(axis cs:174,3.50898718197412);

\path [draw=steelblue31119180, semithick]
(axis cs:221,3.223938760215)
--(axis cs:221,3.49008838910627);

\path [draw=steelblue31119180, semithick]
(axis cs:280,3.32644307371341)
--(axis cs:280,3.5678426405723);

\path [draw=steelblue31119180, semithick]
(axis cs:356,3.35969192600062)
--(axis cs:356,3.59012080808177);

\path [draw=steelblue31119180, semithick]
(axis cs:452,3.4100077586705)
--(axis cs:452,3.61403353926461);

\path [draw=steelblue31119180, semithick]
(axis cs:573,3.42069733691116)
--(axis cs:573,3.62933174976714);

\path [draw=steelblue31119180, semithick]
(axis cs:727,3.43163930739673)
--(axis cs:727,3.63310163712413);

\path [draw=steelblue31119180, semithick]
(axis cs:923,3.48868825195324)
--(axis cs:923,3.67974800662386);

\path [draw=steelblue31119180, semithick]
(axis cs:1172,3.48129018200766)
--(axis cs:1172,3.67297602959644);

\path [draw=steelblue31119180, semithick]
(axis cs:1487,3.52508256520614)
--(axis cs:1487,3.70226556301623);

\path [draw=steelblue31119180, semithick]
(axis cs:1887,3.61254029845782)
--(axis cs:1887,3.7075798216623);

\path [draw=steelblue31119180, semithick]
(axis cs:2395,3.59063028919832)
--(axis cs:2395,3.68744904274323);

\path [draw=steelblue31119180, semithick]
(axis cs:3039,3.60247295462599)
--(axis cs:3039,3.70894527505483);

\path [draw=steelblue31119180, semithick]
(axis cs:3856,3.62121567875697)
--(axis cs:3856,3.70622208057913);

\path [draw=steelblue31119180, semithick]
(axis cs:4893,3.65641935233598)
--(axis cs:4893,3.69608422420193);

\path [draw=steelblue31119180, semithick]
(axis cs:6210,3.65847477638698)
--(axis cs:6210,3.705282604182);

\path [draw=steelblue31119180, semithick]
(axis cs:7880,3.66596737718)
--(axis cs:7880,3.69965022011273);

\path [draw=steelblue31119180, semithick]
(axis cs:10000,3.66485498306198)
--(axis cs:10000,3.69617835027135);

\path [draw=darkorange25512714, semithick]
(axis cs:10,2.34296024464393)
--(axis cs:10,2.63703975535607);

\path [draw=darkorange25512714, semithick]
(axis cs:12,2.34107714822287)
--(axis cs:12,2.64225618511046);

\path [draw=darkorange25512714, semithick]
(axis cs:16,2.31990512478123)
--(axis cs:16,2.70926154188544);

\path [draw=darkorange25512714, semithick]
(axis cs:20,2.37358215503495)
--(axis cs:20,2.81641784496505);

\path [draw=darkorange25512714, semithick]
(axis cs:25,2.44437861447668)
--(axis cs:25,2.85428805218999);

\path [draw=darkorange25512714, semithick]
(axis cs:32,2.53362471790226)
--(axis cs:32,3.13929194876441);

\path [draw=darkorange25512714, semithick]
(axis cs:41,2.68738758944518)
--(axis cs:41,3.30610834551417);

\path [draw=darkorange25512714, semithick]
(axis cs:52,2.75406666095322)
--(axis cs:52,3.42926667238011);

\path [draw=darkorange25512714, semithick]
(axis cs:67,2.9261134642535)
--(axis cs:67,3.433090515846);

\path [draw=darkorange25512714, semithick]
(axis cs:85,3.18835592651696)
--(axis cs:85,3.61007544603205);

\path [draw=darkorange25512714, semithick]
(axis cs:108,3.26807711006337)
--(axis cs:108,3.70044140845515);

\path [draw=darkorange25512714, semithick]
(axis cs:137,3.4073239438911)
--(axis cs:137,3.78051060598724);

\path [draw=darkorange25512714, semithick]
(axis cs:174,3.44854993773205)
--(axis cs:174,3.79053052203806);

\path [draw=darkorange25512714, semithick]
(axis cs:221,3.61583101679478)
--(axis cs:221,3.74857320643298);

\path [draw=darkorange25512714, semithick]
(axis cs:280,3.63735299903429)
--(axis cs:280,3.73836128668);

\path [draw=darkorange25512714, semithick]
(axis cs:356,3.63204523924989)
--(axis cs:356,3.75147536000104);

\path [draw=darkorange25512714, semithick]
(axis cs:452,3.65264437858935)
--(axis cs:452,3.75178040017172);

\path [draw=darkorange25512714, semithick]
(axis cs:573,3.68380337528052)
--(axis cs:573,3.76715648510342);

\path [draw=darkorange25512714, semithick]
(axis cs:727,3.68654322255673)
--(axis cs:727,3.7802151451645);

\path [draw=darkorange25512714, semithick]
(axis cs:923,3.71177334378006)
--(axis cs:923,3.7931020625038);

\path [draw=darkorange25512714, semithick]
(axis cs:1172,3.70912701800859)
--(axis cs:1172,3.77540085457389);

\path [draw=darkorange25512714, semithick]
(axis cs:1487,3.72637447610372)
--(axis cs:1487,3.78857732842442);

\path [draw=darkorange25512714, semithick]
(axis cs:1887,3.73320478102833)
--(axis cs:1887,3.7959243481008);

\path [draw=darkorange25512714, semithick]
(axis cs:2395,3.74278794378223)
--(axis cs:2395,3.78350294000344);

\path [draw=darkorange25512714, semithick]
(axis cs:3039,3.75365940826153)
--(axis cs:3039,3.79419624601071);

\path [draw=darkorange25512714, semithick]
(axis cs:3856,3.75154376579472)
--(axis cs:3856,3.79961460211676);

\path [draw=darkorange25512714, semithick]
(axis cs:4893,3.76523638047164)
--(axis cs:4893,3.80448907766583);

\path [draw=darkorange25512714, semithick]
(axis cs:6210,3.7734856992649)
--(axis cs:6210,3.81029315205019);

\path [draw=darkorange25512714, semithick]
(axis cs:7880,3.76958747263573)
--(axis cs:7880,3.81034484546918);

\path [draw=darkorange25512714, semithick]
(axis cs:10000,3.77914206121311)
--(axis cs:10000,3.81639793878689);

\addplot [semithick, steelblue31119180]
table {%
10 2.52333331108093
11.9999990463257 2.61944437026978
16.0000019073486 2.625
20.0000019073486 2.68666672706604
25.0000019073486 2.76399993896484
31.9999980926514 2.93541669845581
41.0000038146973 3.05772352218628
51.9999961853027 3.06217956542969
67.0000076293945 3.04676628112793
85 3.25098037719727
107.999977111816 3.25154328346252
137.000030517578 3.29659366607666
174.000030517578 3.34827589988708
221.000015258789 3.357013463974
280.000030517578 3.44714283943176
356 3.47490644454956
452.000061035156 3.51202058792114
573 3.52501463890076
726.999938964844 3.53237056732178
923.000305175781 3.58421802520752
1171.99975585938 3.57713317871094
1487.00036621094 3.61367416381836
1886.99975585938 3.66006016731262
2395.00048828125 3.63903975486755
3038.99951171875 3.65570902824402
3856.00024414062 3.66371893882751
4893 3.67625188827515
6209.9990234375 3.68187880516052
7880.001953125 3.6828088760376
10000 3.68051671981812
};
\addlegendentry{Kauri}
\addplot [semithick, darkorange25512714]
table {%
10 2.49000000953674
11.9999990463257 2.49166655540466
16.0000019073486 2.51458334922791
20.0000019073486 2.59500002861023
25.0000019073486 2.64933323860168
31.9999980926514 2.83645844459534
41.0000038146973 2.99674797058105
51.9999961853027 3.09166669845581
67.0000076293945 3.1796019077301
85 3.39921569824219
107.999977111816 3.48425936698914
137.000030517578 3.59391736984253
174.000030517578 3.61954021453857
221.000015258789 3.68220210075378
280.000030517578 3.68785715103149
356 3.69176030158997
452.000061035156 3.7022123336792
573 3.7254798412323
726.999938964844 3.73337912559509
923.000305175781 3.75243759155273
1171.99975585938 3.74226403236389
1487.00036621094 3.75747585296631
1886.99975585938 3.76456451416016
2395.00048828125 3.76314544677734
3038.99951171875 3.77392792701721
3856.00024414062 3.77557921409607
4893 3.78486275672913
6209.9990234375 3.79188942909241
7880.001953125 3.78996610641479
10000 3.79777002334595
};
\addlegendentry{KMeans+DT}
\end{axis}

\end{tikzpicture}

%% file: figs/synthetic_example.tex
\begin{tikzpicture}

\definecolor{darkslategray38}{RGB}{38,38,38}
\definecolor{lightgray204}{RGB}{204,204,204}

\begin{axis}[
axis line style={lightgray204},
tick align=outside,
tick pos=left,
x grid style={lightgray204},
xlabel=\textcolor{darkslategray38}{X},
xmajorgrids,
xmin=-2.40861826267323, xmax=2.41818800875769,
xtick style={color=darkslategray38},
y grid style={lightgray204},
ylabel=\textcolor{darkslategray38}{Y},
ymajorgrids,
ymin=-2.39031661817807, ymax=2.06881179909104,
ytick style={color=darkslategray38},
width=0.3\linewidth
]
\addplot [
  colormap={mymap}{[1pt]
 rgb(0pt)=(0.050383,0.029803,0.527975);
  rgb(1pt)=(0.063536,0.028426,0.533124);
  rgb(2pt)=(0.075353,0.027206,0.538007);
  rgb(3pt)=(0.086222,0.026125,0.542658);
  rgb(4pt)=(0.096379,0.025165,0.547103);
  rgb(5pt)=(0.10598,0.024309,0.551368);
  rgb(6pt)=(0.115124,0.023556,0.555468);
  rgb(7pt)=(0.123903,0.022878,0.559423);
  rgb(8pt)=(0.132381,0.022258,0.56325);
  rgb(9pt)=(0.140603,0.021687,0.566959);
  rgb(10pt)=(0.148607,0.021154,0.570562);
  rgb(11pt)=(0.156421,0.020651,0.574065);
  rgb(12pt)=(0.16407,0.020171,0.577478);
  rgb(13pt)=(0.171574,0.019706,0.580806);
  rgb(14pt)=(0.17895,0.019252,0.584054);
  rgb(15pt)=(0.186213,0.018803,0.587228);
  rgb(16pt)=(0.193374,0.018354,0.59033);
  rgb(17pt)=(0.200445,0.017902,0.593364);
  rgb(18pt)=(0.207435,0.017442,0.596333);
  rgb(19pt)=(0.21435,0.016973,0.599239);
  rgb(20pt)=(0.221197,0.016497,0.602083);
  rgb(21pt)=(0.227983,0.016007,0.604867);
  rgb(22pt)=(0.234715,0.015502,0.607592);
  rgb(23pt)=(0.241396,0.014979,0.610259);
  rgb(24pt)=(0.248032,0.014439,0.612868);
  rgb(25pt)=(0.254627,0.013882,0.615419);
  rgb(26pt)=(0.261183,0.013308,0.617911);
  rgb(27pt)=(0.267703,0.012716,0.620346);
  rgb(28pt)=(0.274191,0.012109,0.622722);
  rgb(29pt)=(0.280648,0.011488,0.625038);
  rgb(30pt)=(0.287076,0.010855,0.627295);
  rgb(31pt)=(0.293478,0.010213,0.62949);
  rgb(32pt)=(0.299855,0.009561,0.631624);
  rgb(33pt)=(0.30621,0.008902,0.633694);
  rgb(34pt)=(0.312543,0.008239,0.6357);
  rgb(35pt)=(0.318856,0.007576,0.63764);
  rgb(36pt)=(0.32515,0.006915,0.639512);
  rgb(37pt)=(0.331426,0.006261,0.641316);
  rgb(38pt)=(0.337683,0.005618,0.643049);
  rgb(39pt)=(0.343925,0.004991,0.64471);
  rgb(40pt)=(0.35015,0.004382,0.646298);
  rgb(41pt)=(0.356359,0.003798,0.64781);
  rgb(42pt)=(0.362553,0.003243,0.649245);
  rgb(43pt)=(0.368733,0.002724,0.650601);
  rgb(44pt)=(0.374897,0.002245,0.651876);
  rgb(45pt)=(0.381047,0.001814,0.653068);
  rgb(46pt)=(0.387183,0.001434,0.654177);
  rgb(47pt)=(0.393304,0.001114,0.655199);
  rgb(48pt)=(0.399411,0.000859,0.656133);
  rgb(49pt)=(0.405503,0.000678,0.656977);
  rgb(50pt)=(0.41158,0.000577,0.65773);
  rgb(51pt)=(0.417642,0.000564,0.65839);
  rgb(52pt)=(0.423689,0.000646,0.658956);
  rgb(53pt)=(0.429719,0.000831,0.659425);
  rgb(54pt)=(0.435734,0.001127,0.659797);
  rgb(55pt)=(0.441732,0.00154,0.660069);
  rgb(56pt)=(0.447714,0.00208,0.66024);
  rgb(57pt)=(0.453677,0.002755,0.66031);
  rgb(58pt)=(0.459623,0.003574,0.660277);
  rgb(59pt)=(0.46555,0.004545,0.660139);
  rgb(60pt)=(0.471457,0.005678,0.659897);
  rgb(61pt)=(0.477344,0.00698,0.659549);
  rgb(62pt)=(0.48321,0.00846,0.659095);
  rgb(63pt)=(0.489055,0.010127,0.658534);
  rgb(64pt)=(0.494877,0.01199,0.657865);
  rgb(65pt)=(0.500678,0.014055,0.657088);
  rgb(66pt)=(0.506454,0.016333,0.656202);
  rgb(67pt)=(0.512206,0.018833,0.655209);
  rgb(68pt)=(0.517933,0.021563,0.654109);
  rgb(69pt)=(0.523633,0.024532,0.652901);
  rgb(70pt)=(0.529306,0.027747,0.651586);
  rgb(71pt)=(0.534952,0.031217,0.650165);
  rgb(72pt)=(0.54057,0.03495,0.64864);
  rgb(73pt)=(0.546157,0.038954,0.64701);
  rgb(74pt)=(0.551715,0.043136,0.645277);
  rgb(75pt)=(0.557243,0.047331,0.643443);
  rgb(76pt)=(0.562738,0.051545,0.641509);
  rgb(77pt)=(0.568201,0.055778,0.639477);
  rgb(78pt)=(0.573632,0.060028,0.637349);
  rgb(79pt)=(0.579029,0.064296,0.635126);
  rgb(80pt)=(0.584391,0.068579,0.632812);
  rgb(81pt)=(0.589719,0.072878,0.630408);
  rgb(82pt)=(0.595011,0.07719,0.627917);
  rgb(83pt)=(0.600266,0.081516,0.625342);
  rgb(84pt)=(0.605485,0.085854,0.622686);
  rgb(85pt)=(0.610667,0.090204,0.619951);
  rgb(86pt)=(0.615812,0.094564,0.61714);
  rgb(87pt)=(0.620919,0.098934,0.614257);
  rgb(88pt)=(0.625987,0.103312,0.611305);
  rgb(89pt)=(0.631017,0.107699,0.608287);
  rgb(90pt)=(0.636008,0.112092,0.605205);
  rgb(91pt)=(0.640959,0.116492,0.602065);
  rgb(92pt)=(0.645872,0.120898,0.598867);
  rgb(93pt)=(0.650746,0.125309,0.595617);
  rgb(94pt)=(0.65558,0.129725,0.592317);
  rgb(95pt)=(0.660374,0.134144,0.588971);
  rgb(96pt)=(0.665129,0.138566,0.585582);
  rgb(97pt)=(0.669845,0.142992,0.582154);
  rgb(98pt)=(0.674522,0.147419,0.578688);
  rgb(99pt)=(0.67916,0.151848,0.575189);
  rgb(100pt)=(0.683758,0.156278,0.57166);
  rgb(101pt)=(0.688318,0.160709,0.568103);
  rgb(102pt)=(0.69284,0.165141,0.564522);
  rgb(103pt)=(0.697324,0.169573,0.560919);
  rgb(104pt)=(0.701769,0.174005,0.557296);
  rgb(105pt)=(0.706178,0.178437,0.553657);
  rgb(106pt)=(0.710549,0.182868,0.550004);
  rgb(107pt)=(0.714883,0.187299,0.546338);
  rgb(108pt)=(0.719181,0.191729,0.542663);
  rgb(109pt)=(0.723444,0.196158,0.538981);
  rgb(110pt)=(0.72767,0.200586,0.535293);
  rgb(111pt)=(0.731862,0.205013,0.531601);
  rgb(112pt)=(0.736019,0.209439,0.527908);
  rgb(113pt)=(0.740143,0.213864,0.524216);
  rgb(114pt)=(0.744232,0.218288,0.520524);
  rgb(115pt)=(0.748289,0.222711,0.516834);
  rgb(116pt)=(0.752312,0.227133,0.513149);
  rgb(117pt)=(0.756304,0.231555,0.509468);
  rgb(118pt)=(0.760264,0.235976,0.505794);
  rgb(119pt)=(0.764193,0.240396,0.502126);
  rgb(120pt)=(0.76809,0.244817,0.498465);
  rgb(121pt)=(0.771958,0.249237,0.494813);
  rgb(122pt)=(0.775796,0.253658,0.491171);
  rgb(123pt)=(0.779604,0.258078,0.487539);
  rgb(124pt)=(0.783383,0.2625,0.483918);
  rgb(125pt)=(0.787133,0.266922,0.480307);
  rgb(126pt)=(0.790855,0.271345,0.476706);
  rgb(127pt)=(0.794549,0.27577,0.473117);
  rgb(128pt)=(0.798216,0.280197,0.469538);
  rgb(129pt)=(0.801855,0.284626,0.465971);
  rgb(130pt)=(0.805467,0.289057,0.462415);
  rgb(131pt)=(0.809052,0.293491,0.45887);
  rgb(132pt)=(0.812612,0.297928,0.455338);
  rgb(133pt)=(0.816144,0.302368,0.451816);
  rgb(134pt)=(0.819651,0.306812,0.448306);
  rgb(135pt)=(0.823132,0.311261,0.444806);
  rgb(136pt)=(0.826588,0.315714,0.441316);
  rgb(137pt)=(0.830018,0.320172,0.437836);
  rgb(138pt)=(0.833422,0.324635,0.434366);
  rgb(139pt)=(0.836801,0.329105,0.430905);
  rgb(140pt)=(0.840155,0.33358,0.427455);
  rgb(141pt)=(0.843484,0.338062,0.424013);
  rgb(142pt)=(0.846788,0.342551,0.420579);
  rgb(143pt)=(0.850066,0.347048,0.417153);
  rgb(144pt)=(0.853319,0.351553,0.413734);
  rgb(145pt)=(0.856547,0.356066,0.410322);
  rgb(146pt)=(0.85975,0.360588,0.406917);
  rgb(147pt)=(0.862927,0.365119,0.403519);
  rgb(148pt)=(0.866078,0.36966,0.400126);
  rgb(149pt)=(0.869203,0.374212,0.396738);
  rgb(150pt)=(0.872303,0.378774,0.393355);
  rgb(151pt)=(0.875376,0.383347,0.389976);
  rgb(152pt)=(0.878423,0.387932,0.3866);
  rgb(153pt)=(0.881443,0.392529,0.383229);
  rgb(154pt)=(0.884436,0.397139,0.37986);
  rgb(155pt)=(0.887402,0.401762,0.376494);
  rgb(156pt)=(0.89034,0.406398,0.37313);
  rgb(157pt)=(0.89325,0.411048,0.369768);
  rgb(158pt)=(0.896131,0.415712,0.366407);
  rgb(159pt)=(0.898984,0.420392,0.363047);
  rgb(160pt)=(0.901807,0.425087,0.359688);
  rgb(161pt)=(0.904601,0.429797,0.356329);
  rgb(162pt)=(0.907365,0.434524,0.35297);
  rgb(163pt)=(0.910098,0.439268,0.34961);
  rgb(164pt)=(0.9128,0.444029,0.346251);
  rgb(165pt)=(0.915471,0.448807,0.34289);
  rgb(166pt)=(0.918109,0.453603,0.339529);
  rgb(167pt)=(0.920714,0.458417,0.336166);
  rgb(168pt)=(0.923287,0.463251,0.332801);
  rgb(169pt)=(0.925825,0.468103,0.329435);
  rgb(170pt)=(0.928329,0.472975,0.326067);
  rgb(171pt)=(0.930798,0.477867,0.322697);
  rgb(172pt)=(0.933232,0.48278,0.319325);
  rgb(173pt)=(0.93563,0.487712,0.315952);
  rgb(174pt)=(0.93799,0.492667,0.312575);
  rgb(175pt)=(0.940313,0.497642,0.309197);
  rgb(176pt)=(0.942598,0.502639,0.305816);
  rgb(177pt)=(0.944844,0.507658,0.302433);
  rgb(178pt)=(0.947051,0.512699,0.299049);
  rgb(179pt)=(0.949217,0.517763,0.295662);
  rgb(180pt)=(0.951344,0.52285,0.292275);
  rgb(181pt)=(0.953428,0.52796,0.288883);
  rgb(182pt)=(0.95547,0.533093,0.28549);
  rgb(183pt)=(0.957469,0.53825,0.282096);
  rgb(184pt)=(0.959424,0.543431,0.278701);
  rgb(185pt)=(0.961336,0.548636,0.275305);
  rgb(186pt)=(0.963203,0.553865,0.271909);
  rgb(187pt)=(0.965024,0.559118,0.268513);
  rgb(188pt)=(0.966798,0.564396,0.265118);
  rgb(189pt)=(0.968526,0.5697,0.261721);
  rgb(190pt)=(0.970205,0.575028,0.258325);
  rgb(191pt)=(0.971835,0.580382,0.254931);
  rgb(192pt)=(0.973416,0.585761,0.25154);
  rgb(193pt)=(0.974947,0.591165,0.248151);
  rgb(194pt)=(0.976428,0.596595,0.244767);
  rgb(195pt)=(0.977856,0.602051,0.241387);
  rgb(196pt)=(0.979233,0.607532,0.238013);
  rgb(197pt)=(0.980556,0.613039,0.234646);
  rgb(198pt)=(0.981826,0.618572,0.231287);
  rgb(199pt)=(0.983041,0.624131,0.227937);
  rgb(200pt)=(0.984199,0.629718,0.224595);
  rgb(201pt)=(0.985301,0.63533,0.221265);
  rgb(202pt)=(0.986345,0.640969,0.217948);
  rgb(203pt)=(0.987332,0.646633,0.214648);
  rgb(204pt)=(0.98826,0.652325,0.211364);
  rgb(205pt)=(0.989128,0.658043,0.2081);
  rgb(206pt)=(0.989935,0.663787,0.204859);
  rgb(207pt)=(0.990681,0.669558,0.201642);
  rgb(208pt)=(0.991365,0.675355,0.198453);
  rgb(209pt)=(0.991985,0.681179,0.195295);
  rgb(210pt)=(0.992541,0.68703,0.19217);
  rgb(211pt)=(0.993032,0.692907,0.189084);
  rgb(212pt)=(0.993456,0.69881,0.186041);
  rgb(213pt)=(0.993814,0.704741,0.183043);
  rgb(214pt)=(0.994103,0.710698,0.180097);
  rgb(215pt)=(0.994324,0.716681,0.177208);
  rgb(216pt)=(0.994474,0.722691,0.174381);
  rgb(217pt)=(0.994553,0.728728,0.171622);
  rgb(218pt)=(0.994561,0.734791,0.168938);
  rgb(219pt)=(0.994495,0.74088,0.166335);
  rgb(220pt)=(0.994355,0.746995,0.163821);
  rgb(221pt)=(0.994141,0.753137,0.161404);
  rgb(222pt)=(0.993851,0.759304,0.159092);
  rgb(223pt)=(0.993482,0.765499,0.156891);
  rgb(224pt)=(0.993033,0.77172,0.154808);
  rgb(225pt)=(0.992505,0.777967,0.152855);
  rgb(226pt)=(0.991897,0.784239,0.151042);
  rgb(227pt)=(0.991209,0.790537,0.149377);
  rgb(228pt)=(0.990439,0.796859,0.14787);
  rgb(229pt)=(0.989587,0.803205,0.146529);
  rgb(230pt)=(0.988648,0.809579,0.145357);
  rgb(231pt)=(0.987621,0.815978,0.144363);
  rgb(232pt)=(0.986509,0.822401,0.143557);
  rgb(233pt)=(0.985314,0.828846,0.142945);
  rgb(234pt)=(0.984031,0.835315,0.142528);
  rgb(235pt)=(0.982653,0.841812,0.142303);
  rgb(236pt)=(0.98119,0.848329,0.142279);
  rgb(237pt)=(0.979644,0.854866,0.142453);
  rgb(238pt)=(0.977995,0.861432,0.142808);
  rgb(239pt)=(0.976265,0.868016,0.143351);
  rgb(240pt)=(0.974443,0.874622,0.144061);
  rgb(241pt)=(0.97253,0.88125,0.144923);
  rgb(242pt)=(0.970533,0.887896,0.145919);
  rgb(243pt)=(0.968443,0.894564,0.147014);
  rgb(244pt)=(0.966271,0.901249,0.14818);
  rgb(245pt)=(0.964021,0.90795,0.14937);
  rgb(246pt)=(0.961681,0.914672,0.15052);
  rgb(247pt)=(0.959276,0.921407,0.151566);
  rgb(248pt)=(0.956808,0.928152,0.152409);
  rgb(249pt)=(0.954287,0.934908,0.152921);
  rgb(250pt)=(0.951726,0.941671,0.152925);
  rgb(251pt)=(0.949151,0.948435,0.152178);
  rgb(252pt)=(0.946602,0.95519,0.150328);
  rgb(253pt)=(0.944152,0.961916,0.146861);
  rgb(254pt)=(0.941896,0.96859,0.140956);
  rgb(255pt)=(0.940015,0.975158,0.131326)
},
  only marks,
  scatter,
  scatter src=explicit,
  opacity=0.3
]
table [x=x, y=y, meta=colordata]{%
x  y  colordata
0.147596190201397 0.995472567979418 1.0
0.503826554111882 0.755142461593843 2.0
0.145286108948902 0.445924717327541 1.0
0.0136855518912866 -0.0311446606935464 1.0
0.0230650155938266 0.387857899462499 1.0
-0.379380471140132 -0.145643814095632 1.0
0.210384098149148 0.139283364677667 1.0
1.06593466886272 0.56286441777195 2.0
1.39327513776958 1.28417048718907 2.0
0.181754534193827 -0.0863912472669262 1.0
-0.339231584352806 1.39396725410199 2.0
-0.335271714566185 0.0146725179841787 1.0
-1.15506176867567 -0.516062291157299 1.0
1.1512590301764 1.29148544152463 2.0
-0.690734468706352 -0.689298737495081 0.0
-1.32449701250804 -1.61865916837727 0.0
-1.70791157212716 -0.727030043250435 0.0
1.78766215747572 1.33724183309464 2.0
0.720479253325951 0.825411079665826 2.0
1.05033725809727 0.706831350352174 2.0
1.03019445645379 1.34775514238721 2.0
0.984015363616924 1.15027080842046 2.0
0.483606064457534 -0.282359925906791 1.0
1.12736222780347 0.86164213431951 2.0
-1.29176598786433 -1.17483966459049 0.0
-0.0994947436668174 -0.384120660217161 1.0
-1.90814970216248 -0.0767307245875026 0.0
1.22303747619944 0.669986875324635 2.0
0.0267857165877605 -0.0950431949783286 1.0
-0.340827971059556 -0.397027346918401 1.0
-1.46745151608301 -0.458359360328161 0.0
2.19878772369265 0.910195702448995 2.0
-0.289428335077448 0.211195345955395 1.0
0.416091328829561 -0.0784038786347838 1.0
-1.0607985544475 -0.491566726037602 0.0
-0.383060495858423 -0.689418843784651 1.0
1.16026194814929 -0.180688734831173 1.0
-0.653706731177203 -0.305643555413219 1.0
1.15163190893756 0.736308534497098 2.0
0.599379012881145 1.24521979475377 2.0
-0.233257523135127 -0.529807949638536 1.0
0.429635148898665 0.59437497424642 1.0
0.617524840095813 1.56224572038044 2.0
-1.50393182143625 -1.32676902501051 0.0
1.36575893014642 1.10652636432296 2.0
1.04706124237642 0.95900866654203 2.0
-1.02756016485892 -1.04798837842629 0.0
-1.3218168659418 -1.36358151752752 0.0
-0.877232552638725 -1.39842933755157 0.0
0.0201845453206999 1.04631202162112 1.0
-0.123624526533815 -0.116086355355028 1.0
0.163000969965163 0.657995182144447 1.0
0.291401880848641 1.37765144669206 2.0
-0.654190755744736 -1.52944751774365 0.0
-2.18921797760819 -0.728845665489417 0.0
-1.7852581411237 -0.798335978000938 0.0
-0.354321366692985 -0.237741389510823 1.0
-0.522148069706387 -1.20276029548902 0.0
0.215161225521768 1.23402112180279 1.0
-1.19718659572488 -1.12537878901964 0.0
-1.16309591124303 -0.929919905454941 0.0
-0.741277320926648 -0.843629773105497 0.0
-0.423729781223509 0.109318436837037 1.0
-0.389762806035892 -1.30833104873709 0.0
-1.29171456614855 -1.23308294629033 0.0
0.386177731209603 -0.359896813842702 1.0
0.712462876879175 1.0671335704087 2.0
-0.687640467045058 -0.998313607133498 0.0
0.493041876613268 1.4457644637147 2.0
-1.00701325351481 -0.928030735238129 0.0
0.979824717250422 1.03538431064068 2.0
0.308769246431871 0.356282598016297 1.0
0.547520874013508 1.03799890606757 2.0
0.100883186843467 0.62124467129625 1.0
1.83480608242457 1.33875013536411 2.0
-0.712301025042076 -1.70240414175137 0.0
-1.09253013693945 -0.606372024045112 0.0
-1.7594160090477 -0.826802906131641 0.0
-2.00871898891205 -1.45727896196559 0.0
-0.982723892015402 -1.74090551762848 0.0
-1.44073380040355 -1.6582246256705 0.0
0.619507797017482 -0.135549659703537 1.0
-0.746300075149338 -1.0995833323297 0.0
1.1078696920872 1.50436017217545 2.0
0.129298620698502 0.184641444088751 1.0
-0.616324757151791 -0.489683731380353 0.0
-0.513227216878402 -0.16015348329449 1.0
-1.26772617959153 -1.49904430329891 0.0
-0.188864930929995 0.0468295376021475 1.0
0.307599617530327 1.2379062910252 2.0
-0.635735661579214 -0.97364767539513 0.0
0.8209585669837 0.341557169423474 2.0
-1.64690998400719 -0.642095876604403 0.0
0.196153326976811 0.524418332154794 2.0
-0.22510352799894 -0.693881798877801 0.0
-0.834162017875802 -0.936468029319002 0.0
0.0961355346926673 0.0434777622957696 1.0
-0.584312677059651 -0.73962124709244 0.0
0.643671250461179 1.6923165203187 2.0
-1.05818559878452 -0.957982838477648 0.0
0.917821939181181 1.36878047891929 2.0
-1.25460409039387 -0.879295102985845 0.0
1.06997830947933 0.903248031601604 2.0
1.19804680278186 1.09767014032111 2.0
-1.43429116655492 -0.859209466043233 0.0
0.44064738588206 0.342976902928825 1.0
0.180059611416743 -0.794201257668159 1.0
0.746511654911048 0.135053680566194 1.0
-1.03842940726144 -0.9131095565963 0.0
1.18062779168618 1.86612414376062 2.0
0.194782163904807 -0.267981243633626 1.0
1.09171752705889 1.51945370858441 2.0
1.44314805437749 0.916501146794945 2.0
-0.792881820240346 -0.875184579717242 0.0
0.626490474084185 0.576604816810728 2.0
-0.603028665194615 0.565717743192526 1.0
1.62737124014843 1.15019432538935 2.0
0.0938475395357585 -0.554824187833236 1.0
0.267274679379936 0.205574859878452 2.0
1.09639825620583 1.25402061339008 2.0
1.03694477671854 0.632598594816376 2.0
0.127149318293912 0.185772807602923 1.0
0.571945885408352 1.45346213277734 2.0
-0.0275653749315501 -0.640738789853826 1.0
0.0391452638302805 0.419820365538238 1.0
0.271508593667566 -0.468756056333073 1.0
-0.495277390548529 -0.778181633414135 0.0
-0.655016772102172 -0.539620960032793 0.0
0.266944693656126 -0.441528031970462 1.0
-0.614235336309881 -2.18762896284766 0.0
-0.466501693388458 -0.383297018380223 1.0
0.30297029256861 0.0231747827006419 1.0
-1.46060106118088 -1.15649943029925 0.0
-0.507937885032175 -0.419506926718044 0.0
0.961318956533149 -0.0403582764807779 1.0
0.327207819436903 -0.0292872917118143 1.0
0.15570601553496 0.296618035029857 1.0
-0.493999560228743 -0.013835114846799 1.0
0.706089980404738 -0.355758661211526 1.0
-0.253319603622056 -0.137603722455272 1.0
1.0170468879135 0.74109993713585 2.0
0.566803723130003 0.223382400935397 1.0
-0.0277510015002842 0.563116659368613 1.0
0.507067673255835 0.98902116398544 2.0
0.626193802741089 0.578682938822595 2.0
0.953841854317366 0.529803919541098 2.0
-0.872866266290761 -0.220654828837468 0.0
0.624504954401249 0.388102191388536 2.0
1.69822546302976 0.579498367035985 2.0
0.704864650027215 1.09526417334954 2.0
-0.58107603482491 -0.368387848558787 0.0
1.20607348807726 0.820673538581437 2.0
-2.00158138357041 -0.820444163983237 0.0
0.36875319885677 0.237550187150223 1.0
-1.57229439627104 -1.26181279378293 0.0
1.1590754481989 0.786551090930207 2.0
1.2085712863271 0.957786387819601 2.0
-0.662433332574049 -0.569105299050523 1.0
0.679209823782615 -0.52375888087248 1.0
-1.17201866267617 -1.19848953503259 0.0
0.0513810553853635 0.177642798760275 2.0
0.546219959209318 -0.0862414875059327 1.0
1.0387211753116 1.63471015201329 2.0
1.41777788416981 0.377699925312601 2.0
1.38863145108425 1.0823683656118 2.0
0.847196909470561 1.01086322704954 2.0
-1.60281057471977 -1.3405859731993 0.0
-0.203851292709375 0.637797583244269 1.0
0.196283550643627 0.641221474569019 2.0
-1.32603868313962 -0.912096858097954 0.0
0.729672551975189 0.168939036528353 1.0
-0.996141486303461 -0.764316553093382 0.0
-0.797062559277348 -1.81828478803691 0.0
-0.983450538050491 -0.656583603244783 0.0
1.00340664881158 1.39661794520862 2.0
-1.36028712706675 -1.50012429656635 0.0
-0.279757610861503 0.443697078309686 1.0
1.02019878482824 1.83152495924145 2.0
0.272686718744005 0.939705566891977 2.0
-1.24789480111919 -1.21047555928108 0.0
-1.09702279721354 -0.800814082855923 0.0
0.429346402171272 -0.186264052178947 1.0
-0.123799006809897 0.502625732725312 1.0
-1.09818173263976 -1.48479591478432 0.0
0.527083407820222 0.751355934849248 2.0
0.00840394476525241 -0.265545347347886 1.0
-0.89974023128503 0.263723469728361 1.0
-0.40086872651986 -0.877760438810692 1.0
1.64021656401071 1.6539220567057 2.0
-1.62876573092377 -1.26388180136156 0.0
0.389505950148021 -0.392640396164321 1.0
-0.948504305638846 -1.16955994482058 0.0
-1.7792053798639 -1.58282787787584 0.0
1.43323566316366 0.673374377296618 2.0
0.659061556089173 0.352908610419942 2.0
0.336280737099043 0.251776656278412 1.0
-0.424890468438179 -1.67395017207808 0.0
-0.697477563666024 -1.17083959853714 0.0
-1.10029164430966 -1.13517018869571 0.0
-0.0725462951971934 0.344099636468845 1.0
-0.918008217503925 -0.252731558430891 0.0
1.6846537611708 1.04134388417528 2.0
0.464698055172829 0.214257897755827 1.0
-0.791524036511956 -1.16557745311407 0.0
-1.20294733685128 -0.819654615055957 0.0
-1.41054421379689 -0.88708007555334 0.0
0.910399199836041 0.23345527358899 2.0
1.03789324056027 0.0523646299368612 1.0
0.238902026334225 0.142162537291741 1.0
0.389661564936677 0.990219479457179 2.0
0.878712390401741 0.410941526796263 2.0
-1.26862222602436 -0.305815746360194 0.0
-0.348962739216014 0.121245328484983 1.0
-0.858161455964628 -0.767430766951754 0.0
-0.89910474245807 -0.798882634715413 0.0
1.63592680258425 1.2040919006367 2.0
-1.16578387014645 -1.42289223989228 0.0
0.0588416672623635 -0.628585562485218 1.0
-0.156427346409494 0.90492416604994 1.0
1.34760882566596 0.869842300770101 2.0
-0.707713045357051 1.00336927014682 1.0
0.973005071849456 0.682012978162611 2.0
-1.24404981400787 -1.84281241176159 0.0
0.673143395770733 0.265807996300086 2.0
-0.901832342307194 -0.824216912037001 0.0
0.992943319577146 0.780091610379688 2.0
1.42672655326233 1.24334893117446 2.0
-1.26448331345873 -0.497144453645988 0.0
1.20853124692191 1.76689600939299 2.0
0.6278350915429 1.07707240610005 2.0
-0.0533422763602571 -1.72040672272008 0.0
-1.01599596802071 0.0647016884887948 0.0
0.224648018365008 -0.385617525240606 1.0
-1.67189217166 -1.79499683042792 0.0
-1.23823193093404 -0.512201889472923 0.0
0.092694295996034 0.137622191174848 1.0
0.0712189745464558 -0.875889720189297 1.0
-0.646859164630859 -0.20229719316265 1.0
-1.35270361296941 -0.510462836672775 0.0
-0.428049362891566 -0.603033688013999 1.0
1.03067850717015 0.878234425838198 2.0
-0.43423999144101 -0.905415660498677 0.0
0.269539058067114 -0.265935461579885 1.0
-1.31140674651567 -1.12986953157326 0.0
1.65105710849838 1.02481687766302 2.0
-0.363851191145008 -0.233285175235988 1.0
-0.0327000997661773 -0.580205820668912 1.0
-1.17301390241419 -1.22820985084908 0.0
-0.917746018371963 -1.17239626176599 0.0
-0.357732247494135 -0.0984208807759184 1.0
0.585254458057862 -0.0115374659957044 1.0
1.4714426929532 1.42522575635959 2.0
1.25012610842728 0.524300870220012 2.0
0.807207856894951 1.25614219809973 2.0
1.30301271714705 1.58459038733825 2.0
-0.747767715136978 -1.57446925179297 0.0
-0.434871792421878 0.602048366493557 1.0
-0.209069303063215 -0.385718729664521 1.0
1.20399783925044 0.840085795437929 2.0
-1.34243243147475 -0.751444513460492 0.0
-0.995371645821845 -0.677991113603468 0.0
-1.81583237821669 -0.864224206277723 0.0
-0.65444036367519 -1.74308934604593 0.0
0.0255522005054456 0.718411768659998 2.0
0.707842199565711 1.03511253949532 2.0
1.18684610101757 0.445358392997296 2.0
-0.982215775644281 -1.70078133877383 0.0
0.110357120831311 0.682428967383641 1.0
-0.676619628287725 -0.988993122974986 0.0
-0.560176771953006 -0.262210286588296 1.0
1.00833041340818 1.47150846285226 2.0
1.01366084633741 0.983505183556044 2.0
0.34472876860695 0.108748131921456 1.0
1.00783322039154 1.14472984178727 2.0
-1.16325765594767 -1.80110603347762 0.0
1.34660094014259 1.19357086897866 2.0
0.638330524779581 0.506220742432968 2.0
0.647102776750568 1.00055832640199 2.0
0.395867902995811 0.659087661899377 1.0
0.732498985480176 0.320217765968071 2.0
-0.763403776782327 -0.810994734524404 0.0
0.436605972590389 1.01279947858284 2.0
0.185187283195479 -0.226842147465515 1.0
-1.40132989512476 -0.413045408123725 0.0
-0.0175611794601431 -0.424375654228004 1.0
-0.966643944954236 -1.48169334753782 0.0
0.489326160611343 0.413716227041074 2.0
-0.209882141778112 0.391623013070759 1.0
-0.610519398766935 0.870768984848781 1.0
-0.214770770195992 -0.234004935739904 1.0
0.456705589205308 0.316938161201539 1.0
-1.75499942075066 -1.05029631643189 0.0
1.16599341182107 0.557337898710097 2.0
-0.547490307461244 -1.2942446523906 0.0
-0.200694709472687 -0.7800244778355 1.0
-1.29861020652742 -0.247821731272346 0.0
-1.38128798218556 -0.989732085142768 0.0
0.144148259302333 1.29675375850305 2.0
0.931096728733128 1.5337993713786 2.0
0.956100971797965 0.603494048938771 2.0
};
\addplot [thick, red]
table {%
-2.39031661817807 -2.39031661817807
2.06881179909104 2.06881179909104
};
\addplot [thick, blue]
table {%
-2.40861826267323 0
2.41818800875769 0
};
\draw [black,thick,domain=0:45] plot ({cos(\x)}, {sin(\x)});
\node[draw, anchor=north west, fill=white] at (1,-0.5) {$\theta$};
\end{axis}

\end{tikzpicture}

%% file: figs_tex/tab_kernels_ari.tex
\begin{table*}
\centering
\caption{ARI scores \std{std} (greater is better) after 30 runs on random subsamples of 80\% of the input datasets for varying kernels. All models are limited to finding 4 times more leaves than clusters.} 
\label{tab:benchmark_kernels_ari}
\resizebox{\linewidth}{!}{\begin{tabular}{ccccccccc}
\toprule
Kernel & \multicolumn{2}{c}{Additive $\chi^2$} & \multicolumn{2}{c}{$\chi^2$} & \multicolumn{2}{c}{Laplacian} & \multicolumn{2}{c}{RBF} \\
\cmidrule(lr){2-3}\cmidrule(lr){4-5}\cmidrule(lr){6-7}\cmidrule(lr){8-9}
Dataset & Kauri & KMeans+DT & Kauri & KMeans+DT & Kauri & KMeans+DT & Kauri & KMeans+DT \\
\midrule
Cancer & 0.86\std{0.01} & 0.84\std{0.02} & 0.82\std{0.02} & 0.86\std{0.02} & \textbf{0.87\std{0.01}} & 0.79\std{0.03} & \textbf{0.87\std{0.02}} & 0.78\std{0.03} \\
Iris & 0.67\std{0.04} & 0.62\std{0.09} & 0.67\std{0.04} & 0.61\std{0.11} & \textbf{0.78\std{0.05}} & 0.71\std{0.14} & 0.72\std{0.03} & 0.71\std{0.05} \\
Lsun & \textbf{0.98\std{0.00}} & 0.89\std{0.21} & \textbf{0.98\std{0.01}} & 0.81\std{0.26} & \textbf{0.98\std{0.01}} & 0.96\std{0.01} & 0.88\std{0.02} & 0.93\std{0.02} \\
Twodiamonds & 0.98\std{0.00} & 0.95\std{0.02} & 0.98\std{0.00} & 0.97\std{0.02} & \textbf{1.00\std{0.00}} & 0.77\std{0.43} & \textbf{1.00\std{0.00}} & \textbf{1.00\std{0.00}} \\
Wine & 0.87\std{0.03} & 0.74\std{0.03} & \textbf{0.90\std{0.03}} & 0.73\std{0.10} & 0.89\std{0.03} & 0.83\std{0.03} & 0.85\std{0.03} & 0.80\std{0.04} \\
\bottomrule
\end{tabular}}
\end{table*}

%% file: figs_tex/tab_kernels_kmeans.tex
\begin{table*}
\centering
\caption{Relative Kernel KMeans scores \std{std} (lower is better) of Kauri and Kernel-KMeans + Decision Tree after 30 runs on random subsamples of 80\% of the input datasets for varying kernels. All models are limited to finding 4 times more leaves than clusters.} 
\label{tab:benchmark_kernels_kmeans}
\resizebox{\linewidth}{!}{\begin{tabular}{ccccccccc}
\toprule
Kernel & \multicolumn{2}{c}{Additive $\chi^2$} & \multicolumn{2}{c}{$\chi^2$} & \multicolumn{2}{c}{Laplacian} & \multicolumn{2}{c}{RBF} \\
\cmidrule(lr){2-3}\cmidrule(lr){4-5}\cmidrule(lr){6-7}\cmidrule(lr){8-9}
Dataset & Kauri & KMeans+DT & Kauri & KMeans+DT & Kauri & KMeans+DT & Kauri & KMeans+DT \\
\midrule
Cancer & 1.04\std{0.02} & 1.05\std{0.02} & 1.02\std{0.01} & 1.03\std{0.01} & \textbf{1.01\std{0.02}} & 1.04\std{0.02} & 1.04\std{0.02} & 1.07\std{0.03} \\
Iris & 1.11\std{0.05} & 1.18\std{0.16} & 1.10\std{0.04} & 1.20\std{0.16} & \textbf{1.05\std{0.02}} & 1.09\std{0.08} & 1.06\std{0.05} & 1.06\std{0.09} \\
Lsun & 1.08\std{0.03} & 1.13\std{0.11} & \textbf{1.04\std{0.02}} & 1.11\std{0.12} & \textbf{1.04\std{0.01}} & \textbf{1.04\std{0.01}} & 1.05\std{0.03} & 1.07\std{0.04} \\
Twodiamonds & 1.03\std{0.02} & 1.03\std{0.02} & 1.05\std{0.02} & 1.05\std{0.02} & \textbf{1.01\std{0.01}} & 1.06\std{0.09} & 1.04\std{0.01} & 1.03\std{0.02} \\
Wine & 1.03\std{0.03} & 1.05\std{0.04} & 1.05\std{0.02} & 1.08\std{0.05} & \textbf{1.02\std{0.02}} & \textbf{1.02\std{0.02}} & 1.06\std{0.03} & 1.08\std{0.03} \\
\bottomrule
\end{tabular}}
\end{table*}

%% file: figs_tex/fig_wine_pca.tex
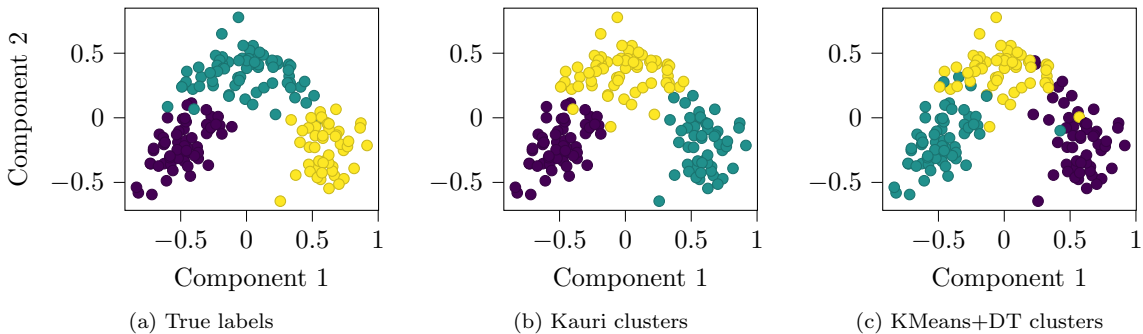
\begin{figure*}[!th]
    \centering
    \subfloat[True labels]{
    \input{figs/pca_wine_labels}
    \label{sfig:pca_wine_labels}
    }
    \subfloat[Kauri clusters]{
    \input{figs/pca_wine_kauri}
    \label{sfig:pca_wine_kauri}
    }
    \subfloat[KMeans+DT clusters]{
    \input{figs/pca_wine_tree}
    \label{sfig:pca_wine_tree}
    }
    \caption{PCA of the wine dataset with samples coloured according to clusters found by Kauri or KMeans+DT with a $\chi^2$ kernel.}
    \label{fig:pca_wine}
\end{figure*}

%% file: figs/pca_wine_labels.tex
\begin{tikzpicture}

\definecolor{darkgray176}{RGB}{176,176,176}

\begin{axis}[
tick align=outside,
tick pos=left,
x grid style={darkgray176},
xlabel={Component 1},
xmin=-0.923263684098084, xmax=1.00276311853182,
xtick style={color=black},
y grid style={darkgray176},
ylabel={Component 2},
ymin=-0.716448288694247, ymax=0.848685539548014,
ytick style={color=black},
width=0.3\linewidth
]
\addplot [colormap/viridis, only marks, scatter, scatter src=explicit]
table [x=x, y=y, meta=colordata]{%
x  y  colordata
-0.706335755973939 -0.253192752857043 0.0
-0.484976801771771 -0.00882289142063541 0.0
-0.521172266128946 -0.189187221677481 0.0
-0.821643663444924 -0.580905512257258 0.0
-0.202546381570955 -0.0594665740151188 0.0
-0.608190152120248 -0.487519191054382 0.0
-0.544047398643627 -0.300196496926384 0.0
-0.47435749548749 -0.298197020592497 0.0
-0.500432011957094 -0.307602858921239 0.0
-0.62751796867287 -0.206328232885875 0.0
-0.727467157091796 -0.356512043609109 0.0
-0.374967743832642 -0.225424534904278 0.0
-0.448188282848203 -0.231938139182621 0.0
-0.626345328675927 -0.355138676571354 0.0
-0.83571701125127 -0.538047802465009 0.0
-0.471931567976655 -0.337405385467885 0.0
-0.426990905027504 -0.450842684419911 0.0
-0.366595703552089 -0.315750340879042 0.0
-0.718788533345653 -0.59388133189827 0.0
-0.458884985639634 -0.175782239611429 0.0
-0.661852288038142 -0.127831031585363 0.0
-0.267900031953801 0.00981127565393747 0.0
-0.599782398827993 0.000782494522600041 0.0
-0.417894799937014 0.113786900972701 0.0
-0.449940391686938 0.0983771974086478 0.0
-0.215787795169716 0.00926270331364054 0.0
-0.415531995702532 -0.155470384900159 0.0
-0.283946970073552 -0.0897364262250318 0.0
-0.47509922696507 -0.088051647935278 0.0
-0.514396469295947 -0.104602501953416 0.0
-0.520965811951038 -0.258430884386146 0.0
-0.575738232641882 -0.361155787969285 0.0
-0.302144638806723 -0.0514943064152287 0.0
-0.367021971487879 -0.268616629871898 0.0
-0.303022468117356 -0.121413201641094 0.0
-0.458338877641686 -0.0303098821876442 0.0
-0.271566324740124 -0.0761238405816806 0.0
-0.232004188780535 -0.0503302762652722 0.0
-0.305480403391417 0.0616775562692842 0.0
-0.519605033037914 -0.313830061471668 0.0
-0.530575114358287 -0.129536622670152 0.0
-0.187044525570531 -0.124830305999712 0.0
-0.705508197447484 -0.236951581729675 0.0
-0.111517298776384 -0.0693158637483764 0.0
-0.467360651664167 -0.00417305794718525 0.0
-0.290122800504937 -0.367221062794388 0.0
-0.583259370217135 -0.309000481165462 0.0
-0.620279028606546 -0.193810630781977 0.0
-0.394822369792498 -0.293598970181061 0.0
-0.559620161183027 -0.408270579047439 0.0
-0.631721342609359 -0.153075103820135 0.0
-0.619429906890197 -0.189577074330231 0.0
-0.731057466432358 -0.352882172937049 0.0
-0.446372391091788 -0.375445366898909 0.0
-0.472271686019544 -0.217583775069567 0.0
-0.52651947486138 -0.247353411608355 0.0
-0.580922837679425 -0.296118381635174 0.0
-0.456499659915844 -0.254172860565824 0.0
-0.668673305794483 -0.373626691948152 0.0
0.236612131137147 0.41589767921192 1.0
0.411083017208874 0.239042056297528 1.0
0.453466145053352 0.115387043091102 1.0
-0.0102908723290331 0.14405831199131 1.0
-0.453680708330411 0.358591309928006 1.0
0.185889068991341 0.431970611825086 1.0
-0.132514399811759 0.177025232135253 1.0
-0.481660732929306 0.248221509282394 1.0
-0.137359790161575 0.431281610691773 1.0
0.220157829060032 0.0260739073368095 1.0
-0.345031544353912 0.316134485985404 1.0
0.367943408415772 0.155916428246418 1.0
-0.408324350732455 0.221997603171009 1.0
0.075421131530484 0.196129619803228 1.0
-0.599276556982756 0.0858331792398659 1.0
-0.45956226089733 0.277406318199334 1.0
0.231860903132459 0.4394992763828 1.0
-0.0649397655069122 0.310816393994891 1.0
0.323736370617222 0.322191312231979 1.0
-0.133928622084176 0.167857292506533 1.0
-0.0863464613983471 0.250941890871835 1.0
-0.187477315533606 0.649776371819141 1.0
-0.25358904005712 0.249115556915997 1.0
0.0645922121194815 0.487258388505616 1.0
0.568567488594932 -0.00272502413683529 1.0
-0.207555330349635 0.383332423041426 1.0
-0.184826649009305 0.418166207649886 1.0
0.194604608620517 0.444698701185889 1.0
0.0497903089669534 0.558265148478847 1.0
0.201995377848432 0.391867102817966 1.0
0.0260011769622084 0.520397483042804 1.0
0.32077702968927 0.407964175196049 1.0
0.3214334318896 0.414255535091991 1.0
0.441539232565149 0.280747043338404 1.0
-0.183814798699787 0.470714917361633 1.0
-0.234623773126082 0.482068799807107 1.0
-0.39876299122206 0.066835947163574 1.0
0.132311031969631 0.233219802053533 1.0
-0.203237436026396 0.446450045406071 1.0
-0.491302147167694 0.239224246612165 1.0
-0.222211503722452 0.405617951465357 1.0
-0.178415143839603 0.383375676136284 1.0
0.0882037060890757 0.391561686268333 1.0
-0.045962789794469 0.316898298360223 1.0
0.127165256509655 0.486899646495961 1.0
-0.168359803032324 0.379932118646268 1.0
0.334941837930111 0.401936154740588 1.0
0.0308559770308645 0.412679141675129 1.0
0.331338297416002 0.269832519967769 1.0
-0.0325248675471024 0.459852303632666 1.0
-0.363261831596653 0.391179205652417 1.0
-0.258928676357361 0.282827495514696 1.0
0.00122122332849961 0.383225394801441 1.0
0.311152098767665 0.270193984495277 1.0
0.0992201563195347 0.507021834970808 1.0
0.018732699925375 0.456899550181211 1.0
-0.0613720074246993 0.777543092809729 1.0
-0.0253109370219557 0.559167632135412 1.0
0.00125729313961842 0.442898173939617 1.0
0.512737594142009 0.171179320015026 1.0
0.0910642272368429 0.41136667831199 1.0
-0.220708758352761 0.359648555968802 1.0
-0.357855635721731 0.2433096368635 1.0
0.196736865630536 0.268406268216407 1.0
0.0466832534077146 0.102594072565724 1.0
-0.276379589796079 0.348247689039462 1.0
-0.0580494972316168 0.443300887226118 1.0
-0.0561569952617862 0.289123497036994 1.0
0.318351542211413 0.369040097918439 1.0
0.0616894440958766 0.430372898704839 1.0
0.335113252407486 0.233117539074665 1.0
0.338878409658823 -0.0163054182203504 2.0
0.525473385691879 -0.0685182057994609 2.0
0.610608686811755 -0.0387597324912212 2.0
0.470503300711184 -0.0780055196287675 2.0
0.559522390560273 0.0455607504869564 2.0
0.70088762628385 -0.105783624917323 2.0
0.868367097504515 -0.0407280840587472 2.0
0.862780922737714 -0.0782006550163326 2.0
0.683749542109501 -0.13388880697513 2.0
0.490719364904787 0.000556742596972359 2.0
0.569453098116811 0.0053646151062483 2.0
0.424677352021391 -0.0995305796215085 2.0
0.615620030410754 -0.0477516359244463 2.0
0.522173329442338 -0.138443761782292 2.0
0.477614571768238 -0.227298229593696 2.0
0.579679657370043 -0.157739251288844 2.0
0.915216445685003 -0.214231373001592 2.0
0.736788844386602 -0.252267555063631 2.0
0.603862840859783 -0.327962912612413 2.0
0.654244134785643 -0.37855515353659 2.0
0.520842262793416 -0.383062058379084 2.0
0.549621418081059 -0.334917376900725 2.0
0.406086595209614 -0.190280414009748 2.0
0.667875156431584 -0.447135703759609 2.0
0.62678579818521 -0.0889615490328759 2.0
0.817055087023848 -0.391798467813144 2.0
0.65097987150659 -0.412211744935431 2.0
0.713430626539064 -0.2043714725598 2.0
0.256575999481253 -0.645305841955962 2.0
0.373642517001621 -0.443532276221601 2.0
0.675012038973718 -0.143459279796379 2.0
0.517484678555807 -0.229187275777799 2.0
0.626069564878052 -0.046280519483056 2.0
0.579008532115868 -0.13907481600186 2.0
0.624375675319297 -0.348868311850617 2.0
0.766349787578844 -0.223018531815834 2.0
0.563011191989018 -0.476565598054751 2.0
0.624388906267386 -0.288769367399123 2.0
0.472514205644082 -0.360781031699105 2.0
0.482393360937639 -0.415988803475432 2.0
0.679182280356414 0.0657518883953787 2.0
0.774530366872655 -0.181913660378091 2.0
0.537656393695778 -0.462404696296112 2.0
0.739509502688215 -0.471901149485667 2.0
0.581781138998914 -0.348365752315639 2.0
0.626312887668318 -0.546857013070591 2.0
0.572991102409656 -0.42551608676412 2.0
0.701763997101223 -0.513504982683948 2.0
};
\end{axis}

\end{tikzpicture}

%% file: figs/pca_wine_kauri.tex
\begin{tikzpicture}

\definecolor{darkgray176}{RGB}{176,176,176}

\begin{axis}[
tick align=outside,
tick pos=left,
x grid style={darkgray176},
xlabel={Component 1},
xmin=-0.923263684098084, xmax=1.00276311853182,
xtick style={color=black},
y grid style={darkgray176},
ymin=-0.716448288694247, ymax=0.848685539548014,
ytick style={color=black},
width=0.3\linewidth
]
\addplot [colormap/viridis, only marks, scatter, scatter src=explicit]
table [x=x, y=y, meta=colordata]{%
x  y  colordata
-0.706335755973939 -0.253192752857043 0.0
-0.484976801771771 -0.00882289142063541 0.0
-0.521172266128946 -0.189187221677481 0.0
-0.821643663444924 -0.580905512257258 0.0
-0.202546381570955 -0.0594665740151188 0.0
-0.608190152120248 -0.487519191054382 0.0
-0.544047398643627 -0.300196496926384 0.0
-0.47435749548749 -0.298197020592497 0.0
-0.500432011957094 -0.307602858921239 0.0
-0.62751796867287 -0.206328232885875 0.0
-0.727467157091796 -0.356512043609109 0.0
-0.374967743832642 -0.225424534904278 0.0
-0.448188282848203 -0.231938139182621 0.0
-0.626345328675927 -0.355138676571354 0.0
-0.83571701125127 -0.538047802465009 0.0
-0.471931567976655 -0.337405385467885 0.0
-0.426990905027504 -0.450842684419911 0.0
-0.366595703552089 -0.315750340879042 0.0
-0.718788533345653 -0.59388133189827 0.0
-0.458884985639634 -0.175782239611429 0.0
-0.661852288038142 -0.127831031585363 0.0
-0.267900031953801 0.00981127565393747 0.0
-0.599782398827993 0.000782494522600041 0.0
-0.417894799937014 0.113786900972701 0.0
-0.449940391686938 0.0983771974086478 0.0
-0.215787795169716 0.00926270331364054 0.0
-0.415531995702532 -0.155470384900159 0.0
-0.283946970073552 -0.0897364262250318 0.0
-0.47509922696507 -0.088051647935278 0.0
-0.514396469295947 -0.104602501953416 0.0
-0.520965811951038 -0.258430884386146 0.0
-0.575738232641882 -0.361155787969285 0.0
-0.302144638806723 -0.0514943064152287 0.0
-0.367021971487879 -0.268616629871898 0.0
-0.303022468117356 -0.121413201641094 0.0
-0.458338877641686 -0.0303098821876442 0.0
-0.271566324740124 -0.0761238405816806 0.0
-0.232004188780535 -0.0503302762652722 0.0
-0.305480403391417 0.0616775562692842 0.0
-0.519605033037914 -0.313830061471668 0.0
-0.530575114358287 -0.129536622670152 0.0
-0.187044525570531 -0.124830305999712 0.0
-0.705508197447484 -0.236951581729675 0.0
-0.111517298776384 -0.0693158637483764 2.0
-0.467360651664167 -0.00417305794718525 0.0
-0.290122800504937 -0.367221062794388 0.0
-0.583259370217135 -0.309000481165462 0.0
-0.620279028606546 -0.193810630781977 0.0
-0.394822369792498 -0.293598970181061 0.0
-0.559620161183027 -0.408270579047439 0.0
-0.631721342609359 -0.153075103820135 0.0
-0.619429906890197 -0.189577074330231 0.0
-0.731057466432358 -0.352882172937049 0.0
-0.446372391091788 -0.375445366898909 0.0
-0.472271686019544 -0.217583775069567 0.0
-0.52651947486138 -0.247353411608355 0.0
-0.580922837679425 -0.296118381635174 0.0
-0.456499659915844 -0.254172860565824 0.0
-0.668673305794483 -0.373626691948152 0.0
0.236612131137147 0.41589767921192 2.0
0.411083017208874 0.239042056297528 2.0
0.453466145053352 0.115387043091102 1.0
-0.0102908723290331 0.14405831199131 2.0
-0.453680708330411 0.358591309928006 2.0
0.185889068991341 0.431970611825086 2.0
-0.132514399811759 0.177025232135253 2.0
-0.481660732929306 0.248221509282394 2.0
-0.137359790161575 0.431281610691773 2.0
0.220157829060032 0.0260739073368095 2.0
-0.345031544353912 0.316134485985404 2.0
0.367943408415772 0.155916428246418 1.0
-0.408324350732455 0.221997603171009 2.0
0.075421131530484 0.196129619803228 2.0
-0.599276556982756 0.0858331792398659 0.0
-0.45956226089733 0.277406318199334 2.0
0.231860903132459 0.4394992763828 2.0
-0.0649397655069122 0.310816393994891 2.0
0.323736370617222 0.322191312231979 2.0
-0.133928622084176 0.167857292506533 2.0
-0.0863464613983471 0.250941890871835 2.0
-0.187477315533606 0.649776371819141 2.0
-0.25358904005712 0.249115556915997 2.0
0.0645922121194815 0.487258388505616 2.0
0.568567488594932 -0.00272502413683529 1.0
-0.207555330349635 0.383332423041426 2.0
-0.184826649009305 0.418166207649886 2.0
0.194604608620517 0.444698701185889 2.0
0.0497903089669534 0.558265148478847 2.0
0.201995377848432 0.391867102817966 2.0
0.0260011769622084 0.520397483042804 2.0
0.32077702968927 0.407964175196049 2.0
0.3214334318896 0.414255535091991 2.0
0.441539232565149 0.280747043338404 2.0
-0.183814798699787 0.470714917361633 2.0
-0.234623773126082 0.482068799807107 2.0
-0.39876299122206 0.066835947163574 2.0
0.132311031969631 0.233219802053533 2.0
-0.203237436026396 0.446450045406071 2.0
-0.491302147167694 0.239224246612165 2.0
-0.222211503722452 0.405617951465357 2.0
-0.178415143839603 0.383375676136284 2.0
0.0882037060890757 0.391561686268333 2.0
-0.045962789794469 0.316898298360223 2.0
0.127165256509655 0.486899646495961 2.0
-0.168359803032324 0.379932118646268 2.0
0.334941837930111 0.401936154740588 2.0
0.0308559770308645 0.412679141675129 2.0
0.331338297416002 0.269832519967769 2.0
-0.0325248675471024 0.459852303632666 2.0
-0.363261831596653 0.391179205652417 2.0
-0.258928676357361 0.282827495514696 2.0
0.00122122332849961 0.383225394801441 2.0
0.311152098767665 0.270193984495277 2.0
0.0992201563195347 0.507021834970808 2.0
0.018732699925375 0.456899550181211 2.0
-0.0613720074246993 0.777543092809729 2.0
-0.0253109370219557 0.559167632135412 2.0
0.00125729313961842 0.442898173939617 2.0
0.512737594142009 0.171179320015026 1.0
0.0910642272368429 0.41136667831199 2.0
-0.220708758352761 0.359648555968802 2.0
-0.357855635721731 0.2433096368635 2.0
0.196736865630536 0.268406268216407 2.0
0.0466832534077146 0.102594072565724 2.0
-0.276379589796079 0.348247689039462 2.0
-0.0580494972316168 0.443300887226118 2.0
-0.0561569952617862 0.289123497036994 2.0
0.318351542211413 0.369040097918439 2.0
0.0616894440958766 0.430372898704839 2.0
0.335113252407486 0.233117539074665 2.0
0.338878409658823 -0.0163054182203504 1.0
0.525473385691879 -0.0685182057994609 1.0
0.610608686811755 -0.0387597324912212 1.0
0.470503300711184 -0.0780055196287675 1.0
0.559522390560273 0.0455607504869564 1.0
0.70088762628385 -0.105783624917323 1.0
0.868367097504515 -0.0407280840587472 1.0
0.862780922737714 -0.0782006550163326 1.0
0.683749542109501 -0.13388880697513 1.0
0.490719364904787 0.000556742596972359 1.0
0.569453098116811 0.0053646151062483 1.0
0.424677352021391 -0.0995305796215085 1.0
0.615620030410754 -0.0477516359244463 1.0
0.522173329442338 -0.138443761782292 1.0
0.477614571768238 -0.227298229593696 1.0
0.579679657370043 -0.157739251288844 1.0
0.915216445685003 -0.214231373001592 1.0
0.736788844386602 -0.252267555063631 1.0
0.603862840859783 -0.327962912612413 1.0
0.654244134785643 -0.37855515353659 1.0
0.520842262793416 -0.383062058379084 1.0
0.549621418081059 -0.334917376900725 1.0
0.406086595209614 -0.190280414009748 1.0
0.667875156431584 -0.447135703759609 1.0
0.62678579818521 -0.0889615490328759 1.0
0.817055087023848 -0.391798467813144 1.0
0.65097987150659 -0.412211744935431 1.0
0.713430626539064 -0.2043714725598 1.0
0.256575999481253 -0.645305841955962 1.0
0.373642517001621 -0.443532276221601 1.0
0.675012038973718 -0.143459279796379 1.0
0.517484678555807 -0.229187275777799 1.0
0.626069564878052 -0.046280519483056 1.0
0.579008532115868 -0.13907481600186 1.0
0.624375675319297 -0.348868311850617 1.0
0.766349787578844 -0.223018531815834 1.0
0.563011191989018 -0.476565598054751 1.0
0.624388906267386 -0.288769367399123 1.0
0.472514205644082 -0.360781031699105 1.0
0.482393360937639 -0.415988803475432 1.0
0.679182280356414 0.0657518883953787 1.0
0.774530366872655 -0.181913660378091 1.0
0.537656393695778 -0.462404696296112 1.0
0.739509502688215 -0.471901149485667 1.0
0.581781138998914 -0.348365752315639 1.0
0.626312887668318 -0.546857013070591 1.0
0.572991102409656 -0.42551608676412 1.0
0.701763997101223 -0.513504982683948 1.0
};
\end{axis}

\end{tikzpicture}

%% file: figs/pca_wine_tree.tex
\begin{tikzpicture}

\definecolor{darkgray176}{RGB}{176,176,176}

\begin{axis}[
tick align=outside,
tick pos=left,
x grid style={darkgray176},
xlabel={Component 1},
xmin=-0.923263684098084, xmax=1.00276311853182,
xtick style={color=black},
y grid style={darkgray176},
ymin=-0.716448288694247, ymax=0.848685539548014,
ytick style={color=black},
width=0.3\linewidth
]
\addplot [colormap/viridis, only marks, scatter, scatter src=explicit]
table [x=x, y=y, meta=colordata]{%
x  y  colordata
-0.706335755973939 -0.253192752857043 1.0
-0.484976801771771 -0.00882289142063541 1.0
-0.521172266128946 -0.189187221677481 1.0
-0.821643663444924 -0.580905512257258 1.0
-0.202546381570955 -0.0594665740151188 1.0
-0.608190152120248 -0.487519191054382 1.0
-0.544047398643627 -0.300196496926384 1.0
-0.47435749548749 -0.298197020592497 1.0
-0.500432011957094 -0.307602858921239 1.0
-0.62751796867287 -0.206328232885875 1.0
-0.727467157091796 -0.356512043609109 1.0
-0.374967743832642 -0.225424534904278 1.0
-0.448188282848203 -0.231938139182621 1.0
-0.626345328675927 -0.355138676571354 1.0
-0.83571701125127 -0.538047802465009 1.0
-0.471931567976655 -0.337405385467885 1.0
-0.426990905027504 -0.450842684419911 1.0
-0.366595703552089 -0.315750340879042 1.0
-0.718788533345653 -0.59388133189827 1.0
-0.458884985639634 -0.175782239611429 1.0
-0.661852288038142 -0.127831031585363 1.0
-0.267900031953801 0.00981127565393747 1.0
-0.599782398827993 0.000782494522600041 1.0
-0.417894799937014 0.113786900972701 1.0
-0.449940391686938 0.0983771974086478 1.0
-0.215787795169716 0.00926270331364054 1.0
-0.415531995702532 -0.155470384900159 1.0
-0.283946970073552 -0.0897364262250318 1.0
-0.47509922696507 -0.088051647935278 1.0
-0.514396469295947 -0.104602501953416 1.0
-0.520965811951038 -0.258430884386146 1.0
-0.575738232641882 -0.361155787969285 1.0
-0.302144638806723 -0.0514943064152287 1.0
-0.367021971487879 -0.268616629871898 1.0
-0.303022468117356 -0.121413201641094 1.0
-0.458338877641686 -0.0303098821876442 1.0
-0.271566324740124 -0.0761238405816806 1.0
-0.232004188780535 -0.0503302762652722 1.0
-0.305480403391417 0.0616775562692842 1.0
-0.519605033037914 -0.313830061471668 1.0
-0.530575114358287 -0.129536622670152 1.0
-0.187044525570531 -0.124830305999712 1.0
-0.705508197447484 -0.236951581729675 1.0
-0.111517298776384 -0.0693158637483764 2.0
-0.467360651664167 -0.00417305794718525 1.0
-0.290122800504937 -0.367221062794388 1.0
-0.583259370217135 -0.309000481165462 1.0
-0.620279028606546 -0.193810630781977 1.0
-0.394822369792498 -0.293598970181061 1.0
-0.559620161183027 -0.408270579047439 1.0
-0.631721342609359 -0.153075103820135 1.0
-0.619429906890197 -0.189577074330231 1.0
-0.731057466432358 -0.352882172937049 1.0
-0.446372391091788 -0.375445366898909 1.0
-0.472271686019544 -0.217583775069567 1.0
-0.52651947486138 -0.247353411608355 1.0
-0.580922837679425 -0.296118381635174 1.0
-0.456499659915844 -0.254172860565824 1.0
-0.668673305794483 -0.373626691948152 1.0
0.236612131137147 0.41589767921192 0.0
0.411083017208874 0.239042056297528 0.0
0.453466145053352 0.115387043091102 0.0
-0.0102908723290331 0.14405831199131 2.0
-0.453680708330411 0.358591309928006 2.0
0.185889068991341 0.431970611825086 2.0
-0.132514399811759 0.177025232135253 2.0
-0.481660732929306 0.248221509282394 2.0
-0.137359790161575 0.431281610691773 2.0
0.220157829060032 0.0260739073368095 0.0
-0.345031544353912 0.316134485985404 1.0
0.367943408415772 0.155916428246418 0.0
-0.408324350732455 0.221997603171009 2.0
0.075421131530484 0.196129619803228 2.0
-0.599276556982756 0.0858331792398659 1.0
-0.45956226089733 0.277406318199334 1.0
0.231860903132459 0.4394992763828 0.0
-0.0649397655069122 0.310816393994891 2.0
0.323736370617222 0.322191312231979 2.0
-0.133928622084176 0.167857292506533 1.0
-0.0863464613983471 0.250941890871835 2.0
-0.187477315533606 0.649776371819141 2.0
-0.25358904005712 0.249115556915997 1.0
0.0645922121194815 0.487258388505616 2.0
0.568567488594932 -0.00272502413683529 0.0
-0.207555330349635 0.383332423041426 2.0
-0.184826649009305 0.418166207649886 2.0
0.194604608620517 0.444698701185889 2.0
0.0497903089669534 0.558265148478847 2.0
0.201995377848432 0.391867102817966 2.0
0.0260011769622084 0.520397483042804 2.0
0.32077702968927 0.407964175196049 2.0
0.3214334318896 0.414255535091991 2.0
0.441539232565149 0.280747043338404 0.0
-0.183814798699787 0.470714917361633 2.0
-0.234623773126082 0.482068799807107 2.0
-0.39876299122206 0.066835947163574 1.0
0.132311031969631 0.233219802053533 2.0
-0.203237436026396 0.446450045406071 2.0
-0.491302147167694 0.239224246612165 2.0
-0.222211503722452 0.405617951465357 2.0
-0.178415143839603 0.383375676136284 2.0
0.0882037060890757 0.391561686268333 2.0
-0.045962789794469 0.316898298360223 2.0
0.127165256509655 0.486899646495961 2.0
-0.168359803032324 0.379932118646268 2.0
0.334941837930111 0.401936154740588 2.0
0.0308559770308645 0.412679141675129 2.0
0.331338297416002 0.269832519967769 2.0
-0.0325248675471024 0.459852303632666 2.0
-0.363261831596653 0.391179205652417 2.0
-0.258928676357361 0.282827495514696 2.0
0.00122122332849961 0.383225394801441 2.0
0.311152098767665 0.270193984495277 2.0
0.0992201563195347 0.507021834970808 2.0
0.018732699925375 0.456899550181211 2.0
-0.0613720074246993 0.777543092809729 2.0
-0.0253109370219557 0.559167632135412 2.0
0.00125729313961842 0.442898173939617 2.0
0.512737594142009 0.171179320015026 0.0
0.0910642272368429 0.41136667831199 2.0
-0.220708758352761 0.359648555968802 2.0
-0.357855635721731 0.2433096368635 2.0
0.196736865630536 0.268406268216407 2.0
0.0466832534077146 0.102594072565724 2.0
-0.276379589796079 0.348247689039462 2.0
-0.0580494972316168 0.443300887226118 2.0
-0.0561569952617862 0.289123497036994 2.0
0.318351542211413 0.369040097918439 2.0
0.0616894440958766 0.430372898704839 2.0
0.335113252407486 0.233117539074665 2.0
0.338878409658823 -0.0163054182203504 0.0
0.525473385691879 -0.0685182057994609 0.0
0.610608686811755 -0.0387597324912212 0.0
0.470503300711184 -0.0780055196287675 0.0
0.559522390560273 0.0455607504869564 0.0
0.70088762628385 -0.105783624917323 0.0
0.868367097504515 -0.0407280840587472 0.0
0.862780922737714 -0.0782006550163326 0.0
0.683749542109501 -0.13388880697513 0.0
0.490719364904787 0.000556742596972359 0.0
0.569453098116811 0.0053646151062483 2.0
0.424677352021391 -0.0995305796215085 1.0
0.615620030410754 -0.0477516359244463 0.0
0.522173329442338 -0.138443761782292 0.0
0.477614571768238 -0.227298229593696 0.0
0.579679657370043 -0.157739251288844 0.0
0.915216445685003 -0.214231373001592 0.0
0.736788844386602 -0.252267555063631 0.0
0.603862840859783 -0.327962912612413 0.0
0.654244134785643 -0.37855515353659 0.0
0.520842262793416 -0.383062058379084 0.0
0.549621418081059 -0.334917376900725 0.0
0.406086595209614 -0.190280414009748 0.0
0.667875156431584 -0.447135703759609 0.0
0.62678579818521 -0.0889615490328759 0.0
0.817055087023848 -0.391798467813144 0.0
0.65097987150659 -0.412211744935431 0.0
0.713430626539064 -0.2043714725598 0.0
0.256575999481253 -0.645305841955962 0.0
0.373642517001621 -0.443532276221601 0.0
0.675012038973718 -0.143459279796379 0.0
0.517484678555807 -0.229187275777799 0.0
0.626069564878052 -0.046280519483056 0.0
0.579008532115868 -0.13907481600186 0.0
0.624375675319297 -0.348868311850617 0.0
0.766349787578844 -0.223018531815834 0.0
0.563011191989018 -0.476565598054751 0.0
0.624388906267386 -0.288769367399123 0.0
0.472514205644082 -0.360781031699105 0.0
0.482393360937639 -0.415988803475432 0.0
0.679182280356414 0.0657518883953787 0.0
0.774530366872655 -0.181913660378091 0.0
0.537656393695778 -0.462404696296112 0.0
0.739509502688215 -0.471901149485667 0.0
0.581781138998914 -0.348365752315639 0.0
0.626312887668318 -0.546857013070591 0.0
0.572991102409656 -0.42551608676412 0.0
0.701763997101223 -0.513504982683948 0.0
};
\end{axis}

\end{tikzpicture}

%% file: figs_tex/fig_kernel_kmeans_clusters.tex
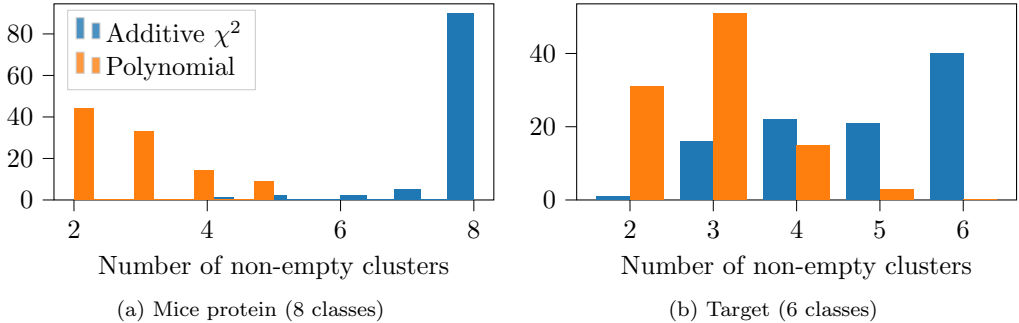
\begin{figure}
    \centering
    \subfloat[Mice protein (8 classes)]{
    \input{figs/kmeans_mice_clusters}
    \label{sfig:kernel_kmeans_clusters_mice}
    }
    \subfloat[Target (6 classes)]{
    \input{figs/kmeans_target_clusters}
    \label{sfig:kernel_kmeans_clusters_target}
    }
    \caption{Number of non-empty clusters for 100 runs of kernel KMeans with an additive $\chi^2$ or polynomial kernel. The algorithm had to find the same number of clusters as classes per dataset.}
    \label{fig:kernel_kmeans_clusters}
\end{figure}

%% file: figs/kmeans_mice_clusters.tex
\begin{tikzpicture}

\definecolor{darkgray176}{RGB}{176,176,176}
\definecolor{darkorange25512714}{RGB}{255,127,14}
\definecolor{lightgray204}{RGB}{204,204,204}
\definecolor{steelblue31119180}{RGB}{31,119,180}

\begin{axis}[
legend cell align={left},
legend style={
  fill opacity=0.8,
  draw opacity=1,
  text opacity=1,
  at={(0.03,0.97)},
  anchor=north west,
  draw=lightgray204
},
tick align=outside,
tick pos=left,
x grid style={darkgray176},
xlabel={Number of non-empty clusters},
xmin=1.7, xmax=8.3,
xtick style={color=black},
y grid style={darkgray176},
ymin=0, ymax=94.5,
ytick style={color=black},
width=0.45\linewidth,
height=0.15\paperheight
]
\draw[draw=none,fill=steelblue31119180] (axis cs:4,0) rectangle (axis cs:4.4,1);
\addlegendimage{ybar,ybar legend,draw=none,fill=steelblue31119180}
\addlegendentry{Additive $\chi^2$}

\draw[draw=none,fill=steelblue31119180] (axis cs:4.4,0) rectangle (axis cs:4.8,0);
\draw[draw=none,fill=steelblue31119180] (axis cs:4.8,0) rectangle (axis cs:5.2,2);
\draw[draw=none,fill=steelblue31119180] (axis cs:5.2,0) rectangle (axis cs:5.6,0);
\draw[draw=none,fill=steelblue31119180] (axis cs:5.6,0) rectangle (axis cs:6,0);
\draw[draw=none,fill=steelblue31119180] (axis cs:6,0) rectangle (axis cs:6.4,2);
\draw[draw=none,fill=steelblue31119180] (axis cs:6.4,0) rectangle (axis cs:6.8,0);
\draw[draw=none,fill=steelblue31119180] (axis cs:6.8,0) rectangle (axis cs:7.2,5);
\draw[draw=none,fill=steelblue31119180] (axis cs:7.2,0) rectangle (axis cs:7.6,0);
\draw[draw=none,fill=steelblue31119180] (axis cs:7.6,0) rectangle (axis cs:8,90);
\draw[draw=none,fill=darkorange25512714] (axis cs:2,0) rectangle (axis cs:2.3,44);
\addlegendimage{ybar,ybar legend,draw=none,fill=darkorange25512714}
\addlegendentry{Polynomial}

\draw[draw=none,fill=darkorange25512714] (axis cs:2.3,0) rectangle (axis cs:2.6,0);
\draw[draw=none,fill=darkorange25512714] (axis cs:2.6,0) rectangle (axis cs:2.9,0);
\draw[draw=none,fill=darkorange25512714] (axis cs:2.9,0) rectangle (axis cs:3.2,33);
\draw[draw=none,fill=darkorange25512714] (axis cs:3.2,0) rectangle (axis cs:3.5,0);
\draw[draw=none,fill=darkorange25512714] (axis cs:3.5,0) rectangle (axis cs:3.8,0);
\draw[draw=none,fill=darkorange25512714] (axis cs:3.8,0) rectangle (axis cs:4.1,14);
\draw[draw=none,fill=darkorange25512714] (axis cs:4.1,0) rectangle (axis cs:4.4,0);
\draw[draw=none,fill=darkorange25512714] (axis cs:4.4,0) rectangle (axis cs:4.7,0);
\draw[draw=none,fill=darkorange25512714] (axis cs:4.7,0) rectangle (axis cs:5,9);
\end{axis}

\end{tikzpicture}

%% file: figs/kmeans_target_clusters.tex
\begin{tikzpicture}

\definecolor{darkgray176}{RGB}{176,176,176}
\definecolor{darkorange25512714}{RGB}{255,127,14}
\definecolor{lightgray204}{RGB}{204,204,204}
\definecolor{steelblue31119180}{RGB}{31,119,180}

\begin{axis}[
legend cell align={left},
legend style={fill opacity=0.8, draw opacity=1, text opacity=1, draw=lightgray204},
legend pos=outer north east,
tick align=outside,
tick pos=left,
x grid style={darkgray176},
xlabel={Number of non-empty clusters},
xmin=1.36, xmax=6.64,
xtick style={color=black},
y grid style={darkgray176},
ymin=0, ymax=53.55,
ytick style={color=black},
width=0.45\linewidth,
height=0.15\paperheight
]
\draw[draw=none,fill=steelblue31119180] (axis cs:1.6,0) rectangle (axis cs:2,1);
\addlegendimage{ybar,ybar legend,draw=none,fill=steelblue31119180}

\draw[draw=none,fill=steelblue31119180] (axis cs:2.6,0) rectangle (axis cs:3,16);
\draw[draw=none,fill=steelblue31119180] (axis cs:3.6,0) rectangle (axis cs:4,22);
\draw[draw=none,fill=steelblue31119180] (axis cs:4.6,0) rectangle (axis cs:5,21);
\draw[draw=none,fill=steelblue31119180] (axis cs:5.6,0) rectangle (axis cs:6,40);
\draw[draw=none,fill=darkorange25512714] (axis cs:2,0) rectangle (axis cs:2.4,31);
\addlegendimage{ybar,ybar legend,draw=none,fill=darkorange25512714}

\draw[draw=none,fill=darkorange25512714] (axis cs:3,0) rectangle (axis cs:3.4,51);
\draw[draw=none,fill=darkorange25512714] (axis cs:4,0) rectangle (axis cs:4.4,15);
\draw[draw=none,fill=darkorange25512714] (axis cs:5,0) rectangle (axis cs:5.4,3);
\draw[draw=none,fill=darkorange25512714] (axis cs:6,0) rectangle (axis cs:6.4,0);
\end{axis}

\end{tikzpicture}

%% file: figs_tex/fig_us_votes_tree.tex
\begin{wrapfigure}{R}{0.5\linewidth}
    \centering
    \begin{scaletikzpicturetowidth}{\linewidth}\begin{tikzpicture}[scale=\tikzscale]
        \node [rectangle,draw, fill=red, align=center] (root){$\leaf_1$ [$\cluster_1$]\\$\stargain$}
          child {
            node [rectangle,draw, fill=cyan, align=center] (l11) {$\leaf_2$ [$\cluster_2$]\\$\switchgain$}
            child {
              node [rectangle,draw, fill=red, align=center] (l21) {$\leaf_4$ [$\cluster_1$]\\$\switchgain$}
              child {
                node [rectangle,draw, fill=cyan] (l31) {$\leaf_6$ [$\cluster_2$]}
                edge from parent node[above, anchor=east] {MX: no}
              }
              child {
                node [rectangle,draw, fill=red] (l32) {$\leaf_7$ [$\cluster_1$]}
                edge from parent node[above, anchor=west] {MX: yes, ?}
              }
              edge from parent node [above, anchor=east] {NC: no}
            }
            child {
              node [rectangle,draw, fill=cyan, align=center] (l22) {$\leaf_5$ [$\cluster_2$]}
              edge from parent node [above, anchor=west] {NC: yes,?}
            }
            edge from parent node [above, anchor=south east] {SA: no, ?}
          }
          child {
            node [rectangle ,draw, fill=red, align=center] (l12) {$\leaf_3$ [$\cluster_1$]}
            edge from parent node[above, anchor = south west] {SA: yes}
          };
    \end{tikzpicture}\end{scaletikzpicturetowidth}
    \caption{The unsupervised Kauri tree for 2 clusters on the Congressional votes dataset. SA stands for the El Salvador Aid vote, NC for the Nicaraguan Contras vote and MX for the MX-missile vote. The question mark means that the voter did not vote or was missing. Nodes contain their name, the associated cluster to which they assign samples and the type of split that occurred during learning. See Section~\ref{ssec:gain_metrics} for the split notations and App.~\ref{app:kauri_gains} for their computation.}
    \label{fig:kauri_us_votes}
\end{wrapfigure}
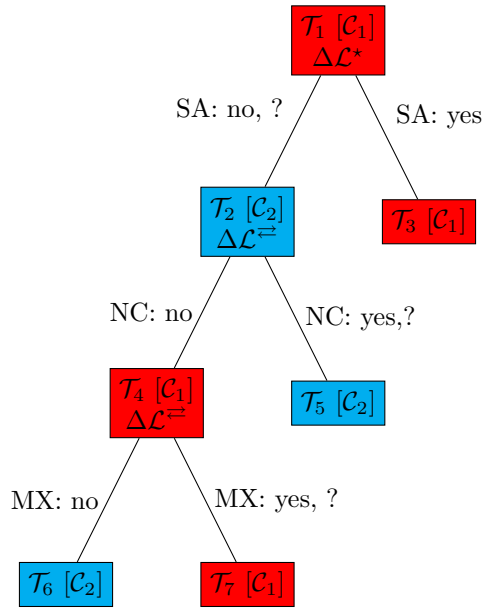

%% file: sections/conclusion.tex
We introduced a novel model optimising a kernel KMeans objective: Kauri. By rephrasing the kernel KMeans objective to drop the requirement for centroids, we leveraged an easy criterion to maximise and derived gains for an iterative splitting procedure in a CART-like binary tree. The performances of Kauri in clustering and structure are on par with those of related work, with an improvement on the weighted average explanation sizes. Moreover, we also compared recent unsupervised trees with a naive KMeans+Decision Tree combination and observed that remains a strong baseline to include in future works. In contrast to related work on unsupervised binary decision trees, Kauri is compatible with kernels other than the linear kernel. Moreover, we showed that the algorithm does not suffer from an empty cluster phenomenon that arises in the kernel KMeans algorithm and will fill all clusters. Hence, the strong advantage of this method is building an interpretable by-nature clustering instead of seeking to explain another clustering output from a different algorithm. Future work will focus on the integration of linear splits using several features, such as in oblique trees~\citep{gabidolla_optimal_2022}.

%% file: appendix/related_hierarchical_methods.tex
Trees are closely related to divisive hierarchical clustering algorithms. Indeed, divisive algorithms are built top-down~\cite{roux_comparative_2018} and if we give up the interpretability of trees by not restraining the splits on data-dependent thresholds, we can sort the data per node anyhow for the best split. In such case, the proposals of split become combinatorial and reach $2^{n-1}-1$ proposals for $n$ samples~\cite{edwards_method_1965}. These models are therefore costly, yet account for few iterations as we only need $K-1$ splits to obtain $K$ clusters.

Multiple splitting procedures were proposed to alleviate computation of all splits~\cite{williams_multivariate_1959, hubert_monotone_1973} to even KMeans~\cite{mollineda_relative_2000}. One of the most known divisive clustering algorithms is DIANA~\cite{kaufman_finding_1990}. \citet{roux_comparative_2018} suggests that ratio-based splits are among the most efficient.

To further accelerate the evaluation of the gains procedures and constraining the search space, combinations of divisive clustering algorithms with model-based clustering algorithms were proposed, hence using the maximum likelihood as a global objective for the model~\cite{sharma_divisive_2017, burghardt_agglomerative_2022}. We can still note the usage of KMeans as a heuristic for proposing splits at each node~\cite{sharma_divisive_2017}. However, the improvements brought by model-based clustering require parametric hypotheses which may constrain the exploration of the tree.


%% file: appendix/proof_equality.tex
We prove in this appendix the equality linking the definitions of the cluster sum of squares with and without centroids. Let $\cluster \subseteq \{\x_1 \ldots \x_n\}$ be a set of samples, $\rkhs$ be a kernel Hilbert space with projection $\varphi$. For this specific set, the cluster sum of squares (CSS) is defined as:

\begin{equation}
    \text{CSS} = \sum_{\x\in\cluster} \norm{\varphi(\x)-\vec{\mu}}^2_\rkhs,
\end{equation}

where $\vec{\mu} = \sum_{\x\in\cluster} \varphi(\x) / \card{\cluster}$ is the set centroid. We start by rewriting this equation using the kernel trick:

\begin{equation}
    \text{CSS} = \sum_{\x\in\cluster} \left(\langle \varphi(\x), \varphi(\x)\rangle + \langle \vec{\mu}, \vec{\mu} \rangle - 2 \langle \varphi(\x), \vec{\mu} \rangle \right). 
\end{equation}

There, we start with the second term, which does not depend on the sum over $\x$. Then, we can perform the sum over the last term using the bilinearity of the dot product to get a new term depending only on the centroid:

\begin{align}
    \text{CSS} &= \card{\cluster} \langle \vec{\mu}, \vec{\mu} \rangle + \sum_{\x\in\cluster} \left( \langle \varphi(\x), \varphi(x)\rangle - 2 \langle \varphi(\x), \vec{\mu} \rangle \right) \\
    &= \card{\cluster} \langle \vec{\mu}, \vec{\mu} \rangle - 2 \langle \sum_{\x\in\cluster} \varphi(\x), \vec{\mu} \rangle +
    3\sum_{\x\in\cluster} \langle \varphi(\x), \varphi(\x)\rangle\\
    &= - \card{\cluster}  \langle \vec{\mu}, \vec{\mu} \rangle + \sum_{\x\in\cluster} \langle \varphi(\x), \varphi(\x)\rangle.
\end{align}

Now, we can work on the term that depends only on $\x$. We simply sum it $\card{\cluster}$ additional times, here on a phantom variable $\vec{y}$ that we use to break the sum in two identical sums:

\begin{align}
    \text{CSS} &= \card{\cluster} \langle \vec{\mu}, \vec{\mu} \rangle + \inv{\card{\cluster}}\sum_{\x, \vec{y} \in \cluster\times\cluster} \langle \varphi(\x), \varphi(\x) \rangle,\\
    &=\card{\cluster} \langle \vec{\mu}, \vec{\mu} \rangle + \inv{2\card{\cluster}} \sum_{\x, \vec{y} \in \cluster\times\cluster} \langle \varphi(\x), \varphi(\x) \rangle + \inv{2\card{\cluster}} \sum_{\x, \vec{y}\in\cluster \times \cluster} \langle \varphi(\vec{y}), \varphi(\vec{y}) \rangle.
\end{align}

Finally, we express in the first term each centroid depending on a sample $\x$ or $\vec{y}$ before using the dot product bilinearity and get:

\begin{align}
    \text{CSS} &=\card{\cluster} \langle \inv{\card{\cluster}} \sum_{\x\in\cluster} \varphi(\x), \vec{\mu} \rangle + \inv{2\card{\cluster}} \sum_{\x, \vec{y} \in \cluster\times\cluster} \langle \varphi(\x), \varphi(\x) \rangle + \inv{2\card{\cluster}} \sum_{\x, \vec{y}\in\cluster \times \cluster} \langle \varphi(\vec{y}), \varphi(\vec{y}) \rangle,\\
    &= \inv{\card{\cluster}} \sum_{\x,\vec{y}\in \cluster \times \cluster}\langle \varphi(\x), \varphi(\vec{y}) \rangle + \inv{2\card{\cluster}} \sum_{\x, \vec{y} \in \cluster\times\cluster} \langle \varphi(\x), \varphi(\x) \rangle + \inv{2\card{\cluster}} \sum_{\x, \vec{y}\in\cluster \times \cluster} \langle \varphi(\vec{y}), \varphi(\vec{y}) \rangle.
\end{align}

The expression above is exactly the norm between the samples $\x$ and $\vec{y}$ expressed using the kernel trick in the Hilbert space. Therefore, we conclude:

\begin{equation}
    \sum_{\x\in\cluster} \langle \varphi(\x), \varphi(x)\rangle + \langle \vec{\mu}, \vec{\mu} \rangle - 2 \langle \varphi(\x), \vec{\mu} \rangle = \inv{2\card{\cluster}} \sum_{\x, \vec{y} \in \cluster\times \cluster} \norm{\varphi(\x) - \varphi(\vec{y})}^2_\rkhs
\end{equation}

%% file: appendix/kauri_objective.tex
We show here how to derive the Kauri objective function from both the one-vs-all and one-vs-one squared-MMD-GEMINI defined by~\citet{ohl_generalised_2022}.

\subsection{MMD-OvA}

We start by recovering the definition of the OvA squared-MMD-GEMINI using only the outputs of the outputs of our model $p_\theta(y|\vec{x})$:
\begin{equation}
    \gemini[{\mmd}^2][\text{ova}]{\x;y|\theta} = \expectation{y\sim \p(y)}{\expectation{\x_a, \x_b \sim p(\x)}{\kappa(\x_a,\x_b)\left(\frac{\p(y|\x_a)\p(y|\x_b)}{\p(y)^2}+1-2\frac{\p(y|\x_a)}{\p(y)}\right)}}.
\end{equation}
We can replace the expectations with discrete sums for both the clusters and the data. Notice the factor $\frac{1}{n}$ in front of the kernel as we are simply doing a Monte Carlo estimate. We replace at the same time, the values of the distributions by either the indicator functions or cluster sizes:
\begin{equation}
    \gemini[{\mmd}^2][\text{ova}]{\x;y|\theta} = \sum_{k=1}^K \frac{\card{\cluster_k}}{n}\sum_{\substack{i=1\\j=1}}^n \frac{\kappa(\x_i, \x_j)}{n^2}\left(\frac{\ind[\x_i\in\dataspace_k]\ind[\x_j\in\dataspace_k]n^2}{\card{\cluster_k}^2}+1-2\frac{\ind[\x_i\in\dataspace_k]n}{\card{\cluster_k}}\right).
\end{equation}
By applying the indicator functions, we see that we sum the terms of a kernel on condition that the respective samples belong specifically to some subset of data. We can consequently rewrite the inner sum as a combination of \emph{kernel stocks} $\sigma$:
\begin{equation}
    \gemini[{\mmd}^2][\text{ova}]{\x;y|\theta} = \sum_{k=1}^K \frac{\card{\cluster_k}}{n}\left(\frac{\sigma(\cluster_k^2)}{\card{\cluster_k}^2} + \frac{\sigma(\dataset^2)}{n^2} -2\frac{\sigma(\cluster_k\times \dataset)}{n\card{\cluster_k}}\right).
\end{equation}
Then, a couple of simplifications give:
\begin{equation}
    \gemini[{\mmd}^2][\text{ova}]{\x;y|\theta} = \sum_{k=1}^K \frac{\sigma(\cluster_k^2)}{n\card{\cluster_k}} + \frac{\card{\cluster_k}\sigma(\dataset^2)}{n^3} -2\frac{\sigma(\cluster_k\times \dataset)}{n^2}.
\end{equation}
We can obtain the final form of the GEMINI by summing constant terms. Observing that $\sum_k \card{\cluster_k}=n$ and using the bilinearity of $\sigma$, we have:
\begin{equation}
    \gemini[{\mmd}^2][\text{ova}]{\x;y|\theta} = -\frac{\sigma(\dataset^2)}{n^2} + \sum_{k=1}^K \frac{\sigma(\cluster_k^2)}{n\card{\cluster_k}}.
\end{equation}
As we are interested in optimising the clustering assignments in the tree, we can remove all constant terms and factors that do not bring extra information. Hence, the final objective $\objective$ to maximise is:
\begin{equation}
    \objective = \sum_{k=1}^K \card{\cluster_k}^{-1} \sigma(\cluster_k^2).
\end{equation}

\subsection{MMD-OvO}

We now prove that the objective function $\mathcal{L}$ obtained in the previous section is also equivalent to maximising the OvO squared MMD-GEMINI in case of delta Dirac classifiers. We start by expressing the complete squared MMD GEMINI:

\begin{multline}
    \gemini[{\mmd}^2][\text{ovo}]{\x;y|\theta} = \mathbb{E}_{y_a,y_b \sim \p(y)} \left[\mathbb{E}_{\x_a,\x_b\sim p(\x)}\left[\kappa(\x_a,\x_b)\left(\frac{\p(y_a|\x_a)\p(y_a|\x_b)}{\p(y_a)^2} \right.\right.\right.\\\left.\left.\left.+ \frac{\p(y_b|\x_a)\p(y_b|\x_b)}{\p(y_b)^2} - 2\frac{\p(y_a|\x_a)\p(y_b|\x_b)}{\p(y_a)\p(y_b)}\right)\right]\right].
\end{multline}

As we exactly did for the OvA MMD-GEMINI demonstration, we apply the following tricks: discretising the expectations on the dataset $\dataset$, re-expressing the cluster proportions, simplifying the sums. Our discrete version is:

\begin{multline}
    \gemini[{\mmd}^2][\text{ovo}]{\x;y|\theta} = \sum_{k,k^\prime}^K\frac{\card{\cluster_k}\card{\cluster_{k^\prime}}}{n^2}\sum_{\substack{x_i\in\dataset\\x_j\in\dataset}} \frac{\kappa(\x_i,\x_j)}{n^2}\left(\frac{n^2\ind[\x_i\in\dataspace_k]\ind[\x_j\in\dataspace_k]}{\card{\cluster_k}^2}+ \frac{n^2\ind[\x_i\in\dataspace_{k^\prime}]\ind[\x_j\in\dataspace_{k^\prime}]}{\card{\cluster_{k^\prime}}^2} \right.\\\left. - 2\frac{n^2 \ind[\x_i\in\dataspace_k] \ind[\x_j\in\dataspace_{k^\prime}]}{\card{\cluster_k}\card{\cluster_{k^\prime}}}\right).
\end{multline}

We can cancel the factors in $n^2$ and replace our sum over indicator functions by the \emph{kernel stock} function $\sigma$. Thus, we obtain:

\begin{equation}
    \gemini[{\mmd}^2][\text{ovo}]{\x;y|\theta} = \sum_{k,k^\prime}^K \frac{\card{\cluster_k}\card{\cluster_{k^\prime}}}{n^2} \left(\frac{\sigma(\cluster_k\times \cluster_k)}{\card{\cluster_k}^2} + \frac{\sigma(\cluster_{k^\prime}\times\cluster_{k^\prime})}{\card{\cluster_{k^\prime}}^2} - 2\frac{\sigma(\cluster_k\times \cluster_{k^\prime})}{\card{\cluster_k}\card{\cluster_{k^\prime}}}\right).
\end{equation}

For the first two terms, we can cancel one part of the summation. Indeed, the \emph{kernel stock} on $\cluster_k$ does not depend on $k^\prime$, consequently, the sum over $k^\prime$ just multiplies this \emph{kernel stock} up to a factor $n$ which will be cancelled by the denominator $n$ at the very start. The same reasoning goes for the second term. Last but not least, we can cancel cluster sizes on the last term. Our expression is then:

\begin{equation}
    \gemini[{\mmd}^2][\text{ovo}]{\x;y|\theta} = 2\sum_{k=1}^K \frac{\sigma(\cluster_k\times\cluster_k)}{n\card{\cluster_k}} - 2 \sum_{k, k^\prime}^K \frac{\sigma(\cluster_k\times \cluster_{k^\prime})}{n^2}.
\end{equation}

As we now look forward to maximising this expression, we can realise that the last term is simply the full \emph{kernel stock}, in other words, a constant with respect to the clustering. We can then discard this term. For the first term, we simply remove the constant factors $2$ and $1/n$ to obtain the equivalent:

\begin{equation}
    \gemini[{\mmd}^2][\text{ovo}]{\x;y|\theta} = \frac{2}{n}\left(\sum_{k=1}^K \frac{\sigma(\cluster_k\times\cluster_k)}{\card{\cluster_k}} - \frac{\sigma(\dataset\times\dataset)}{n}\right) \propto \objective + \text{constant}.
\end{equation}

Therefore, the maximisation of an MMD-GEMINI for Dirac distributions as clustering models is equivalent to the minimisation of a kernel KMeans objective. This joins the observation of \citet{franca_kernel_2020}  who connected an objective similar to a one-vs-one squared MMD to the kernel KMeans. Notice that we removed a factor 2 in the final equation as it does not affect the argmin operator.

%% file: appendix/expression_gains.tex
We can derive from our objective (Eq.~\ref{eq:objective}) four metrics which correspond to different types of splits: the \emph{star gain}, the \emph{double star gain}, the \emph{switch gain} and the \emph{reallocation gain}.


\subsection{Creating a new cluster: the \emph{star gain}}

In this case, we assign one of the splits $\split_L$ or $\split_R$ to a new cluster and let the other split in the same cluster as the parent node, i.e. either $k_L=K+1$ and $k_R=k_p$ or $k_L=k_p$ and $k_R=K+1$. Taking the case where the left split is given to a new cluster, we derive from the global gain a variation that we call \emph{star gain}:

\begin{equation}
     \stargain(\split_L: k_p\rightarrow k_L) = \frac{\sigma(\split_L^2)}{\card{\split_L}} + \frac{\sigma({\cluster^\prime_{k_p}}^2)}{\card{\cluster^\prime_{k_p}}} - \frac{\sigma(\cluster^2_{k_p})}{\card{\cluster_{k_p}}}.
\end{equation}

However, that expression is not convenient since there is a clear dependence: $\cluster^\prime_{k_p} = \cluster_{k_p}\setminus \split_L$ and we would be interested in avoiding the evaluation of $\sigma({\cluster^\prime_{k_p}}^2)$. We can use the bilinearity of the sigma function and decompose over the new cluster $\cluster_{k_p}^\prime = \cluster_{k_p} \setminus \split_L$. Similarly, we can reexpress the cardinal as $\card{\cluster^\prime_{k_p}} = \card{\cluster_{k_p}}-\card{\split_L}$. Consequently, our term becomes:

\begin{equation}\label{eq:star_gain}
    \stargain(\split_L: k_p \rightarrow k_L) = \frac{\sigma(\split_L^2)}{\card{\split_L}} + \frac{\sigma(\cluster_{k_p}^2) - 2\sigma(\cluster_{k_p}\times\split_L) + \sigma(\split_L^2)}{\card{\cluster_{k_p}}-\card{\split_L}} - \frac{\sigma(\cluster_{k_p}^2)}{\card{\cluster_{k_p}}}.
\end{equation}

It is then just a matter of reordering with respect to the \emph{kernel stocks} $\sigma$ to obtain the final equation:

\begin{multline}
    \stargain(\split_L: k_p \rightarrow k_L) = \sigma(\split_L^2)\left(\inv{\card{\split_L}}+\inv{\card{\cluster_{k_p}}-\card{\split_L}}\right) + \sigma(\cluster_{k_p}^2)\left(\inv{\card{\cluster_{k_p}}-\card{\split_L}}-\inv{\card{\cluster_{k_p}}}\right) \\- 2\frac{\sigma(\cluster_{k_p}\times\split_L)}{\card{\cluster_{k_p}}-\card{\split_L}},
\end{multline}

which will be the used equation for the \emph{star gain}.

\subsection{Creating two clusters: the \emph{double star gain}}

The operation of creating two clusters can be seen as assigning in a first step the complete node $p$ to a new cluster, then taking one of its split and assigning it to a second new cluster. The \emph{double star gain} $\doublestargain$ can be thus computed by the sum of $\stargain$ with $\leaf_p$ replacing the source cluster $\cluster_{k_p}$ and another $\stargain$ with the $\leaf_p$ replacing $\split_L$:
\begin{equation}\label{eq:double_star_gain}
    \doublestargain(\split_L \rightarrow k_L, \split_R \rightarrow k_R) = \stargain(\leaf_p: k_p \rightarrow k_L) + \stargain(\split_R: k_L \rightarrow k_R)
\end{equation}

\subsubsection{Merging with another cluster: the \emph{switch gain}}

This type of split is very similar to the creation of a new one. The main difference is that as one of the child nodes will join another cluster, e.g. $k_p \neq k_L \leq K$, we must take into account in the gain that we must subtract the \emph{kernel stock} of the former target cluster. We call this type of gain the \emph{switch gain}:
\begin{equation}
    \switchgain(\split_L: k_p \rightarrow k_L) = \frac{\sigma({\cluster^\prime_{k_L}}^2)}{\card{\cluster^\prime_{k_L}}}+\frac{\sigma({\cluster^\prime_{k_p}}^2)}{\card{\cluster^\prime_{k_p}}}-\frac{\sigma(\cluster_{k_L}^2)}{\card{\cluster_{k_L}}}-\frac{\sigma(\cluster_{k_p}^2)}{\card{\cluster_{k_p}}},
\end{equation}
where we arbitrarily chose the left split for the equation. Similarly to the new cluster case, this expression can be completely re-written using only the original clusters and $\split_L$ to remove dependencies in the equation. We start by exploiting the bilinearity of $\sigma$. The first new cluster is the source one without the split elements and the second new cluster is the target one with added split elements. Therefore, we have $\cluster^\prime_{k_p} = \cluster_{k_p}\setminus \split_L$ and $\cluster^\prime_{k_L} = \cluster_{k_L} \cup \split_L$. We can deduce:

\begin{multline}
    \switchgain(\split_L: k_p \rightarrow k_L) = \frac{\sigma(\cluster_{k_p}^2)-2\sigma(\cluster_{k_p}\times\split_L)+\sigma(\split_L^2)}{\card{\cluster_{k_p}}-\card{\split_L}}+\frac{\sigma(\cluster_{k_L}^2)+2\sigma(\cluster_{k_L}\times\split_L) + \sigma(\split_L^2)}{\card{\cluster_{k_L}}+\card{\split_L}}\\-\frac{\sigma(\cluster_{k_p}^2)}{\card{\cluster_{k_p}}}-\frac{\sigma(\cluster_{k_L}^2)}{\card{\cluster_{k_L}}}.
\end{multline}

Then we finish again the demonstration by reordering the factors according to the respective stocks:

\begin{multline}\label{eq:switch_gain}
     \switchgain(\split_L: k_p \rightarrow k_L) = \sigma(\split_L^2)\left(\inv{\card{\cluster_{k_L}}+\card{\split_L}} + \inv{\card{\cluster_{k_p}}-\card{\split_L}}\right) - 2 \frac{\sigma(\cluster_{k_p}\times \split_L)}{\card{\cluster_{k_p}}-\card{\split_L}} \\+ \sigma(\cluster_{k_p}^2)\left(\inv{\card{\cluster_{k_p}}-\card{\split_L}} - \inv{\card{\cluster_{k_p}}}\right)  + \sigma(\cluster_{k_L}^2)\left(\inv{\card{\cluster_{k_L}} + \card{\split_L}} - \inv{\card{\cluster_{k_p}}}\right) + 2\frac{\sigma(\cluster_{k_L}\times \split_L)}{\card{\cluster_{k_L}} + \card{\split_L}}.
\end{multline}

which is the equation we use in Kauri switch splits.

\subsection{Reallocating content to the different clusters: the \emph{reallocation gain}}

As the tree grows, it may as well be interesting to reconsider whether samples that are currently in a cluster should all be in a new cluster. We call this process \emph{reallocation} as both splits of a node end up in two different clusters: $k_L\neq k_p$ and $k_R\neq k_p$. 

Contrary to the double cluster creation case, we cannot simply sum two \emph{switch gains} $\switchgain$ to compute the \emph{reallocation gain}. Indeed, the \emph{switch gain} assumes that the final state of the original cluster $\cluster_{k_p}$ still contains the complementary of the chosen split $\split_L$ (or $\split_R$) from the leaf samples $\leaf_p$ which is not true when both parts of the leaf go to different clusters. Hence, a corrective term $\epsilon$ is required.


When we sum two switch gains, the final state of the target clusters is correct: we simply added elements from a split. The corrective term thus only focuses on the state of the source cluster. Let $\cluster_k^\prime$ the state of the source cluster according to the first switch gain on the left split, $\cluster_k^\second$ the state of the source cluster according to the second switch gain on the right split and $\cluster_k^{\hookrightarrow}$ the true state after reallocating both left and right splits. Notice that we lighten the notation $k_p$ to $k$. The corrective term must satisfy:

\begin{equation}
    \frac{\sigma({\cluster_k^\prime}^2)}{\card{\cluster^\prime_k}} + \frac{\sigma({\cluster_k^\second}^2)}{\card{\cluster_k^\second}} + \epsilon = \frac{\sigma({\cluster_k^{\hookrightarrow}}^2)}{\card{\cluster_k^{\hookrightarrow}}}.
\end{equation}

We can rewrite each new definition of the source clusters using the left split $\split_L$ and right split $\split_R$. Thus we get:

\begin{equation}
    \frac{\sigma({\cluster_k\setminus\split_L}^2)}{\card{\cluster_k}-\card{\split_L}}+\frac{\sigma({\cluster_k\setminus\split_R}^2)}{\card{\cluster_k}-\card{\split_R}}+\epsilon = \frac{\sigma({\cluster_k\setminus\leaf_p}^2)}{\card{\cluster_k}-\card{\leaf_p}},
\end{equation}

which allows us to use the bilinearity of $\sigma$:

\begin{multline}
    \frac{\sigma(\cluster_k^2)-2\sigma(\cluster_k\times\split_L)+\sigma(\split_L^2)}{\card{\cluster_k}-\card{\split_L}}+\frac{\sigma(\cluster_k^2)-2\sigma(\cluster_k\times\split_R)+\sigma(\split_R^2)}{\card{\cluster_k}-\card{\split_R}}+\epsilon\\ = \frac{\sigma(\cluster_k^2)-2\sigma(\cluster_k\times\leaf_p)+\sigma({\leaf_p}^2)}{\card{\cluster_k}-\card{\leaf_p}}.
\end{multline}

Then, by reordering the terms and simplifying for the factor $\sigma(\cluster_k^2)$, we get the expression of $\epsilon$:

\begin{multline}\label{eq:corrective_term}
   \epsilon = \frac{\sigma(\cluster_{k_p}^2) + \sigma(\leaf_p^2) - 2 \sigma(\cluster_{k_p}\times \leaf)}{\card{\cluster_{k_p}}-\card{\leaf_p}} + \frac{\sigma (\cluster_{k_p}^2)}{\card{\cluster_{k_p}}} - \frac{\sigma(\cluster_{k_p}^2) + \sigma(\split_L^2) - 2\sigma(\cluster_{k_p}\times \split_L)}{\card{\cluster_{k_p}}-\card{\split_L}} \\ - \frac{\sigma(\cluster_{k_p}^2) + \sigma(\split_R^2) - 2\sigma(\cluster_{k_p}\times \split_R)}{\card{\cluster_{k_p}}-\card{\split_R}}.
\end{multline}

Thus, we can express the \emph{reallocation gain} $\reallocationgain$ as the sum of two switch gains assigning both left and right children nodes to different clusters plus the corrective term $\epsilon$.
\begin{equation}\label{eq:reallocation_gain}
   \reallocationgain(\split_L: k_p \rightarrow k_L, \split_R: k_p \rightarrow k_R) = \switchgain(\split_L: k_p\rightarrow k_L) + \switchgain(\split_R: k_p\rightarrow k_R) + \epsilon
\end{equation}

%% file: appendix/dynamic_algorithm.tex
\subsection{The Kauri Algorithm}

Upon looking at one leaf containing a small subset of samples, we need to find the best possible split according to a given threshold on a specified feature. While each feature specifies a different ordering and offers little space for optimisation, computing all possible gains may be time-consuming. Indeed, computing $\sigma(E\times F)$ is done in $\mathcal{O}(|E||F|)$, so evaluating gains for a proposal split $\split$ on a single feature for node samples $\leaf_p$ contributing to a cluster of $\cluster_k$ has a naive complexity of: $\mathcal{O}(|\split|^2+|\cluster_k|^2 + |\cluster_k||\split|)$ for $\stargain$, $\mathcal{O}(|\split|^2+|\cluster_k|^2 +|\leaf_p|^2 + |\cluster_k|(|\split|+|\leaf_p|))$ for $\doublestargain$, $\mathcal{O}(|\leaf_p|^2+|\cluster_k|^2 + \cluster_k||\leaf_p| +|\cluster_{k^\prime}|^2 + |\cluster_{k^\prime}||\leaf_p|)$ for each $k^\prime \neq k$ for $\switchgain$, and at worst $K$ times the previous complexity again for all pairs of assignable new clusters $k^\prime$, $k^\second$ in the \emph{reallocation gain} $\reallocationgain$. Therefore, iterating over all features and all possible splits needs to be optimised as this operation is the core of the tree construction.

\subsubsection{Pre-computing kernel stocks}

Most of the kernel stocks can be computed ahead in fact, and then the splitting choice would just need to access the value of the kernel stocks instead. To that end, we choose to formulate two matrices that will store all structural information. The matrix $\vec{Z}\in\{0,1\}^{\Tmax\times n}$ describes the membership of samples to leaves where $\Tmax$ is the maximal number of leaves allowed ($\Tmax \leq n$). As a sample can only belong to 1 leaf, each column of $\vec{Z}$ has a single 1. Similarly, the matrix $\vec{Y}\in\{0,1\}^{\Kmax\times\Tmax}$ describes the membership of leaves to clusters, and only one cluster is allowed per leaf. We can then compute most of the kernel stocks required for split computations, as:

\begin{equation}
    \vec{\Lambda} = [\sigma(\leaf_i \times \{\x_j\})] = \vec{Z}\vec{\kappa},
\end{equation}

is the matrix containing all stocks between leaves and single samples requiring $\mathcal{O}(n^2\Tmax)$ to compute, and:

\begin{equation}
    \vec{\gamma} = [\sigma(\cluster_i \times \cluster_j)] = \vec{Y}\vec{\Lambda}\vec{Z}^\top\vec{Y}^\top,
\end{equation}

is the matrix with cluster-cluster stocks, requiring $\mathcal{O}(n^2\Tmax + \Tmax^2\Kmax)$ for computations.

\subsubsection{Optimising split evaluation}

\input{figs_tex/alg_choosesplits}

Thanks to the formulation of the \emph{star gain} $\stargain$, the \emph{double star gain} $\doublestargain$ and assuming we know the variables $\sigma(\split_L\times\split_l)$ (resp. $\split_R$) and $\sigma(\split_L\times\cluster_k)$ (resp. $\split_R$), evaluating the creation of clusters is done in $\mathcal{O}(1)$. Inevitably, we achieve $\mathcal{O}(K)$ for the switch gains $\switchgain$ since evaluating these gains is easy but needs iteration over all clusters.

To alleviate the complexity of the \emph{reallocation gain} $\reallocationgain$ due to the exploration of all pairs of clusters, we propose to remember the top two switch gains per left children and right children. Indeed, the corrective term $\epsilon$ does not depend on the two clusters to which the left and right children will be reallocated. Hence maximising the \emph{reallocation gain} is the same as finding the combination of the best switch gains. Thus, remembering the top two switch gains and finding the best combination between left and right child, with different clusters membership per child, will yield the optimal \emph{reallocation gain}. Therefore we achieved the best gain in $\mathcal{O}(K)$ and the evaluation of all types of gain is done in $\mathcal{O}(K)$. 

\subsubsection{An iterative rule for split stocks}

\input{figs_tex/alg_findbestsplit}

Starting from here, we seek an update rule that allows us to easily update the kernel stocks of the splits $\sigma(\split\times \cluster_k)$ and $\sigma(\split^2)$ that are required by Algorithm~\ref{alg:choose_split}. We must explore all possible splits by considering thresholds on chosen variables. Therefore, a split on a variable must be done according to the ordering imposed by that variable, leaving a left child $\split_L^{(l)}$ and a right child $\split_R^{(l)}$ at the $l$-th threshold. The algorithm consist in starting from the split of a single sample to the left child $\split_L^{(1)}$ and all other samples to the right child $\split_R^{(1)}$ then progressively remove or add samples according to an ordering given by a sorted feature: $t = \nu(l)$ to compute $\split_L^{(l)}$ and $\split_R^{(l)}$. For example, if the $p$-th node has the data samples 5, 8, 9 and 15, a feature may order those as 9,8,15,5. Then, $\nu(1)=9$, $\nu(2)=8$, $\nu(3)=15$  and $\nu(4)=5$.

Note that there is a key difference regarding the indices notations. We write $i$ the absolute index of sample from the dataset $\dataset$ while $l$ refers to the ordered count of samples inside a specific node. The index $t$ is the absolute index according to the ordering $\nu(l)$.

We introduce 3 helping variables. The first one is the sample-wise cluster adversarial stocks:

\begin{equation}
    \vec{\omega}_{k,i} = \sigma(\cluster_k\times \{\x_t\})
\end{equation}

which does not depend on the ordering specified by a feature and will ease the computation of both $\sigma(\split_L\times\cluster_k)$ and $\sigma(\split_R\times \cluster_k)$. We can shortly write that $\vec{\omega} = \vec{Y}\vec{Z}\vec{\kappa}$. To alleviate the computations of $\sigma(\split_L^2)$ and $\sigma(\split_R^2)$, we introduce the ordering-dependent variables:

\begin{equation}
    \alpha_t^\nu = \sum_{\substack{l=1\cdots|\leaf_p|\\\nu(l)<t}} \kappa_{\nu(l), t},
\end{equation}

and:

\begin{equation}
    \beta_t^\nu = \sum_{\substack{l=1\cdots |\leaf_p|\\\nu(l) > t}} \kappa_{\nu(l), t}.
\end{equation}

\input{figs_tex/fig_alpha_beta}

These two variables verify a constant sum $\beta_t^\nu+\alpha_t^\nu+\kappa_{tt} = \sigma(\{x_t\}\times\leaf_p)$ for all $t\in\leaf_p$. We provide a visual intuition of the definition of these variables in Fig.~\ref{fig:visualisation_alpha_beta}. Once the variables $\vec{\omega}_{k,i}$, $\alpha_t^\nu$, $\beta_t^\nu$ are initialised, we can compute all split gains with simple additions.

The initialisation of the variables is easy. For the split self-stock, we have:

\begin{equation}
    \sigma(\split_L^{(0)}\times \split_L^{(0)}) = 0,
\end{equation}
\begin{equation}
    \sigma(\split_R^{(0)}\times \split_L^{(0)}) = \sigma(\leaf_p^2),
\end{equation}

because starting from no sample yields all content of the node $\leaf_p$ to the right split $\split_R^{(0)}$. The adversarial stocks follow the same logic:

\begin{equation}
    \sigma(\split_L^{(0)}\times\cluster_k) = 0,
\end{equation}
\begin{equation}
    \sigma(\split_R^{(0)}\times\cluster_k) = \sigma(\leaf_p\times\cluster_k),
\end{equation}

where the last term is simply an element of $\vec{\gamma}$ indexed by the respective leaf and cluster. The iterations then consist in removing adequate adversarial stock or self-kernel stock:

\begin{equation}\label{eq:update_self_left_split}
    \sigma(\split_L^{(l)}\times\split_L^{(l)}) = \sigma(\split_L^{(l-1)}\times \split_L^{(l-1)}) + 2\alpha_{\nu(l)}^\nu + \kappa_{\nu(l),\nu(l)},
\end{equation}
\begin{equation}\label{eq:update_self_right_split}
    \sigma(\split_R^{(l)}\times\split_R^{(l)}) = \sigma(\split_R^{(l-1)}\times\split_R^{(l-1)}) - 2\beta^\nu_{\nu(l)} - \kappa_{\nu(l),\nu(l)}.
\end{equation}

The adversarial scores are easier to update:

\begin{equation}\label{eq:update_cluster_left_split}
    \sigma(\split_L^{(l)} \times \cluster_k) = \sigma(\split_L^{(l-1)}\times \cluster_k) + \vec{\omega}_{k,\nu(l)},
\end{equation}

and conversely:

\begin{equation}\label{eq:update_cluster_right_split}
    \sigma(\split_R^{(l)} \times \cluster_k) = \sigma(\split_R^{(l-1)}\times \cluster_k) - \vec{\omega}_{k, \nu(l)}.
\end{equation}

Thanks to these iterative variables, the iterative computation of all kernel stocks can be achieved in  $\mathcal{O}(|\leaf_p|(|\leaf_p|+K))$ for a specific feature and node samples $\leaf_p$ as summarised in Algorithm~\ref{alg:find_best_split}. The pre-computing of $\vec{\omega}$ takes $\mathcal{O}(n^2)$ and can be done in advance at the tree level.

Finally, we can optimise the computation of all splits.

\subsubsection{Complete picture}
\input{figs_tex/alg_kauri}

The complete algorithm of KAURI is written in Algorithm~\ref{alg:tree_training}. We estimate the complexity of the split search to $\mathcal{O}(n((L+d)(n+K)+dL)+L^2(d+K))$ at worst and $\mathcal{O}(n(n+K)(d+L)+KL^2)$ at best, where $L$ is the current number of leaves. 

%% file: figs_tex/alg_choosesplits.tex
\begin{algorithm}
  \caption{ComputeSplits}\label{alg:choose_split}
    \begin{algorithmic}
        \STATE \textbf{Input:} $\sigma(\split_L^2)$ the stock of the left split, $|\split_L|$ the size of the left split, $\sigma(\split_R^2)$ the stock of the right split, $|\split_R|$ the size of the right split, $\sigma(\cluster_k) \forall k$ the stock of all clusters, $|\cluster_k| \forall k$ the size of all clusters, $\sigma(\cluster_k\times\split_L) \forall k$ the adversarial stocks of the left split, $\sigma(\cluster_k\times\split_R) \forall k$ the adversarial stocks of the right split, $\sigma(\leaf^2)$ the leaf stock, $k$ the indicator of the current cluster of the split leaf, $K$ the current number of clusters
            \Let{$\Delta\objective$}{0}
            \Let{\texttt{Split}}{$(-1, -1)$}\COMMENT{Stores the cluster targets of left and right splits}

            \IF {Creating a new cluster is allowed}
                \STATE Compute $\stargain_L$ using the left split $\split_L$ and Eq.~\ref{eq:star_gain}
                \STATE Compute $\stargain_R$ using the right split $\split_R$ and Eq.~\ref{eq:star_gain}
                \Let{$\Delta\objective$, \texttt{Split}}{\texttt{TakeBest}{\{$\Delta\objective$, \texttt{Split}\}, \{$\stargain_L$, ($K+1$, $k$)\}, \{$\stargain_R$,($k$, $K+1$)\}}}
            \ENDIF

            \IF {Creating two clusters is allowed}
                \STATE Compute $\doublestargain$ using either $\split_L$ or $\split_R$ and Eq.~\ref{eq:double_star_gain}
                \Let{$\Delta\objective$, \texttt{Split}}{\texttt{TakeBest}{\{$\Delta\objective$, \texttt{Split}\},\{$\doublestargain$, ($K+1$, $K+2$)\}}}
            \ENDIF
            
            \Let{\texttt{TopL}, \texttt{SecondL}, \texttt{TopR}, \texttt{SecondR}}{0,0,0,0}
            \Let{\texttt{TopkL}, \texttt{SecondkL}, \texttt{TopkR}, \texttt{SecondkR}}{-1,-1,-1,-1}
            
            \FOR {$k^\prime \in \{1,\cdots, K\}\setminus\{k\}$}
                \STATE Compute $\switchgain_L$ using the left split $\split_L$ and Eq.~\ref{eq:switch_gain}
                \STATE Compute $\switchgain_R$ using the right split $\split_L$ and Eq.~\ref{eq:switch_gain}

                \Let{$\Delta\objective$, \texttt{Split}}{\texttt{TakeBest}{\{$\Delta\objective$, \texttt{Split}\}, \{$\switchgain_L$, ($k^\prime$, $k$)\}, \{$\switchgain_R$, ($k$, $k^\prime$)\}}}
                

                \STATE Update \texttt{TopL}, \texttt{SecondL}, \texttt{TopkL} and \texttt{SecondkL} using $\switchgain_L$ to keep track of the two best switch gains and their respective target cluster $k^\prime$
                \STATE Update \texttt{TopR}, \texttt{SecondR}, \texttt{TopkR} and \texttt{SecondkR} using $\switchgain_R$ to keep track of the two best switch gains and their respective target cluster $k^\prime$                
            \ENDFOR
            
            \IF {Reallocation is allowed}
                \STATE Compute $\epsilon$  using Eq.~\ref{eq:corrective_term}
                \STATE Compute $\reallocationgain$ using \texttt{TopL}, \texttt{SecondL}, \texttt{TopkL}, \texttt{SecondkL}, \texttt{TopR}, \texttt{SecondR}, \texttt{TopkR}, \texttt{SecondkR} and $\epsilon$.
                \Let{$\Delta\objective$, \texttt{Split}}{\texttt{TakeBest}{\{$\Delta\objective$, \texttt{Split}\}, \{$\reallocationgain$, (target left, target right)\}}}
            \ENDIF
            \STATE Return $\Delta\objective$, \texttt{Split}
  \end{algorithmic}
\end{algorithm}

%% file: figs_tex/alg_findbestsplit.tex
\begin{algorithm}
  \caption{FindBestSplit}\label{alg:find_best_split}
    \begin{algorithmic}
        \STATE \textbf{Input :} $\leaf$ the set of samples in the leaf of length $|\leaf|$, $p$ the index of the leaf, $\nu$ an ordering precised by a feature of length $|\leaf|$, $\vec{\kappa}$ a kernel of shape $n\times n$, $\vec{\Lambda}$ the $\Tmax \times n$ leaf-sample stocks, $\vec{\omega}$ the $\Kmax\times n$ cluster-sample stocks, $\vec{\gamma}$ the $\Kmax \times \Kmax$ cluster-cluster stocks, $|\cluster_k|, \forall k$ the size of all clusters, $k$ the cluster of the considered leaf.
        
            \Let{$\sigma(\split_L^{(0)}\times\split_L^{(0)})$}{0} \COMMENT{Initialise all iteration variables}
            \Let{$\sigma(\split_R^{(0)}\times\split_R^{(0)})$}{$\sum_{\x_i\in\leaf} \vec{\Lambda}_{pi}$}
            \Let{$\sigma(\split_L^{(0)}\times\cluster_k)$}{0, $\forall k\leq \Kmax$}\COMMENT{Arrays of size $\Kmax$}
            \Let{$\sigma(\split_R^{(0)}\times\cluster_k)$}{$\sum_{\x_i \in \leaf} \vec{\omega}_{ki}$, $\forall k\leq \Kmax$}
            \Let{$\sigma(\leaf\times\leaf)$}{$\sigma(\split_R^{(0)}\times\split_R^{(0)})$}
            \Let{$K$}{$\card{\{k \text{ s.t. } \card{\cluster_k}\neq 0\}}$}\COMMENT{Current number of clusters}

            \Let{$\Delta\objective$, \texttt{BestSplit}}{0, $\emptyset$}\COMMENT{Best split so far}

            \FOR {$l \gets 1 \textrm{ to } |\leaf|-1$}
                \Let{$\alpha,\beta$}{0,0}
                \FOR {$l^\prime \gets 1 \textrm{ to } |\leaf|$}
                    \IF {$l^\prime < l$}
                        \Let{$\alpha$}{$\alpha + \kappa_{\nu(l), \nu(l^\prime)}$}
                    \ENDIF
                    \IF {$l^\prime > l$}
                        \Let{$\beta$}{$\beta + \kappa_{\nu(l), \nu(l^\prime)}$}
                    \ENDIF
                \ENDFOR
                \STATE Update $\sigma(\split_L^{(l)}\times\split_L^{(l)})$, $\sigma(\split_R^{(l)}\times\split_R^{(l)})$, $\sigma(\split_L^{(l)}\times\cluster_k)$ and $\sigma(\split_R^{(l)}\times\cluster_k)$ using equations~\ref{eq:update_self_left_split}, \ref{eq:update_self_right_split}, \ref{eq:update_cluster_left_split}, \ref{eq:update_cluster_right_split}.

                \Let{$\tilde{\Delta\objective}, \texttt{split}$}{\texttt{ComputeSplits}{$\sigma(\split_L^{(l)}\times\split_L^{(l)})$, $l$, $\sigma(\split_R^{(l)}\times\split_R^{(l)})$, $|\leaf|-1-l$, $\text{diag}(\vec{\gamma})$, $|C_k|$, $\sigma(\split_L^{(l)}\times \cluster_k)$, $\sigma(\split_R^{(l)}\times\cluster_k)$, $\sigma(\leaf\times\leaf)$, $k$, $K$}}
                \IF {$\tilde{\Delta\objective} > \Delta\objective$}
                    \Let{$\Delta\objective$}{$\tilde{\Delta\objective}$}
                    \Let{\texttt{BestSplit}}{$\texttt{split} \cup (\nu(l))$} \COMMENT{The split gives the left target, the right target, the sample on which the split is done}
                \ENDIF
            \ENDFOR
            \STATE Return $\Delta\objective$, \texttt{BestSplit}
  \end{algorithmic}
\end{algorithm}

%% file: figs_tex/fig_alpha_beta.tex
\begin{figure}
    \centering
    \subfloat[][There is no ordering of the samples]{
        \begin{scaletikzpicturetowidth}{0.3\linewidth}\begin{tikzpicture}[scale=\tikzscale]
            \draw[step=1.0, black, thick] (0, 0) grid (4,4);
            \draw[fill={rgb, 255:red, 0; green, 255; blue, 0}, fill opacity = 1] (2,2) rectangle ++(1,1);
            \draw[fill={rgb, 255:red, 0; green, 255; blue, 0}, fill opacity = 1] (2,3) rectangle ++(1,1);
            \draw[fill={rgb, 255:red, 255; green, 0; blue, 0}, fill opacity = 1] (2,0) rectangle ++(1,1);
            
            \foreach \i in {1, ..., 4}
            	\draw[pattern = north west lines, pattern color=gray] ($(-1, 4)+(\i,-\i)$) rectangle ++(1,1);
            
            \draw (0.5,-0.5) node [anchor=center][inner sep=0.75pt]   [align=center] {A};
            \draw (1.5,-0.5) node [anchor=center][inner sep=0.75pt]   [align=center] {B};
            \draw (2.5,-0.5) node [anchor=center][inner sep=0.75pt]   [align=center] {C};
            \draw (3.5,-0.5) node [anchor=center][inner sep=0.75pt]   [align=center] {D};
            \draw (-0.5,3.5) node [anchor=center][inner sep=0.75pt]   [align=center] {A};
            \draw (-0.5,2.5) node [anchor=center][inner sep=0.75pt]   [align=center] {B};
            \draw (-0.5,1.5) node [anchor=center][inner sep=0.75pt]   [align=center] {C};
            \draw (-0.5,0.5) node [anchor=center][inner sep=0.75pt]   [align=center] {D};
        
        \end{tikzpicture}\end{scaletikzpicturetowidth}\label{sfig:alpha_beta_no_order}
    }
    \subfloat[][There exists an ordering $\nu$]{
        \begin{scaletikzpicturetowidth}{0.3\linewidth}\begin{tikzpicture}[scale=\tikzscale]
            \draw[step=1.0, black, thick] (0, 0) grid (4,4);
            \draw[fill={rgb, 255:red, 0; green, 255; blue, 0}, fill opacity = 1] (2,2) rectangle ++(1,1);
            \draw[fill={rgb, 255:red, 0; green, 255; blue, 0}, fill opacity = 1] (2,0) rectangle ++(1,1);
            \draw[fill={rgb, 255:red, 255; green, 0; blue, 0}, fill opacity = 1] (2,3) rectangle ++(1,1);
            
            \foreach \i in {1, ..., 4}
                \draw[pattern = north west lines, pattern color=gray] ($(-1, 4)+(\i,-\i)$) rectangle ++(1,1);
            
            \draw (0.5,-0.5) node [anchor=center][inner sep=0.75pt]   [align=center] {A};
            \draw (1.5,-0.5) node [anchor=center][inner sep=0.75pt]   [align=center] {B};
            \draw (2.5,-0.5) node [anchor=center][inner sep=0.75pt]   [align=center] {C};
            \draw (3.5,-0.5) node [anchor=center][inner sep=0.75pt]   [align=center] {D};
            \draw (-0.5,3.5) node [anchor=center][inner sep=0.75pt]   [align=center] {A};
            \draw (-0.5,2.5) node [anchor=center][inner sep=0.75pt]   [align=center] {B};
            \draw (-0.5,1.5) node [anchor=center][inner sep=0.75pt]   [align=center] {C};
            \draw (-0.5,0.5) node [anchor=center][inner sep=0.75pt]   [align=center] {D};
        
        \end{tikzpicture}\end{scaletikzpicturetowidth}\label{sfig:alpha_beta_nu}
    }
    \subfloat[][Reordering according to $\nu$]{
        \begin{scaletikzpicturetowidth}{0.3\linewidth}\begin{tikzpicture}[scale=\tikzscale]
            \draw[step=1.0, black, thick] (0, 0) grid (4,4);
            \draw[fill={rgb, 255:red, 0; green, 255; blue, 0}, fill opacity = 1] (2,2) rectangle ++(1,1);
            \draw[fill={rgb, 255:red, 0; green, 255; blue, 0}, fill opacity = 1] (2,3) rectangle ++(1,1);
            \draw[fill={rgb, 255:red, 255; green, 0; blue, 0}, fill opacity = 1] (2,0) rectangle ++(1,1);
            
            \foreach \i in {1, ..., 4}
            	\draw[pattern = north west lines, pattern color=gray] ($(-1, 4)+(\i,-\i)$) rectangle ++(1,1);
            
            \draw (0.5,-0.5) node [anchor=center][inner sep=0.75pt]   [align=center] {B};
            \draw (1.5,-0.5) node [anchor=center][inner sep=0.75pt]   [align=center] {D};
            \draw (2.5,-0.5) node [anchor=center][inner sep=0.75pt]   [align=center] {C};
            \draw (3.5,-0.5) node [anchor=center][inner sep=0.75pt]   [align=center] {A};
            \draw (-0.5,3.5) node [anchor=center][inner sep=0.75pt]   [align=center] {B};
            \draw (-0.5,2.5) node [anchor=center][inner sep=0.75pt]   [align=center] {D};
            \draw (-0.5,1.5) node [anchor=center][inner sep=0.75pt]   [align=center] {C};
            \draw (-0.5,0.5) node [anchor=center][inner sep=0.75pt]   [align=center] {A};
        
        \end{tikzpicture}\end{scaletikzpicturetowidth}\label{sfig:alpha_beta_sorted_nu}
    }
    \caption{An example of the value of the variables $\alpha_3$ and $\beta_3$ which respectively are the sum of green squares and red squares on the kernel matrix of the elements A, B, C and D in a leaf. In \ref{sfig:alpha_beta_nu} and \ref{sfig:alpha_beta_sorted_nu}, the ordering is $\nu(\{1,2,3,4\}) = \{B,D,C,A\}$.}
    \label{fig:visualisation_alpha_beta}
\end{figure}
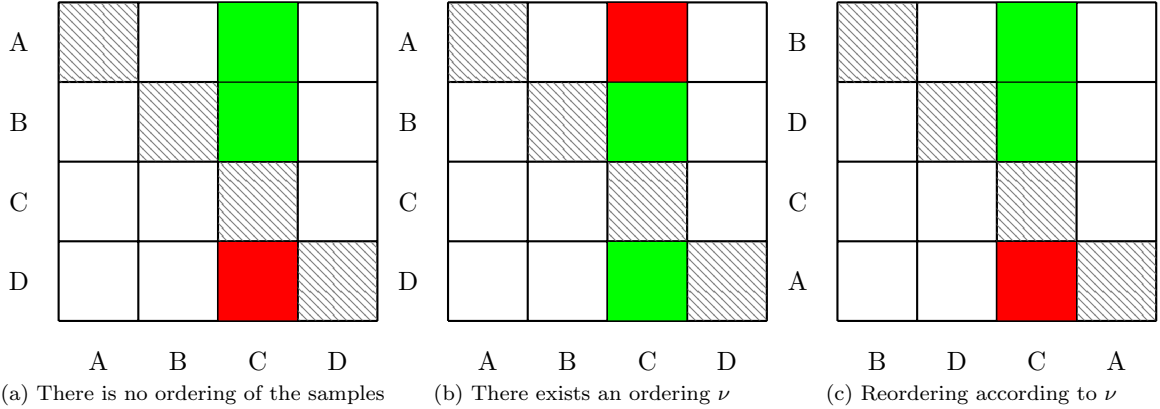

%% file: figs_tex/alg_kauri.tex
\begin{algorithm}
  \caption{TrainKAURI}\label{alg:tree_training}
    \begin{algorithmic}
        \STATE \textbf{Input :} $\dataset=\{\vec{x}_i\}_{i=1}^n$ a dataset, $\vec{x}_i\in \mathbb{R}^d$, $\Kmax>=2$ the maximum number of allowed clusters, $\dmax \in \{1, d\}$ the maximum number of feature to consider per split, $\Tmax \leq n$ the maximum number of leaves
            \Let{$\vec{\kappa}$}{$\langle\varphi(\dataset), \varphi(\dataset)\rangle$} \COMMENT{Kernel value of samples, $n\times n$}
            \STATE Initialise $\vec{Z}$ and $\vec{Y}$ \COMMENT{All samples in one leaf, leaf belongs to only one cluster}
            \STATE Initialise the tree structure in \texttt{Tree}.
            \Let{\texttt{Leaves}}{\texttt{List}(0)}\COMMENT{Only one starting leaf to explore}
            \Let{$\Delta\objective$}{$\infty$}\COMMENT{Last gain value}
            \Let{\texttt{BestSplit}}{$\emptyset$}\COMMENT{The best split proposal}
            \WHILE{$\texttt{Leaves}\neq\emptyset \wedge |\texttt{Tree}|\leq \Tmax \wedge \bar{\Delta\objective}>0$}
                \Let{$\Delta\objective$}{0}\COMMENT{Best split achieved so far}
                \Let{$\vec{\Lambda}$}{$\vec{Z}\vec{\kappa}$}\COMMENT{Compute $\sigma(\leaf_p \times \{\x_j\})$}
                \Let{$\vec{\omega}$}{$\vec{Y}\vec{\Lambda}$}\COMMENT{Compute $\sigma(\cluster_k \times \{\x_j\})$}
                \Let{$\vec{\gamma}$}{$\vec{\omega}\vec{Z}^\top\vec{Y}^\top$}\COMMENT{Compute $\sigma(\cluster_k \times \cluster_{k^\prime})$}
                \Let{$\card{\cluster_k}$}{$\vec{Y}\vec{Z}\vec{1}_n$}\COMMENT{Sizes of clusters}

                \FOR {$p \in \texttt{Leaves}$}
                    \Let{$\leaf_p$}{$\{\x_i | \vec{Z}_{ji}==1\}$} \COMMENT{Find the indices of leaf $p$}
                    \Let{$k$}{$\text{argmax}_{k^\prime} \vec{Y}_{k^\prime,p}$} \COMMENT{The current cluster of leaf $p$}
                    \FOR {$f \gets 1 \textrm{ to } \dmax$}
                        \Let{$\nu$}{Argsort($\{\vec{x}_{if} | \x_i\in\leaf_p\}$)}
                        \Let{$\tilde{\Delta\objective}, \texttt{split}$}{\texttt{FindBestSplit}{$\leaf$, $j$, $\nu$, $\vec{\kappa}$, $\Lambda$, $\vec{\omega}$, $\vec{\gamma}$, $|\cluster_k|$, $k$}}
                        \IF {$\tilde{\Delta\objective} > \Delta\objective$}
                            \Let{$\Delta\objective$}{$\tilde{\Delta\objective}$}
                            \Let{\texttt{BestSplit}}{\texttt{split} $\cup$ $(p, f)$} \COMMENT{Add node and feature information to the best split}
                        \ENDIF
                    \ENDFOR
                \ENDFOR

                \IF {$\Delta\objective >0$}
                    \STATE Remove the best leaf from the list \texttt{Leaves} and add the children of the split in \texttt{Leaves} if they satisfy structural constraints.
                    \STATE Update \texttt{Tree} using \texttt{BestSplit}
                    \STATE Update $\vec{Z}$ and $\vec{Y}$.
                \ENDIF
            \ENDWHILE

            \STATE Return \texttt{Tree}
  \end{algorithmic}
\end{algorithm}

%% file: appendix/dataset_characteristics.tex
\input{figs_tex/tab_datasets}

We summarise here in Table~\ref{tab:dataset_summary} the characteristics of the tested datasets for benchmarking. We insert as well all remaining tables of the benchmark that could not fit the main part of the paper.

\input{figs_tex/tab_benchmark_small_ari_extended}
\input{figs_tex/tab_benchmark_small_kmeans_extended}
\input{figs_tex/tab_benchmark_small_wad_extended}
\input{figs_tex/tab_kernels_ari_extended}
\input{figs_tex/tab_kernels_kmeans_extended}

%% file: figs_tex/tab_datasets.tex
\begin{table}
    \centering
    \caption{Summary of the datasets used in the experiments. *The number of features may be slightly larger than the actual number of variables as discrete variables were one-hot encoded.}
    \label{tab:dataset_summary}
    \begin{tabular}{c c c c}
        \toprule
         Name & Samples & Features & Classes \\
         \cmidrule{2-4}
         Atom&800&3&2\\
         Avila&20,867&10&12\\
         Breast Cancer&683&9&2\\
         Car evaluation*&1,728&21&4\\
         Chainlink&1,000&3&2\\
         US Congress&435&16&2\\
         Engytime&4,096&2&2\\
         Digits&1,797&64&10\\
         Haberman Survival&306&3&2\\
         Hepta&212&3&7\\
         Iris&150&4&3\\
         Lsun&404&3&3\\
         Mice protein&552&77&8\\
         Target&770&2&6\\
         Tetra&400&3&4\\
         Twodiamonds&800&2&2\\
         Vowel&990&10&2\\
         Wine&178&13&3\\
         Wingnut&1,016&2&2\\
         \bottomrule
    \end{tabular}
\end{table}

%% file: figs_tex/tab_benchmark_small_ari_extended.tex
\begin{table*}
    \centering
    \caption{Extended ARI scores \std{std} (greater is better) after 30 runs on random subsamples of 80\% of the remaining datasets. Entries marked X were not run because of excessive runtime due to large numbers of features. All models are limited to finding as many leaves as clusters.}
    \label{tab:benchmark_small_ari_extended}
    \begin{tabular}{ccccccc}
        \toprule
        Dataset & Kauri & KMeans+DT & ICOT & IMM & ExShallow & RDM \\
        \midrule
        Avila & 0.02\std{0.02} & 0.04\std{0.02} & X & \textbf{0.05\std{0.03}} & 0.04\std{0.02} & \textbf{0.05\std{0.03}} \\
        Car & 0.04\std{0.06} & \textbf{0.07\std{0.08}} & X & 0.06\std{0.06} & 0.06\std{0.06} & \textbf{0.07\std{0.05}} \\
        Congress & 0.48\std{0.02} & 0.47\std{0.04} & \textbf{0.49\std{0.03}} & 0.47\std{0.04} & 0.48\std{0.02} & 0.39\std{0.02} \\
        Engytime & \textbf{0.51\std{0.01}} & \textbf{0.51\std{0.01}} & X & 0.50\std{0.01} & 0.49\std{0.01} & \textbf{0.51\std{0.01}} \\
        Haberman & \textbf{-0.00\std{0.00}} & \textbf{-0.00\std{0.00}} & \textbf{-0.00\std{0.00}} & \textbf{-0.00\std{0.00}} & \textbf{-0.00\std{0.00}} & \textbf{-0.00\std{0.00}} \\
        Tetra & 0.94\std{0.07} & \textbf{1.00\std{0.00}} & \textbf{1.00\std{0.00}} & \textbf{1.00\std{0.00}} & \textbf{1.00\std{0.00}} & 0.61\std{0.03} \\
        Twodiamonds & \textbf{1.00\std{0.00}} & \textbf{1.00\std{0.00}} & \textbf{1.00\std{0.00}} & \textbf{1.00\std{0.00}} & \textbf{1.00\std{0.00}} & 0.98\std{0.01} \\
        Vowel & 0.02\std{0.02} & 0.02\std{0.02} & X & 0.03\std{0.03} & \textbf{0.04\std{0.06}} & \textbf{0.04\std{0.03}} \\
        Wingnut & 0.15\std{0.01} & 0.15\std{0.01} & \textbf{0.46\std{0.38}} & 0.15\std{0.01} & 0.15\std{0.01} & 0.13\std{0.01} \\
        \bottomrule
\end{tabular}
\end{table*}

%% file: figs_tex/tab_benchmark_small_kmeans_extended.tex
\begin{table*}
    \centering
    \caption{KMeans score \std{std} (lower is better) after 30 runs on subsamples of 80\% of the remaining datasets divided by the KMeans reference score (=1.0). All models are limited to finding as many leaves as clusters.}
    \label{tab:benchmark_small_kmeans_extended}
    \begin{tabular}{ccccccc}
        \toprule
        Dataset & Kauri & KMeans+DT & ICOT & IMM & ExShallow & RDM \\
        \midrule
        Atom & 1.03\std{0.02} & 1.04\std{0.02} & 1.05\std{0.03} & \textbf{1.02\std{0.02}} & 1.04\std{0.02} & 1.03\std{0.02} \\
        Avila & \textbf{1.45\std{0.58}} & 3.33\std{1.16} & X & 1.49\std{0.66} & 1.52\std{0.71} & 1.99\std{0.68} \\
        Cancer & 1.14\std{0.03} & 1.14\std{0.03} & \textbf{1.09\std{0.04}} & 1.14\std{0.03} & 1.14\std{0.03} & 1.37\std{0.04} \\
        Car & \textbf{1.00\std{0.00}} & \textbf{1.00\std{0.00}} & X & \textbf{1.00\std{0.00}} & \textbf{1.00\std{0.00}} & 1.02\std{0.03} \\
        Chainlink & \textbf{1.02\std{0.01}} & \textbf{1.02\std{0.01}} & 1.04\std{0.02} & \textbf{1.02\std{0.01}} & \textbf{1.02\std{0.01}} & 1.03\std{0.01} \\
        Congress & 1.09\std{0.01} & 1.09\std{0.02} & \textbf{1.08\std{0.01}} & 1.09\std{0.01} & 1.09\std{0.02} & 1.19\std{0.02} \\
        Digits & 1.21\std{0.01} & 1.24\std{0.02} & X & 1.25\std{0.02} & \textbf{1.20\std{0.01}} & 1.38\std{0.04} \\
        Engytime & 1.14\std{0.04} & 1.15\std{0.05} & X & \textbf{1.13\std{0.04}} & 1.14\std{0.06} & 1.15\std{0.05} \\
        Haberman & 1.10\std{0.07} & \textbf{1.08\std{0.04}} & 1.09\std{0.04} & 1.10\std{0.07} & \textbf{1.08\std{0.04}} & 1.10\std{0.06} \\
        Hepta & 1.09\std{0.09} & 1.08\std{0.03} & 3.12\std{1.94} & \textbf{1.07\std{0.04}} & 1.08\std{0.03} & 1.36\std{0.09} \\
        Iris & 1.18\std{0.06} & 1.19\std{0.05} & 1.92\std{0.10} & 1.17\std{0.05} & \textbf{1.16\std{0.04}} & 1.46\std{0.10} \\
        Lsun & 1.09\std{0.04} & 1.09\std{0.04} & 1.14\std{0.12} & \textbf{1.07\std{0.03}} & \textbf{1.07\std{0.04}} & 1.66\std{0.09} \\
        Mice & \textbf{1.08\std{0.02}} & 1.11\std{0.04} & X & 1.14\std{0.05} & \textbf{1.08\std{0.02}} & 1.41\std{0.06} \\
        Target & 1.35\std{0.05} & 1.58\std{0.13} & X & \textbf{1.19\std{0.08}} & \textbf{1.19\std{0.03}} & 1.83\std{0.10} \\
        Tetra & 1.12\std{0.15} & \textbf{1.04\std{0.02}} & \textbf{1.04\std{0.01}} & \textbf{1.04\std{0.02}} & 1.05\std{0.02} & 1.53\std{0.06} \\
        Twodiamonds & 1.05\std{0.02} & \textbf{1.04\std{0.02}} & \textbf{1.04\std{0.02}} & 1.05\std{0.02} & \textbf{1.04\std{0.02}} & \textbf{1.04\std{0.02}} \\
        Vowel & \textbf{1.07\std{0.01}} & 1.08\std{0.01} & X & 1.09\std{0.02} & \textbf{1.07\std{0.01}} & 1.11\std{0.01} \\
        Wine & 1.15\std{0.05} & 1.15\std{0.08} & 1.37\std{0.17} & \textbf{1.11\std{0.04}} & \textbf{1.11\std{0.04}} & 1.54\std{0.10} \\
        Wingnut & \textbf{1.09\std{0.02}} & \textbf{1.09\std{0.01}} & 1.14\std{0.06} & \textbf{1.09\std{0.02}} & \textbf{1.09\std{0.02}} & \textbf{1.09\std{0.01}} \\
        \bottomrule
    \end{tabular}
\end{table*}

%% file: figs_tex/tab_benchmark_small_wad_extended.tex
\begin{table*}
    \centering
    \caption{Extended WAD scores \std{std} (lower is better) after 30 runs on random subsamples of 80\% of the remaining datasets. All models are limited to finding as many leaves as clusters.}
    \label{tab:benchmark_small_wad_extended}
    \begin{tabular}{cccccc}
        \toprule
        Dataset & Kauri & KMeans+DT & IMM & ExShallow & RDM \\
        \midrule
        Atom & \textbf{2.00\std{0.00}} & \textbf{2.00\std{0.00}} & \textbf{2.00\std{0.00}} & \textbf{2.00\std{0.00}} & \textbf{2.00\std{0.00}} \\
        Avila & 5.37\std{0.13} & \textbf{5.12\std{0.17}} & 8.16\std{0.72} & 5.74\std{0.55} & 7.82\std{0.69} \\
        Cancer & \textbf{2.00\std{0.00}} & \textbf{2.00\std{0.00}} & \textbf{2.00\std{0.00}} & \textbf{2.00\std{0.00}} & \textbf{2.00\std{0.00}} \\
        Car & \textbf{3.00\std{0.00}} & 3.06\std{0.06} & 3.04\std{0.05} & 3.04\std{0.06} & \textbf{3.00\std{0.12}} \\
        Chainlink & \textbf{2.00\std{0.00}} & \textbf{2.00\std{0.00}} & \textbf{2.00\std{0.00}} & \textbf{2.00\std{0.00}} & \textbf{2.00\std{0.00}} \\
        Congress & \textbf{2.00\std{0.00}} & \textbf{2.00\std{0.00}} & \textbf{2.00\std{0.00}} & \textbf{2.00\std{0.00}} & \textbf{2.00\std{0.00}} \\
        Engytime & \textbf{2.00\std{0.00}} & \textbf{2.00\std{0.00}} & \textbf{2.00\std{0.00}} & \textbf{2.00\std{0.00}} & \textbf{2.00\std{0.00}} \\
        Haberman & \textbf{2.00\std{0.00}} & \textbf{2.00\std{0.00}} & \textbf{2.00\std{0.00}} & \textbf{2.00\std{0.00}} & \textbf{2.00\std{0.00}} \\
        Tetra & 3.23\std{0.08} & 3.25\std{0.02} & 3.25\std{0.02} & 3.26\std{0.02} & \textbf{3.14\std{0.03}} \\
        Twodiamonds & \textbf{2.00\std{0.00}} & \textbf{2.00\std{0.00}} & \textbf{2.00\std{0.00}} & \textbf{2.00\std{0.00}} & \textbf{2.00\std{0.00}} \\
        Vowel & \textbf{2.00\std{0.00}} & \textbf{2.00\std{0.00}} & \textbf{2.00\std{0.00}} & \textbf{2.00\std{0.00}} & \textbf{2.00\std{0.00}} \\
        Wingnut & \textbf{2.00\std{0.00}} & \textbf{2.00\std{0.00}} & \textbf{2.00\std{0.00}} & \textbf{2.00\std{0.00}} & \textbf{2.00\std{0.00}} \\
        \bottomrule
    \end{tabular}
\end{table*}

%% file: figs_tex/tab_kernels_ari_extended.tex
\begin{table*}[th]
\centering
\caption{ARI scores \std{std} (greater is better) after 30 runs on random subsamples of 80\% of the input datasets for varying kernels. All models are limited to finding 4 times more leaves than clusters. *: Datasets that suffered from the kernel KMeans convergence to empty clusters.} 
\label{tab:benchmark_kernels_ari_extended}
\resizebox{\linewidth}{!}{\begin{tabular}{ccccccccc}
\toprule
Kernel & \multicolumn{2}{c}{Additive $\chi^2$} & \multicolumn{2}{c}{$\chi^2$} & \multicolumn{2}{c}{Laplacian} & \multicolumn{2}{c}{RBF} \\
\cmidrule(lr){2-3}\cmidrule(lr){4-5}\cmidrule(lr){6-7}\cmidrule(lr){8-9}
Dataset & Kauri & KMeans+DT & Kauri & KMeans+DT & Kauri & KMeans+DT & Kauri & KMeans+DT \\
\midrule
Atom & 0.10\std{0.01} & 0.09\std{0.02} & 0.09\std{0.01} & 0.11\std{0.03} & \textbf{0.44\std{0.37}} & 0.15\std{0.04} & 0.18\std{0.03} & 0.14\std{0.02} \\
Avila* & 0.02\std{0.02} & 0.00\std{0.03} & 0.02\std{0.02} & -0.00\std{0.01} & \textbf{0.05\std{0.02}} & 0.04\std{0.04} & 0.03\std{0.03} & 0.00\std{0.01} \\
Car* & 0.06\std{0.06} & 0.06\std{0.06} & 0.06\std{0.06} & 0.02\std{0.04} & 0.05\std{0.06} & 0.06\std{0.06} & \textbf{0.07\std{0.06}} & 0.06\std{0.06} \\
Chainlink & 0.32\std{0.02} & 0.28\std{0.08} & 0.33\std{0.01} & 0.26\std{0.13} & \textbf{0.40\std{0.05}} & 0.36\std{0.11} & 0.11\std{0.01} & 0.09\std{0.02} \\
Congress & 0.50\std{0.04} & \textbf{0.56\std{0.03}} & 0.37\std{0.04} & 0.36\std{0.23} & 0.50\std{0.04} & \textbf{0.56\std{0.03}} & 0.49\std{0.04} & 0.55\std{0.04} \\
Digits & 0.52\std{0.03} & 0.55\std{0.03} & 0.11\std{0.03} & 0.22\std{0.04} & 0.53\std{0.04} & \textbf{0.57\std{0.03}} & 0.56\std{0.02} & 0.54\std{0.04} \\
Engytime & 0.66\std{0.03} & 0.77\std{0.03} & 0.67\std{0.02} & 0.78\std{0.02} & 0.57\std{0.07} & 0.70\std{0.02} & 0.73\std{0.04} & \textbf{0.82\std{0.01}} \\
Haberman & -0.01\std{0.00} & 0.00\std{0.05} & -0.01\std{0.01} & \textbf{0.02\std{0.07}} & -0.00\std{0.00} & -0.00\std{0.00} & -0.00\std{0.00} & -0.00\std{0.00} \\
Hepta & \textbf{1.00\std{0.00}} & 0.48\std{0.21} & \textbf{1.00\std{0.00}} & 0.62\std{0.19} & \textbf{1.00\std{0.00}} & 0.82\std{0.14} & \textbf{1.00\std{0.00}} & 0.57\std{0.25} \\
Mice* & \textbf{0.23\std{0.02}} & 0.19\std{0.04} & 0.19\std{0.02} & 0.21\std{0.04} & 0.20\std{0.01} & 0.20\std{0.02} & 0.21\std{0.01} & 0.17\std{0.05} \\
Target* & \textbf{0.63\std{0.01}} & 0.38\std{0.21} & \textbf{0.63\std{0.02}} & 0.49\std{0.20} & \textbf{0.63\std{0.05}} & \textbf{0.63\std{0.07}} & \textbf{0.63\std{0.02}} & 0.39\std{0.23} \\
Tetra & 0.99\std{0.01} & 0.91\std{0.03} & 0.99\std{0.01} & 0.93\std{0.03} & \textbf{1.00\std{0.00}} & \textbf{1.00\std{0.00}} & \textbf{1.00\std{0.00}} & \textbf{1.00\std{0.00}} \\
Vowel & 0.05\std{0.08} & 0.11\std{0.05} & 0.12\std{0.07} & \textbf{0.14\std{0.08}} & 0.11\std{0.03} & 0.11\std{0.04} & 0.10\std{0.03} & 0.11\std{0.04} \\
Wingnut & 0.13\std{0.01} & 0.29\std{0.04} & 0.14\std{0.02} & 0.29\std{0.03} & 0.17\std{0.13} & \textbf{0.44\std{0.17}} & 0.15\std{0.01} & 0.42\std{0.03} \\
\bottomrule
\end{tabular}}
\end{table*}

%% file: figs_tex/tab_kernels_kmeans_extended.tex
\begin{table*}[th]
\centering
\caption{Relative Kernel KMeans scores \std{std} (lower is better) of Kauri and Kernel-KMeans + Decision Tree after 30 runs on random subsamples of 80\% of the input datasets for varying kernels. All models are limited to finding 4 times more leaves than clusters. *: Datasets that suffered from the kernel KMeans convergence to empty clusters.} 
\label{tab:benchmark_kernels_kmeans_extended}
\resizebox{\linewidth}{!}{\begin{tabular}{ccccccccc}
\toprule
Kernel & \multicolumn{2}{c}{Additive $\chi^2$} & \multicolumn{2}{c}{$\chi^2$} & \multicolumn{2}{c}{Laplacian} & \multicolumn{2}{c}{RBF} \\
\cmidrule(lr){2-3}\cmidrule(lr){4-5}\cmidrule(lr){6-7}\cmidrule(lr){8-9}
Dataset & Kauri & KMeans+DT & Kauri & KMeans+DT & Kauri & KMeans+DT & Kauri & KMeans+DT \\
\midrule
Atom & 1.05\std{0.03} & 1.09\std{0.08} & 1.05\std{0.02} & 1.10\std{0.07} & \textbf{1.00\std{0.02}} & 1.03\std{0.02} & 1.03\std{0.02} & 1.04\std{0.02} \\
Avila* & \textbf{0.47\std{0.12}} & 0.94\std{0.24} & 0.93\std{0.25} & 1.45\std{0.23} & 1.14\std{0.28} & 1.36\std{0.30} & 0.73\std{0.28} & 1.38\std{0.43} \\
Car* & \textbf{1.00\std{0.00}} & 1.01\std{0.01} & \textbf{1.00\std{0.00}} & \textbf{1.00\std{0.00}} & \textbf{1.00\std{0.00}} & 1.01\std{0.01} & \textbf{1.00\std{0.00}} & 1.01\std{0.01} \\
Chainlink & 1.03\std{0.01} & 1.04\std{0.03} & 1.02\std{0.01} & 1.03\std{0.04} & \textbf{1.01\std{0.00}} & \textbf{1.01\std{0.02}} & 1.02\std{0.01} & 1.02\std{0.02} \\
Congress & 1.05\std{0.02} & 1.03\std{0.01} & \textbf{0.99\std{0.00}} & 1.01\std{0.01} & 1.03\std{0.01} & 1.02\std{0.01} & 1.05\std{0.01} & 1.03\std{0.01} \\
Digits & 1.07\std{0.01} & 1.08\std{0.02} & \textbf{0.99\std{0.00}} & 1.00\std{0.00} & 1.04\std{0.01} & 1.05\std{0.01} & 1.07\std{0.01} & 1.10\std{0.03} \\
Engytime & 1.09\std{0.04} & 1.06\std{0.05} & 1.09\std{0.04} & 1.07\std{0.04} & \textbf{1.03\std{0.02}} & \textbf{1.03\std{0.02}} & 1.09\std{0.04} & 1.06\std{0.04} \\
Haberman & \textbf{1.05\std{0.04}} & 1.08\std{0.08} & \textbf{1.05\std{0.03}} & 1.10\std{0.09} & \textbf{1.05\std{0.03}} & 1.07\std{0.05} & 1.06\std{0.06} & 1.09\std{0.07} \\
Hepta & \textbf{1.01\std{0.03}} & 4.33\std{2.01} & 1.10\std{0.05} & 3.34\std{1.65} & 1.03\std{0.02} & 1.31\std{0.28} & 1.05\std{0.03} & 5.88\std{3.22} \\
Mice* & 1.01\std{0.02} & 1.11\std{0.13} & \textbf{0.98\std{0.01}} & 1.02\std{0.01} & 1.02\std{0.01} & 1.04\std{0.01} & 1.02\std{0.02} & 1.12\std{0.13} \\
Target* & 1.07\std{0.08} & 2.08\std{0.76} & 1.12\std{0.05} & 1.71\std{0.69} & \textbf{1.02\std{0.02}} & 1.09\std{0.12} & 1.10\std{0.04} & 2.32\std{0.96} \\
Tetra & 1.05\std{0.02} & 1.06\std{0.03} & \textbf{1.02\std{0.02}} & 1.04\std{0.03} & \textbf{1.02\std{0.01}} & \textbf{1.02\std{0.01}} & 1.03\std{0.02} & \textbf{1.02\std{0.01}} \\
Vowel & 1.04\std{0.02} & 1.03\std{0.02} & 1.03\std{0.01} & 1.03\std{0.01} & \textbf{1.01\std{0.00}} & 1.02\std{0.01} & 1.02\std{0.01} & 1.03\std{0.02} \\
Wingnut & 1.07\std{0.02} & 1.03\std{0.01} & 1.07\std{0.02} & 1.03\std{0.02} & 1.03\std{0.01} & \textbf{1.02\std{0.01}} & 1.09\std{0.02} & 1.05\std{0.02} \\
\bottomrule
\end{tabular}}
\end{table*}

%% file: appendix/example_imm.tex
When trying to motivate their algorithm, \citet[Figure 2b]{moshkovitz_explainable_2020} create a simple dataset where the combination of KMeans+Tree would solve the task with excellent accuracy, yet with non-optimal splits.

\input{figs_tex/fig_imm_dataset}

This dataset consists in 3 clusters. The first two ones are respectively drawn from $\mathcal{N}([2,0]^\top, \epsilon \vec{I}_2)$ and $\mathcal{N}([-2,0], \epsilon\vec{I}_2)$ with $\epsilon$ small enough. The last cluster contains two points located at $(-2, v)$ and $(2, v)$. We plot in Figure~\ref{fig:imm_dataset} a sample of such dataset for $v=1000$.

A decision tree learning from KMeans labels will start by separating the samples along the x-axis. This non-optimal choice then requires two splits on the left- and right-hand sides to then separate the Gaussian distributions from the third cluster.

The optimal choice, achieved by ExKMC as well as Kauri, starts by cutting on the y-axis, separating thus all Gaussian distributions from the third cluster. A single split afterwards is sufficient for separating the two Gaussian distributions.

This shows that the explanation quality brought by Kauri can be of better quality for the same clustering results.

%% file: figs_tex/fig_imm_dataset.tex
\begin{figure}
    \centering
    \includegraphics[width=0.5\linewidth]{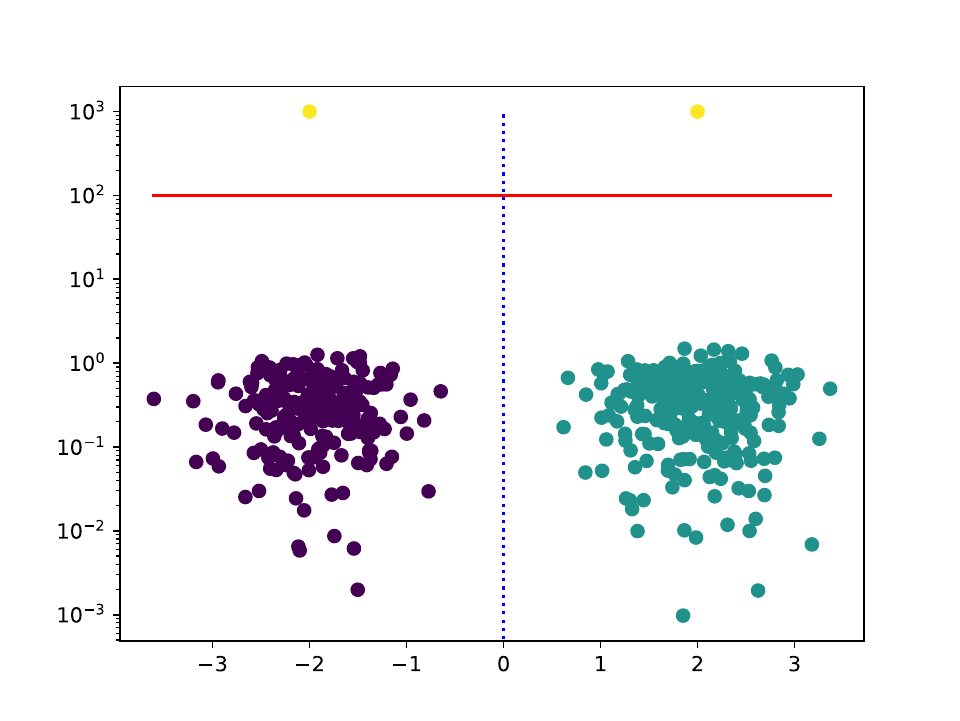}
    \caption{A dataset proposed by~\citet{moshkovitz_explainable_2020} consisting in two isotropic Gaussian distributions and a cluster of two points distant on the y-axis. In order to split optimally the clusters, a decision tree should start with a y-axis split (solid red line) then use an x-axis split (dashed blue line) to separate the two Gaussian distributions.}
    \label{fig:imm_dataset}
\end{figure}